\documentclass[preprint,10pt,nopreprintline]{elsarticle}

\usepackage[a4paper,top=2cm,bottom=2.2cm,left=2cm,right=2cm,footskip=1cm]{geometry}
\usepackage[T1]{fontenc}
\usepackage[utf8]{inputenc}
\usepackage{amsmath}
\IfFileExists{newtxtext.sty}{%
  \usepackage{newtxtext}\usepackage{newtxmath}}{%
  \usepackage{mathptmx}\usepackage{amssymb}}  % fallback if newtx is unavailable
\usepackage[protrusion=true,expansion=true]{microtype}

\usepackage{bm}

\usepackage{booktabs,tabularx,multirow,makecell,longtable,array}
\usepackage{pdflscape}
\usepackage{pifont}
\newcommand{\cmark}{\ding{51}}
\newcommand{\xmark}{\ding{55}}
\usepackage[table]{xcolor}
\renewcommand{\arraystretch}{1.12}

\usepackage{graphicx}
\usepackage[font=small,labelfont=bf,skip=4pt]{caption}
\usepackage{subcaption}
\usepackage{float}
\usepackage[below]{placeins}

\usepackage{algorithm}
\usepackage{algpseudocode}

\usepackage{enumitem}
\setlist{itemsep=1pt,topsep=3pt,parsep=0pt,leftmargin=*}

\biboptions{numbers,sort&compress}
\usepackage{xurl}
\usepackage[colorlinks=true,linkcolor=blue!60!black,citecolor=blue!60!black,urlcolor=blue!60!black]{hyperref}
\usepackage[capitalise,nameinlink]{cleveref}
\newcommand{\model}{ChromaNest}
\newcommand{\ssip}{SSIP}

\newif\ifhl
\ifdefined\highlightchanges \hltrue \else \hlfalse \fi
\definecolor{revblue}{RGB}{0,0,220}
\newcommand{\rev}[1]{\ifhl{\color{revblue}#1}\else#1\fi}
\newcommand{\revon}{\ifhl\color{revblue}\fi}
\newcommand{\zenodoDOI}{10.5281/zenodo.24403413}

\journal{Computer Methods and Programs in Biomedicine}

\begin{document}
% Plain-text PDF metadata (avoids hyperref "Token not allowed in a PDF string" warnings
% caused by the footnote and corresponding-author marks inside \author).
\hypersetup{pdftitle={ChromaNest: Stain-Conditioned Nested Graph Attention with Optical-Density Edge Priors for Leakage-Audited H\&E Histopathology Classification},
            pdfauthor={Touhid Alam, Sumaiya Sarkar Shimla, Md. Manzurul Hasan, Md. Rafiqul Islam}}

\begin{frontmatter}

\title{ChromaNest: Stain-Conditioned Nested Graph Attention with Optical-Density Edge Priors for Leakage-Audited H\&E Histopathology Classification}

\author[aiub]{Touhid Alam\fnref{eq}}
\ead{26-93934-1@student.aiub.edu}
\author[aiub]{Sumaiya Sarkar Shimla\fnref{eq}}
\ead{26-93975-1@student.aiub.edu}
\author[aiub]{Md. Manzurul Hasan\fnref{sup}}
\ead{manzurul@aiub.edu}
\author[aiub]{Md. Rafiqul Islam\corref{cor1}}
\ead{dmri@aiub.edu}
\cortext[cor1]{Corresponding author.}
\fntext[eq]{These authors contributed equally to this work and share first authorship.}
\fntext[sup]{Supervisor.}
\affiliation[aiub]{organization={Department of Computer Science, American International University-Bangladesh (AIUB)},
            city={Dhaka},
            postcode={1229},
            country={Bangladesh}}

\begin{abstract}
\noindent\textbf{Background and Objective:} Deep-learning classifiers for haematoxylin and eosin (H\&E) tiles often report near-ceiling accuracy, but public benchmarks are frequently split so that near-duplicates of test tiles appear in training, stain chemistry is treated purely as a nuisance, and interpretability claims are rarely audited. We aimed to build a lightweight classifier that uses stain physics inside the network and to evaluate it under leakage-audited and patient-isolated protocols.

\noindent\textbf{Methods:} \model{} is a stain-aware nested graph transformer (5.33\,M parameters, 3.04\,GFLOPs per tile). A non-learned, per-image front-end separates H\&E optical densities; stain differences enter windowed attention among 784 patch nodes as learnable edge penalties, stain-conditioned keys softly assign the nodes to 16 compartment slots, and compartment context returns to every node through zero-initialised feature-wise modulation. We evaluated six public datasets (150{,}581 images)\rev{, splitting each at the finest level its metadata allow: patient-grouped cross-validation where patient identifiers exist (BreakHis), source-image grouping on lymphoma, a patient-disjoint external cohort, and otherwise Source-Separated Image Partitioning (\ssip{}), which groups perceptual-hash near-duplicates before splitting. Hyperparameters, checkpoints and calibration temperatures were chosen on validation data only.}

\noindent\textbf{Results:} Under 5-fold patient-grouped cross-validation on BreakHis, \model{} reached $91.84\pm2.37\%$ accuracy, and on the external 50-patient CRC-VAL-HE-7K cohort, without adaptation, $96.84\pm0.14\%$. Random tile splits leaked near-duplicates for 86.9\% (LC25000), 100\% (lymphoma) and 35.9\% (oral cancer) of test images. Under image-level \ssip{}, five-seed accuracies ranged from $94.66\%$ (Kather2016) to $99.88\%$ (LC25000), with calibration error $\le 2.16\%$ after temperature scaling. Stain conditioning contributed at least as much macro-F1 as the entire compartment hierarchy on all five ablated datasets ($+0.65$ to $+2.40$ vs.\ $+0.28$ to $+1.40$\,pp), almost entirely through the per-image H\&E optical-density input ($+0.60$ to $+2.15$\,pp); the stain edge penalty added only $+0.05$ to $+0.25$\,pp, within seed variation. Replacing the stain representation with RGB cost 1.53--2.60\,pp, and on BreakHis \model{} lost 34--38\% less macro-F1 than Swin-Tiny and ConvNeXt-Tiny under $\pm30\%$ stain drift.

\noindent\textbf{Conclusions:} To our knowledge, \model{} is among the first tile classifiers to embed Beer--Lambert H\&E optical densities directly into both local message passing and hierarchical pooling. It matches or approaches substantially larger models with about five times fewer parameters, exposes 72 inspectable stain and gating priors, and is accompanied by a leakage-audited, patient-isolated evaluation protocol and openly available notebooks, providing a reproducible and resource-efficient basis for computational pathology.

\noindent\textbf{Availability:} The Kaggle notebooks that generate all results are on Zenodo (\url{https://doi.org/\zenodoDOI}) and require the Kaggle environment to run.
\end{abstract}


\end{frontmatter}

% =====================================================================
%  Section 1 -- Introduction
% =====================================================================
\section{Introduction}
\label{sec:introduction}

Cancer remains one of the heaviest burdens on global health. The most recent GLOBOCAN estimates report close to 20 million new cancer cases and 9.7 million cancer deaths in 2022, and demographic projections indicate that annual incidence will reach about 35 million cases by 2050~\cite{bray2024globocan}. The organs studied in this work account for a large share of this burden: lung cancer was the most frequently diagnosed cancer, with almost 2.5 million new cases (12.4\% of all cancers), followed by female breast (11.6\%) and colorectal cancer (9.6\%); lung cancer was also the leading cause of cancer death, with an estimated 1.8 million deaths (18.7\%), while colorectal and breast cancers caused 9.3\% and 6.9\% of cancer deaths, respectively~\cite{bray2024globocan}. Beyond these common epithelial tumours, haematological malignancies such as lymphoma and head-and-neck cancers such as oral squamous cell carcinoma also depend on tissue examination for a definitive diagnosis. For almost every solid tumour, the diagnosis, histological subtype and grade are established by a pathologist who examines a stained tissue section under the microscope, so histopathology sits on the critical path between clinical suspicion and treatment~\cite{srinidhi2021deep,vanderlaak2021deep}.

Histopathological diagnosis is a reading of tissue architecture. A tissue section is not a uniform texture but an organised arrangement of cells within compartments: epithelial cells line glands and surfaces, fibroblasts and collagen form the supporting stroma, lymphocytes cluster in immune aggregates, and fat, mucus, necrotic debris and background fill the remaining space. Pathologists judge malignancy at two coupled levels. At the cellular level they assess nuclear size, shape, chromatin texture and pleomorphism; at the tissue level they assess how cells are organised, for example whether glands are preserved, whether tumour cells invade the stroma, and how immune cells are distributed around them. Neither level is sufficient on its own: the same nucleus can be benign in one architectural context and suspicious in another.

This reading is made possible by haematoxylin and eosin (H\&E) staining. H\&E has been used for at least a century and remains essential for recognising tissue types and the morphological changes that form the basis of contemporary cancer diagnosis~\cite{fischer2008hematoxylin}. Haematoxylin, applied with a metal mordant, binds nucleic acids and colours nuclei deep blue-purple, revealing chromatin patterns that are cell-type and cancer-type specific; eosin is an acidic dye that binds proteins non-specifically and colours cytoplasm, collagen and other extracellular matrix in shades of pink~\cite{fischer2008hematoxylin,chan2014wonderful}. The two dyes therefore encode chemically distinct information: the haematoxylin signal marks nuclei and the cellular level, while the eosin signal marks cytoplasm and stroma and hence the tissue compartments in which cells are embedded~\cite{chan2014wonderful}. Because absorbed light follows the Beer--Lambert law, the optical density of a stained pixel is approximately a linear mixture of the two dye concentrations, which allows the stains to be separated computationally by colour deconvolution~\cite{ruifrok2001quantification,macenko2009method}. In practice, however, the colour of H\&E varies with fixation, section thickness, staining time, reagent batch and scanner, so the same tissue can look markedly different from one laboratory to the next~\cite{tellez2019quantifying,hoque2024stain}.

The manual workflow built on this stain is under growing strain. In the United States, the number of active pathologists fell by 17.53\% between 2007 and 2017 while the diagnostic workload per pathologist, adjusted for new cancer cases, rose by 41.73\%~\cite{metter2019workforce}. The shortage is far more severe in low-resource settings, where sub-Saharan Africa has been estimated to have roughly one pathologist per million people, about fifty times fewer than high-income countries~\cite{wilson2018access}. Manual reading is also subjective. In a study of 6{,}900 breast-biopsy interpretations, individual pathologists agreed with an expert consensus diagnosis in only 75.3\% of cases, and agreement fell to 48\% for atypia~\cite{elmore2015concordance}. Stain variability adds a further, purely technical source of disagreement that affects both human readers and algorithms~\cite{tellez2019quantifying}.

Deep learning has become the dominant computational answer to these pressures. Convolutional neural networks (CNNs) classify tissue tiles, detect tumour regions and predict outcomes directly from H\&E images~\cite{srinidhi2021deep,campanella2019clinical}, Vision Transformers (ViTs) and windowed transformers such as Swin model dependencies across image patches~\cite{dosovitskiy2021image,liu2021swin}, and pathology foundation models pretrained on millions of histology images now provide general-purpose tissue representations~\cite{wang2022ctranspath,chen2024uni}. In parallel, graph neural networks (GNNs) have emerged as a natural fit for tissue, because a histology image is fundamentally a set of interacting biological entities rather than a regular pixel grid~\cite{ahmedt2022survey,brussee2025graph}. Hierarchical entity graphs such as HACT-Net link a cell graph to a tissue graph and outperform patch-based CNNs on breast-cancer subtyping~\cite{pati2022hierarchical}, while Vision GNN showed that a generic image can be processed as a graph of patch nodes~\cite{han2022vision}. In the broader machine-learning literature, differentiable graph pooling methods such as DiffPool~\cite{ying2018hierarchical} and MinCutPool~\cite{bianchi2020spectral}, as well as latent-slot bottlenecks such as Slot Attention~\cite{locatello2020object} and Perceiver~\cite{jaegle2021perceiver}, have shown how fine-grained nodes or tokens can be softly clustered into a small set of coarse latent nodes. Relatedly, Graph-in-Graph (GiG) learning~\cite{zaripova2023gig} and GiGCN~\cite{jia2024gigcn} nest internal graphs within external graphs.

Despite this progress, systematic reviews of computational pathology point to limitations that remain unresolved for tile-level histopathology classification. First, most histopathology entity graphs rely on offline nuclei segmentation or superpixels before learning~\cite{ahmedt2022survey,pati2022hierarchical,jia2024gigcn}; errors in pre-processing propagate into the graph, and its topology cannot adapt end to end~\cite{brussee2025graph}. Conversely, generic token-clustering and windowed attention backbones~\cite{liu2021swin,ying2018hierarchical,locatello2020object} operate purely on abstract feature similarity, ignoring the physical chemistry of H\&E staining. Stain is treated almost exclusively as a nuisance to be removed by normalisation or absorbed by colour augmentation~\cite{tellez2019quantifying,hoque2024stain}, even though haematoxylin and eosin concentrations carry specific information about nuclear and stromal compartments~\cite{fischer2008hematoxylin,chan2014wonderful}.

The second group of limitations concerns how tile classifiers are evaluated and explained. In public benchmarks lacking patient or slide metadata, tiles, crops and augmented copies of the same source image are frequently split at random across training and test sets. In digital pathology, cross-split leakage has been shown to inflate predictive scores by up to 41\%~\cite{bussola2021slipping}, echoing broader reproducibility warnings across machine learning~\cite{kapoor2023leakage,varoquaux2022machine}. Explanations face a similar credibility gap. Post-hoc saliency maps such as Grad-CAM~\cite{selvaraju2017gradcam} and built-in attention weights are usually inspected only qualitatively, even though saliency and attention maps can be unfaithful to the underlying decision~\cite{adebayo2018sanity,poceviciute2020survey,hense2024xmil}.

To address these limitations, we propose \textbf{\model{}}, a \emph{Chromatic Nested Graph Transformer} for histopathology image classification (\cref{fig:overview}). Building on windowed local attention~\cite{liu2021swin}, differentiable soft graph pooling~\cite{ying2018hierarchical,locatello2020object} and nested graph hierarchies~\cite{zaripova2023gig,jia2024gigcn}, \model{} introduces domain-specific H\&E stain conditioning at both levels of a cell-to-compartment hierarchy. A non-learned, per-image stain front-end separates haematoxylin and eosin optical densities on the fly and passes them into the graph operators: stain differences act as learnable edge penalties during local message passing, and node stain profiles condition the soft assignment of 784 patch nodes into 16 tissue compartments.

The paper therefore makes two co-equal contributions, one architectural and one methodological. Architecturally, it asks whether stain chemistry, used as a signal inside the network, rather than the graph hierarchy as such, is what makes a small model competitive; the ablations answer that it is, with most of the gain coming from the per-image H\&E optical-density representation rather than from the stain penalty, and the learned stain priors and the model's behaviour under stain drift are consistent with the histology of each task. Methodologically, it shows that the benchmarks on which near-ceiling tile accuracies are reported leak near-duplicates on a scale that makes many published numbers hard to interpret, and it supplies a reusable partitioning protocol and released split files that remove this leakage.

The main contributions of this work are as follows:
\begin{itemize}
  \item \textbf{A stain-conditioned, segmentation-free nested graph architecture.} \model{} combines windowed local patch-node attention~\cite{liu2021swin} (784 patch nodes across 49 local $4\times4$ neighbourhoods) with differentiable soft assignment~\cite{ying2018hierarchical,locatello2020object} to 16 learnable tissue-compartment nodes. Unlike generic vision transformers or graph poolers, both local message passing and cell-to-compartment assignment are conditioned on per-image Macenko H\&E optical densities (with a Ruifrok--Johnston fallback)~\cite{macenko2009method,ruifrok2001quantification}. Stain conditioning is the component that drives the gains: on every ablated dataset it adds at least as much five-seed macro-F1 as the whole compartment hierarchy, and replacing the per-image stain representation with RGB input lowers macro-F1 by 1.53--2.60\,pp (\cref{sec:r_ablation}). The gain comes almost entirely from the per-image H\&E optical-density input ($+0.60$ to $+2.15$\,pp, V1$\to$V2); the stain edge penalty adds only $+0.05$ to $+0.25$\,pp (V2$\to$V3), within seed variation, and its role is to make the stain conditioning inspectable rather than to raise accuracy.
  \item \textbf{A co-equal methodological contribution: a leakage audit of six public benchmarks and a leakage-free partitioning protocol (\ssip{}), paired with patient- and cohort-isolated evaluation.} We group images by connected components of a dihedral-invariant perceptual hash (or by source image on Lymphoma) before splitting. Our audit demonstrates that random tile splits place near-duplicate or same-source partners in training for 86.9\% of LC25000, 100\% of Lymphoma and 35.9\% of Oral Cancer test tiles. Because perceptual hashing isolates near-duplicates rather than patients or arbitrary-angle augmentations (\cref{sec:d_leakage}), we provide the notebooks that regenerate the \ssip{} group assignments and splits for reuse as leakage-free benchmark splits, and we pair \ssip{} with strict patient-grouped cross-validation on BreakHis ($91.84 \pm 2.37\%$) and external cohort evaluation without adaptation on CRC-VAL-HE-7K ($96.84 \pm 0.14\%$).
  \item \textbf{Structurally transparent stain-affinity priors and bidirectional coupling.} Optical-density differences enter local attention as learnable, head-specific edge penalties $\mathrm{softplus}(\beta_H)\lvert\Delta h\rvert+\mathrm{softplus}(\beta_E)\lvert\Delta e\rvert$, while outer compartments interact via a gated mixture of grid-contact spatial adjacency and semantic attention before returning context to cells via zero-initialised FiLM~\cite{perez2018film}. Together, the model exposes 72 inspectable scalar priors that reveal how strongly each depth relies on nuclear versus stromal stain coherence. Without supervision, and in a single representative run, these priors became haematoxylin-dominant in all four blocks on the nuclear-atypia tasks (BreakHis, LC25000) and eosin-leaning in the deep blocks on the stroma- and keratin-rich tasks (Kather2016, Oral Cancer). This pattern is consistent with stain chemistry, but it has not been validated against pathologist annotations. The stain-conditioned model also degrades less under stain drift than generic backbones of comparable or larger size (\cref{sec:r_graphs,sec:r_robust}).
  \item \textbf{A parameter-efficient design with linear scaling in node count.} Windowed local attention and a fixed-size ($M=16$) compartment graph keep every operator $\mathcal{O}(N)$ in the number of cell nodes $N$. The complete model has 5.33\,M parameters and requires 3.04\,GFLOPs per tile.
  \item \textbf{A quantitative structural-transparency audit.} We show that the learned graph topology encodes diagnostic class (a linear classifier on 43 graph descriptors reaches 94.11\% on BreakHis and 89.7--99.1\% across datasets), that the stain priors differ between datasets in a way consistent with tissue histology (single run, not pathologist-validated), and that the compartment rankings are highly stable under synthetic stain shift (Spearman 0.926--0.993), complementing post-hoc attribution methods such as LIME and Integrated Gradients.
\end{itemize}

The remainder of the paper is organised as follows. \Cref{sec:related} reviews related work. \Cref{sec:methodology} details the datasets, \ssip{} and the \model{} architecture. \Cref{sec:results} reports classification, external validation, baseline comparisons, ablations, graph-prior mechanics, calibration and robustness. \Cref{sec:discussion} discusses limitations and clinical implications, and \cref{sec:conclusion} concludes the paper.

%% =====================================================================
%% 2. RELATED WORK
%% =====================================================================
\section{Related Work}\label{sec:related}

We review recent studies across four strands: task-specific tile classifiers, pathology foundation models, hierarchical graph and token-pooling architectures, and work on stain handling, dataset bias and explanation faithfulness (\cref{tab:rw_summary}).

\subsection{Task-specific CNN, ViT and hybrid tile classifiers}\label{sec:rw_tile}

Most recent work on the six benchmarks used here designs compact CNNs or CNN--transformer hybrids and reports accuracies close to 100\%. On LC25000, \citet{hasan2024lwms} reached 99.20\% on the five-class task with an eight-layer multi-scale CNN of 1.1\,M parameters explained by Grad-CAM and SHAP, and \citet{jim2025xllcnet} obtained $99.62 \pm 0.16$\% on the three lung classes with XLLC-Net (3\,M parameters). On BreakHis, the 7.97\,M-parameter DCT-HistoTransformer of \citet{ranjbar2024dct} low-pass filters tokens in the frequency domain before attention and achieved $96.00\pm0.48$\% binary accuracy. For colorectal tissue, \citet{abdelghany2024efficientnet} fine-tuned EfficientNet-B0 with an adaptive learning rate, reaching 99.87\% on NCT-CRC-HE-100K and 99.00\% accuracy (94.36\% macro-F1) on the external CRC-VAL-HE-7K cohort. The hybrid TransNetV reported 98.54\% on CRC-VAL-HE-7K and 98.96\% on Kather-5K~\cite{tanveer2024transnetv}. On lymphoma, PrecisionLymphoNet stacked InceptionV3 and Xception ($\approx$47\,M parameters) to reach 99\%~\cite{rajadurai2024precisionlymphonet}, and HCTN-LC reported 99.87\% accuracy~\cite{sikkandar2025hctnlc}. On oral squamous cell carcinoma, EfficientNetB3 with LIME explanations achieved 98.33\%~\cite{islam2024oscclime}.

Three weaknesses are common across these studies. First, colour is processed as generic RGB or frequency content without explicit H\&E physics. Second, explanations are displayed as qualitative heatmaps without faithfulness testing. Third, almost all evaluate on random tile-level splits of datasets generated by augmenting a small pool of originals: LC25000 derives from 1{,}250 source images, the 7{,}480 lymphoma tiles come from 374 source images, and the oral cancer dataset derives from 231 normal and 747 OSCC originals augmented $\approx10\times$ and $3\times$ respectively~\cite{diniz2025oscc}.

\subsection{Pathology foundation models}\label{sec:rw_fm}

Foundation models replace task-specific design with self-supervised pre-training at scale. CTransPath~\cite{wang2022ctranspath} pre-trained a hybrid CNN--Swin backbone on 15\,M patches from TCGA and PAIP. UNI~\cite{chen2024uni} trains a ViT-L/16 ($\approx$300\,M parameters) with DINOv2 on more than 100\,M patches from 100{,}426 slides. CONCH~\cite{lu2024conch} aligns a 90\,M-parameter image encoder with a text encoder on 1.17\,M image--caption pairs. Virchow~\cite{vorontsov2024virchow} scales a ViT to 632\,M parameters on $\approx$1.5\,M slides, and Prov-GigaPath~\cite{xu2024gigapath} pre-trains on 1.3\,B tiles from 171{,}189 slides. These models provide strong general-purpose embeddings, but they require tens to hundreds of millions of parameters, are frequently released under gated access, and collapse each tile into a black-box vector without explicit cell--compartment graph structure or inspectable stain priors.

\subsection{Windowed attention, differentiable graph pooling and graph-in-graph learning}\label{sec:rw_graph}

\model{} builds on three architectural primitives from computer vision and graph learning. First, restricting self-attention to local non-overlapping windows with learnable relative positional bias was established by the Swin Transformer~\cite{liu2021swin} to achieve linear complexity in image size. Second, softly assigning fine-grained nodes to a small set of learned coarse nodes is the core mechanism of differentiable graph pooling—notably DiffPool~\cite{ying2018hierarchical} and MinCutPool~\cite{bianchi2020spectral}—and is closely related to object-centric token clustering in Slot Attention~\cite{locatello2020object} and cross-attention bottlenecks in Perceiver~\cite{jaegle2021perceiver}. Third, nesting graphs across scales was formalised as Graph-in-Graph (GiG) learning by \citet{zaripova2023gig} (who link graph-valued samples inside a population graph) and adapted to imagery by GiGCN~\cite{jia2024gigcn} (which embeds superpixel graphs inside a spatial adjacency graph).

In computational pathology and vision, other graph architectures explore parts of this space. Vision GNN (ViG)~\cite{han2022vision} and ViHGNN~\cite{han2023vihgnn} connect image patches via feature-space $k$-NN or fuzzy hypergraphs at a single scale. HACT-Net~\cite{pati2022hierarchical} builds a cell-to-tissue hierarchy for breast pathology but relies on offline nuclei segmentation and superpixels~\cite{ahmedt2022survey,brussee2025graph}. HIGT~\cite{guo2023higt} introduces bidirectional cross-level interaction across a whole-slide patch pyramid, and GAFR-Net~\cite{gao2026gafrnet} applies graph attention across an image-similarity graph on BreakHis. However, none of these methods conditions local message passing or differentiable node pooling on physical H\&E optical-density profiles, nor do they combine grid-contact spatial adjacency with semantic attention at the coarse compartment level.

\subsection{Stain handling, dataset bias and explanation faithfulness}\label{sec:rw_eval}

Stain variation is typically addressed via external preprocessing or augmentation. \citet{franchet2024bias} and \citet{zamanitajeddin2024dg} demonstrated that stain augmentation (such as AugmentHE) and self-supervised pre-training outperform classical stain normalisation for domain generalisation. Yet in all such pipelines, stain decomposition is discarded before or absorbed silently inside the classifier. Meanwhile, \citet{ignatov2024nctcrche} exposed low-level colour and JPEG shortcuts in NCT-CRC-HE, showing that simple ImageNet backbones can match or exceed foundation models when colour statistics leak class identity. Finally, \citet{hense2024xmil} and \citet{adebayo2018sanity} demonstrated that attention weights and saliency maps often fail quantitative faithfulness tests, motivating our separation of structural transparency from post-hoc attribution faithfulness.

\subsection{Patch and cell graph models, token clustering, attention biases and site effects}\label{sec:rw_more}

\paragraph{Patch and cell graph models.} CGC-Net~\cite{zhou2019cgcnet} builds a cell graph over segmented nuclei and pools it with learned clustering to grade colorectal cancer, and HACT-Net~\cite{pati2022hierarchical} links a nuclei graph to a superpixel tissue graph; both depend on offline segmentation. Patch-GCN~\cite{chen2021patchgcn} and SlideGraph+~\cite{lu2022slidegraph} connect pretrained patch embeddings across a whole-slide image for survival and HER2 prediction, and HIPT~\cite{chen2022hipt} nests ViTs over fixed $16\to256\to4096$-pixel windows of a whole slide. Relative to these, \model{} operates \emph{within} a single tile, learns its nodes end to end from pixels without segmentation or a frozen encoder, replaces a fixed spatial upper level with a learned soft grouping into compartments, and conditions both message passing and grouping on H\&E optical densities.

\paragraph{Token clustering.} TCFormer~\cite{zeng2022tcformer} merges vision tokens by progressive clustering so that tokens adapt their shape to semantic regions, and GroupViT~\cite{xu2022groupvit} assigns tokens to learnable group tokens with a hard (straight-through) assignment learned from text supervision. Both merge tokens irreversibly according to appearance features. \model{} instead keeps every patch node, conditions the assignment keys on a physical stain profile, mixes spatial-contact and semantic adjacency among the groups, and feeds group context back to the patch nodes through FiLM.

\paragraph{Attention-bias methods.} Graphormer~\cite{ying2021graphormer} adds learned biases indexed by shortest-path distance and edge type to attention logits, and ALiBi~\cite{press2022alibi} subtracts a fixed per-head slope times token distance. The stain penalty in \cref{eq:attn} belongs to this additive-bias family. What \model{} adds is the distance it uses: a content-derived difference in haematoxylin and eosin concentration computed per image, with learnable non-negative slopes per head and per dye that can be read out after training, used alongside Swin's relative position bias~\cite{liu2021swin}.

\paragraph{Site effects.} Howard et al.~\cite{howard2021site} showed that site-specific digital histology signatures survive colour normalisation and augmentation and inflate accuracy unless training and validation sites are separated. \ssip{} addresses a narrower failure, near-duplicate and same-source leakage when patient or site metadata are absent; it does not separate acquisition sites, which is why we also report patient-grouped cross-validation and an external cohort.

\subsection{Positioning of this study}\label{sec:rw_gap}

Relative to prior work, \model{} does not claim to invent windowed attention~\cite{liu2021swin} or soft latent pooling~\cite{ying2018hierarchical,locatello2020object}; rather, its architectural contribution lies in \emph{domain-specific H\&E stain conditioning} across a two-level cell-to-compartment hierarchy. Specifically, per-image Macenko optical densities parameterise (i) learnable head-specific edge penalties in local window attention and (ii) key projections in soft compartment assignment, while the coarse compartments interact through a gated mixture of spatial grid contact and semantic attention before modulating cell features via zero-initialised FiLM~\cite{perez2018film}. We pair this 5.33\,M-parameter architecture with leakage-audited image-level (\ssip{}), patient-grouped and external-cohort evaluations, alongside an audit of structural transparency versus perturbation faithfulness.

{\scriptsize
\setlength{\tabcolsep}{2.5pt}
\renewcommand{\arraystretch}{1.12}
\begin{longtable}{@{}p{0.95cm}p{2.4cm}p{2.0cm}p{1.35cm}p{2.6cm}p{2.6cm}p{3.5cm}@{}}
\caption{Summary of related studies (2023--2026) alongside foundational graph-in-graph work. Parameters are those reported or derivable from the cited architecture. Published accuracies were obtained under the original authors' splitting protocols.}\label{tab:rw_summary}\\
\toprule
\textbf{Ref.} & \textbf{Method} & \textbf{Proposed model} & \textbf{Params} & \textbf{Results} & \textbf{Significance} & \textbf{Limitations}\\
\midrule
\endfirsthead
\multicolumn{7}{@{}l}{\small\itshape \Cref{tab:rw_summary} (continued)}\\
\toprule
\textbf{Ref.} & \textbf{Method} & \textbf{Proposed model} & \textbf{Params} & \textbf{Results} & \textbf{Significance} & \textbf{Limitations}\\
\midrule
\endhead
\midrule \multicolumn{7}{r}{\itshape continued on next page}\\
\endfoot
\bottomrule
\endlastfoot
\multicolumn{7}{@{}l}{\textbf{(a) Task-specific CNN / ViT / hybrid tile classifiers}}\\
\cite{hasan2024lwms} 2024 & Lightweight multi-scale CNN; Grad-CAM + SHAP & LW-MS CNN & 1.1\,M & LC25000 (5-class) Acc 99.20\% & Sub-2\,M CNN saturates LC25000 & Random split over augmentation-derived tiles; explanations not tested for faithfulness\\
\cite{jim2025xllcnet} 2025 & 4-conv-layer CNN; saliency + Grad-CAM & XLLC-Net & 3\,M & LC25000 lung (3-class) Acc $99.62 \pm 0.16$\% & Fast, compact, explainable & Lung-only subset; same leakage risk; no faithfulness test\\
\cite{ranjbar2024dct} 2024 & DCT low-pass attention + MobileConv branch & DCT-HistoTransformer & 7.97\,M & BreakHis binary $96.00\pm0.48$\%; 8-class $87.85 \pm 0.93$\% & Frequency-domain token reduction cuts ViT cost & RGB only, no H\&E decomposition; split isolation not reported\\
\cite{abdelghany2024efficientnet} 2024 & EfficientNet-B0 fine-tuning with adaptive LR & EfficientNet-B0 (ALR) & $\approx$5.3\,M & NCT-100K Acc 99.87\%; CRC-VAL-7K Acc 99.00\%, F1 94.36\% & Reports external cohort; shows external F1 drop & Generic backbone; NCT colour/JPEG shortcuts unaddressed; no XAI\\
\cite{tanveer2024transnetv} 2024 & Optimised CNN--transformer hybrid & TransNetV & Not Reported & CRC-VAL-7K 98.54\%; Kather-5K 98.96\% & Local + global features transfer across CRC sets & Regular token grid; no cell--tissue relations; no source-group control\\
\cite{rajadurai2024precisionlymphonet} 2024 & Stacked ensemble of ImageNet CNNs & PrecisionLymphoNet & $\approx$47\,M & Lymphoma (CLL/FL/MCL) Acc 99\% & Strong transfer-learning reference & Heavy ensemble; tile-level split over 374 source images (100\% same-source test tiles in our audit)\\
\cite{sikkandar2025hctnlc} 2025 & Dual conv/transformer paths + fusion module & HCTN-LC & Not Reported & Lymphoma Acc 99.87\%, AUC 0.9991 & Latest ceiling on lymphoma benchmark & No source-level split; params not reported; no histological prior\\
\cite{islam2024oscclime} 2024 & CNN comparison; LIME explanations & EfficientNetB3 + LIME & $\approx$12\,M & Oral (Normal/OSCC) Acc 98.33\% & Explainable OSCC baseline & 126-image test set; LIME not stain-aware; no faithfulness test\\
\cite{diniz2025oscc} 2025 & Standardised CNN benchmark & CNN comparison & Not Reported & Shows 231/747 originals augmented $\approx10\times$/$3\times$ & Exposes augmentation imbalance on Oral data & Extended abstract; does not propose a de-duplicated split\\
\midrule
\multicolumn{7}{@{}l}{\textbf{(b) Pathology foundation models}}\\
\cite{chen2024uni} 2024 & DINOv2 SSL on Mass-100K (100\,M patches, 100{,}426 WSIs) & UNI (ViT-L/16) & $\approx$300\,M & SOTA on 34 tasks; CRC-100K-Raw bal.\ Acc 0.925 & Scale replaces task-specific design & Global embedding; no cell/tissue graph; heavy; gated licence\\
\cite{lu2024conch} 2024 & CoCa visual--language pre-training, 1.17\,M pairs & CONCH & 90\,M (img) + 110\,M (text) & SOTA on 14 benchmarks; CRC-100K-Raw bal.\ Acc 0.941 & Zero-shot transfer via language & Heatmap explanations; dataset colour shortcuts unaddressed\\
\cite{vorontsov2024virchow} 2024 & DINOv2 on $\approx$1.5\,M MSKCC slides & Virchow (ViT) & 632\,M & Pan-cancer specimen AUC 0.95 (9 common + 7 rare) & Data scale improves rare-cancer detection & Proprietary data; high compute; no mechanistic interpretability\\
\cite{xu2024gigapath} 2024 & Tile encoder + LongNet slide encoder; 1.3\,B tiles, 171{,}189 slides & Prov-GigaPath & Not Reported & Best on 25 / 26 tasks & Whole-slide context modelling & Context between, not within, tiles; very high cost\\
\midrule
\multicolumn{7}{@{}l}{\textbf{(c) Graph, hypergraph and hierarchical pooling models}}\\
\cite{zaripova2023gig} 2023 & Node-level GNN + learned population graph + NDDL & GiG (original) & Not Reported & Gains over graph baselines on HCP and bioinformatic sets & Founding learned graph-over-graphs & Inner graphs given, not learned from pixels; outer graph links samples; bottom-up only\\
\cite{jia2024gigcn} 2024 & Superpixel internal graphs + external adjacency graph & GiGCN & Not Reported & Effective on 4 HSI scenes with few labels & First GiG for dense imagery & Non-differentiable superpixels; fixed adjacency; no top-down feedback\\
\cite{han2023vihgnn} 2023 & Fuzzy-$C$-means hypergraph over patches & ViHGNN & Not Reported & ImageNet top-1 83.9\%; COCO AP 43.1 & Higher-order patch relations & Natural images; feature-only hyperedges; single level\\
\cite{guo2023higt} 2023 & Hierarchical GNN + ViT with bidirectional interaction over WSI pyramid & HIGT & Not Reported & ESCA typing AUC $96.81$, Acc $96.16$\% & Bidirectional cross-level messaging in pathology & Patch (not cell) nodes; pyramid-fixed graph; no stain decomposition\\
\cite{gao2026gafrnet} 2026 & Image-similarity graph + GAT + fuzzy rules & GAFR-Net & Not Reported & BreakHis AUC up to $0.9641$ ($100\times$) & Topology carries diagnostic signal & Transductive across images; no intra-image structure; preprint\\
\midrule
\multicolumn{7}{@{}l}{\textbf{(d) Stain handling, bias, leakage and explanation faithfulness}}\\
\cite{franchet2024bias} 2024 & Stain augmentation (AugmentHE) + fast normalisation (HEnorm) & AugmentHE / HEnorm & -- & +81\% colour dispersion vs.\ geometric aug. & Model stain variation during training & Stain remains preprocessing; not used inside the model\\
\cite{zamanitajeddin2024dg} 2024 & 30 DG algorithms, 3 tasks, 7{,}560 runs & DG benchmark + HISTOPANTUM & -- & SSL and stain augmentation most effective & Largest controlled DG comparison & Backbones fixed; in-model stain use unexplored\\
\cite{ignatov2024nctcrche} 2024 & Dataset audit of NCT-CRC-HE & Bias analysis & -- & Mean RGB $>$50\% Acc; EfficientNet-B0 97.7\% on CRC-VAL & Low-level shortcuts inflate benchmarks & Diagnostic only; no leakage-aware protocol proposed\\
\cite{hense2024xmil} 2024 & LRP-based MIL explanations; faithfulness tests & xMIL-LRP & -- & Attention explanations often misleading; LRP more faithful & Faithfulness as the explanation standard & Slide-level MIL; no tile graphs or stain attribution\\
\midrule
\multicolumn{7}{@{}l}{\textbf{This work}}\\
\textbf{Chroma\-Nest} & Stain-aware nested graph transformer; leakage-audited image-level (\ssip{}), patient-grouped \& external evaluation; structural-transparency audit & \model{} & \textbf{5.33\,M} (3.04 GFLOPs) & BreakHis patient-CV $91.84 \pm 2.37\%$; external CRC-VAL-7K $96.84 \pm 0.14\%$ & Stain-conditioned window attention \& soft pooling; leakage-audited benchmarks & Tile-level (not WSI); pHash isolates near-duplicates, not patients (\cref{sec:d_limitations})\\
\end{longtable}
\normalsize}

%% =====================================================================
%% 3. METHODOLOGY
%% =====================================================================
\section{Methodology}\label{sec:methodology}

\begin{figure}[!htbp]
\centering
\includegraphics[width=\textwidth]{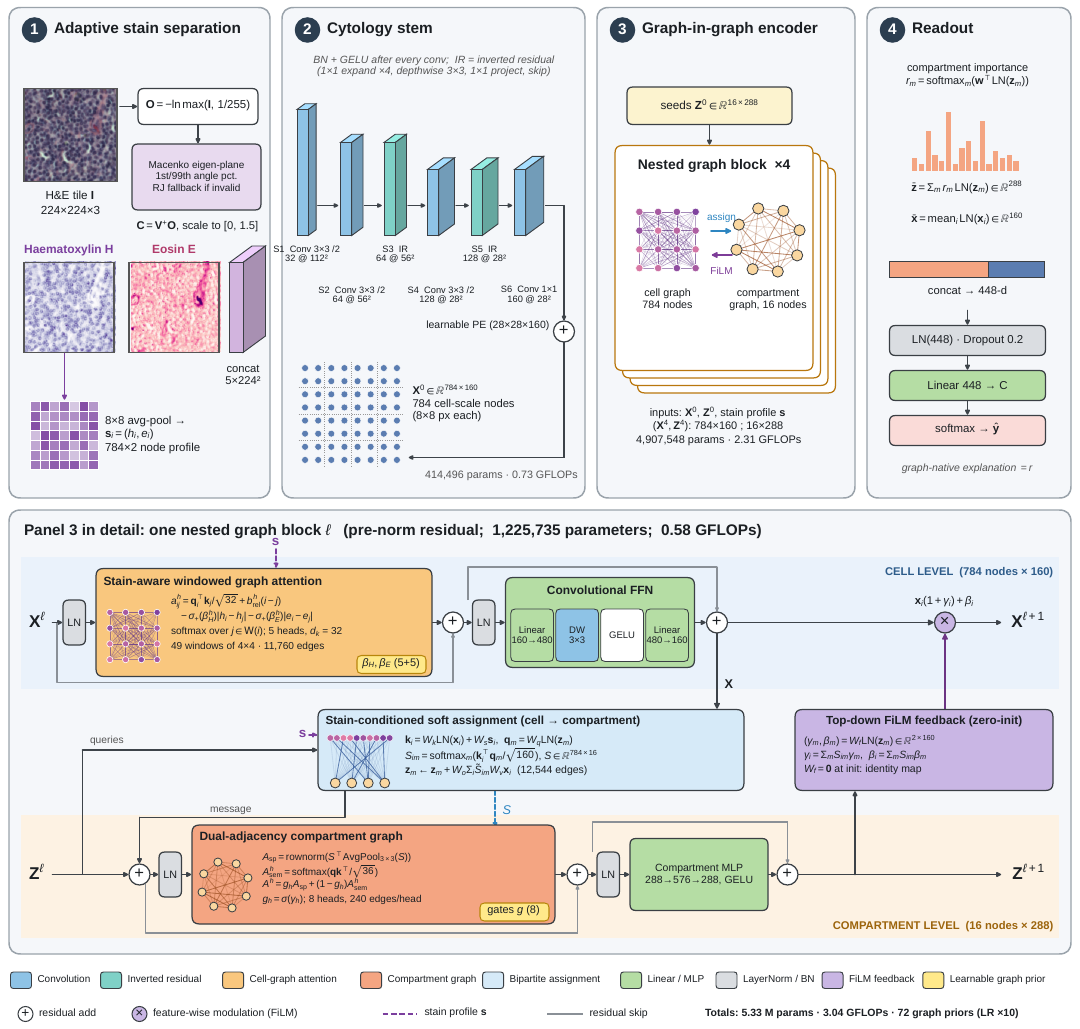}
\caption{End-to-end architecture of \model{}. \textbf{Panel 1:} per-image Macenko stain separation (Ruifrok--Johnston fallback) produces haematoxylin and eosin maps, a five-channel input and a $784\times2$ node stain profile $\mathbf{s}$ routed to every graph block. \textbf{Panel 2:} a cytology stem maps the tile to 784 patch nodes with 160-d features. \textbf{Panel 3:} four nested graph blocks update cell features and 16 compartment states jointly. \textbf{Panel 4:} the readout pools compartments by learned weight $r$, concatenates the mean cell feature and classifies. \textbf{Bottom:} one nested block in detail.}
\label{fig:architecture}
\end{figure}

\rev{This section first gives the overall workflow of the study (\cref{sec:m_workflow}), then formalises the learning problem (\cref{sec:m_problem}), describes the datasets, Source-Separated Image Partitioning (\ssip{}) and the leakage-control protocol (\cref{sec:m_data,sec:m_ssip,sec:m_leakage}), and presents \model{} stage by stage (\cref{sec:m_stain,sec:m_stem,sec:m_block,sec:m_readout}; \cref{fig:architecture}). \Cref{sec:m_complexity,sec:m_training,sec:m_eval,sec:m_impl} detail the complexity, training protocol, evaluation procedures, and implementation.}

\begin{figure}[!htbp]
\centering\revon
\includegraphics[width=\textwidth]{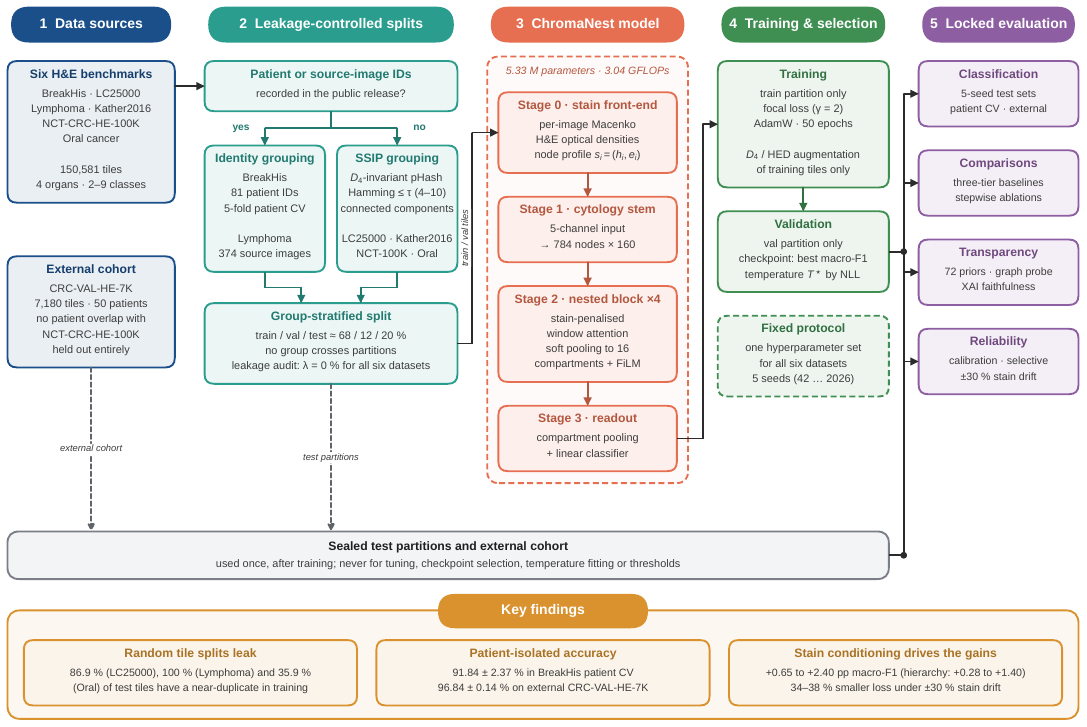}
\caption{Workflow of the study. \textbf{(1)}~Six public H\&E benchmarks and the held-out, patient-disjoint external cohort CRC-VAL-HE-7K. \textbf{(2)}~Leakage-controlled splits, drawn once before training: patient or source-image identifiers define the groups where the public release records them, and \ssip{} near-duplicate groups are used where it does not; whole groups are assigned to train, validation or test, and the leakage rate $\lambda$ is audited. \textbf{(3)}~The \model{} model (Stages~0--3). \textbf{(4)}~Training on the training partition, and checkpoint selection and temperature fitting on the validation partition, with one fixed configuration for all datasets (dashed green box: a constraint on every run, not a processing step). \textbf{(5)}~Locked evaluation: the selected models and the sealed test data feed all four analyses. Dashed arrows and the grey strip mark the sealed test partitions and external cohort, which are used once, only for final reporting. The bottom band lists the key findings.}
\label{fig:workflow}
\end{figure}

\begin{figure}[!htbp]
  \centering
  \includegraphics[width=\textwidth]{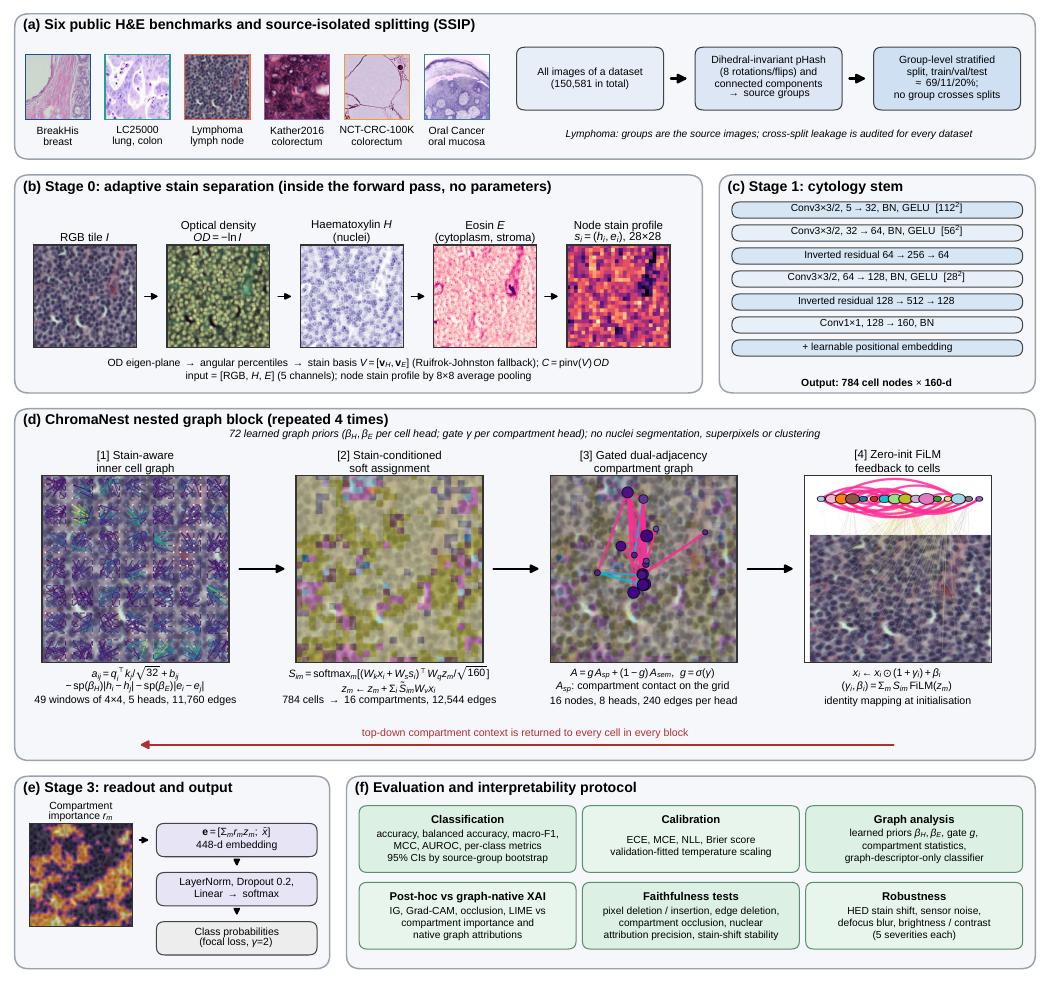}
  \caption{Overview of the \model{} framework. \textbf{(a)}~Six public H\&E benchmarks are partitioned by Source-Separated Image Partitioning (\ssip{}), which groups near-duplicate and known same-source images before splitting. \textbf{(b)}~Stage~0 separates haematoxylin and eosin by per-image Macenko unmixing and attaches a stain profile $\mathbf{s}_i=(h_i,e_i)$ to each cell node. \textbf{(c)}~Stage~1 maps the five-channel input to 784 cell nodes. \textbf{(d)}~Each of four nested graph blocks performs [1]~stain-penalised attention in 49 local $4\times4$ neighbourhoods, [2]~stain-conditioned soft assignment of cells to 16 learned compartments, [3]~message passing over a compartment graph that gates spatial contact and semantic adjacency, and [4]~zero-initialised FiLM feedback of compartment context to every cell. \textbf{(e)}~Stage~3 pools compartments and cell features for classification. \textbf{(f)}~Evaluation and structural-transparency audit protocol.}
  \label{fig:overview}
\end{figure}

%% ---------------------------------------------------------------------
\begingroup\revon
\subsection{Overall workflow}\label{sec:m_workflow}

\Cref{fig:workflow} summarises the workflow of the study in five stages, and \cref{fig:overview} shows the components of \model{} in more detail. \textbf{(1)~Data sources.} Six public H\&E benchmarks (150{,}581 tiles from four organs) are used for development and internal testing, and the CRC-VAL-HE-7K cohort (7{,}180 tiles from 50 patients, with no patient overlap with NCT-CRC-HE-100K~\cite{kather2019predicting}) is held out entirely as an external test set. \textbf{(2)~Leakage-controlled splits.} Before any model is trained, every dataset is split once into training, validation and test partitions at the level of the finest grouping that its metadata allow. Where patient or source-image identifiers are recorded, they define the groups (81 patient identifiers on BreakHis, used for 5-fold patient-grouped cross-validation; 374 source images on Lymphoma). Where they are not (LC25000, Kather2016, NCT-CRC-HE-100K and Oral Cancer), \ssip{} builds near-duplicate groups from a dihedral-invariant perceptual hash and its connected components. Whole groups are assigned to one partition, and the cross-split leakage rate $\lambda$ is audited for every dataset (\cref{sec:m_ssip,sec:m_leakage}). \textbf{(3)~Model.} Only training and validation tiles enter \model{}: a non-learned, per-image stain front-end (Stage~0, \cref{sec:m_stain}) that estimates no dataset-wide statistic, a cytology stem that produces 784 patch nodes (Stage~1), four stain-conditioned nested graph blocks (Stage~2) and a compartment readout (Stage~3). \textbf{(4)~Training and selection.} The model is trained on the training partition only, with augmentation applied on the fly to training tiles; the validation partition alone is used to select the checkpoint (best macro-F1) and to fit the calibration temperature; one hyperparameter configuration and five seeds are used for all datasets (\cref{sec:m_training}). \textbf{(5)~Locked evaluation.} The test partitions and the external cohort, sealed since stage~(2) (dashed paths and grey strip in \cref{fig:workflow}), are used once to report classification, baseline and ablation comparisons, structural transparency and explanation faithfulness, calibration and robustness (\cref{sec:m_eval}). The bottom band of \cref{fig:workflow} summarises the three principal findings.\par
\endgroup

%% ---------------------------------------------------------------------
\subsection{Problem formulation}\label{sec:m_problem}

Let $\mathcal{D}=\{(\mathbf{I}_n,y_n,\varsigma_n)\}_{n=1}^{N_{\mathcal{D}}}$ be a histopathology dataset, where $\mathbf{I}_n\in[0,1]^{224\times224\times3}$ is an H\&E tile, $y_n\in\mathcal{Y}=\{1,\dots,C\}$ is its diagnostic label and $\varsigma_n$ is the identifier of the source or near-duplicate group to which the tile belongs (\cref{sec:m_ssip}). We seek a classifier $f_\theta:[0,1]^{224\times224\times3}\to\Delta^{C-1}$ that minimises expected risk on tiles from unseen groups:
\begin{equation}
\theta^{\star}=\arg\min_{\theta}\;\frac{1}{|\mathcal{D}_{\mathrm{tr}}|}\sum_{(\mathbf{I},y)\in\mathcal{D}_{\mathrm{tr}}}\mathcal{L}\big(f_\theta(\mathbf{I}),y\big)
\quad\text{s.t.}\quad \mathcal{S}(\mathcal{D}_{\mathrm{tr}})\cap\mathcal{S}(\mathcal{D}_{\mathrm{va}})=\mathcal{S}(\mathcal{D}_{\mathrm{tr}})\cap\mathcal{S}(\mathcal{D}_{\mathrm{te}})=\varnothing,
\label{eq:objective}
\end{equation}
where $\mathcal{S}(\cdot)$ returns the set of group identifiers present in a split. Ideally, $\varsigma_n$ represents the patient or whole-slide image (as in our BreakHis patient-grouped cross-validation and our external CRC-VAL-HE-7K evaluation); when public datasets strip patient metadata and distribute augmented crops, $\varsigma_n$ is estimated via dihedral-invariant perceptual hashing (\cref{sec:m_ssip}) to prevent near-duplicate leakage across splits.

\model{} realises $f_\theta$ as a composition $f_\theta=\rho\circ\mathcal{B}_4\circ\cdots\circ\mathcal{B}_1\circ\phi\circ\psi$ of a non-learned stain operator $\psi$, a convolutional cytology stem $\phi$, four nested graph blocks $\mathcal{B}_\ell$ and a readout $\rho$. Its central object is a two-level nested graph built inside every tile,
\begin{equation}
\mathcal{G}(\mathbf{I})=\big(\mathcal{G}^{\mathrm{c}},\,\mathcal{G}^{\mathrm{k}},\,S\big),\qquad
\mathcal{G}^{\mathrm{c}}=(\mathcal{V}^{\mathrm{c}},\mathcal{E}^{\mathrm{c}}),\quad
\mathcal{G}^{\mathrm{k}}=(\mathcal{V}^{\mathrm{k}},\mathcal{E}^{\mathrm{k}}),\quad
S\in[0,1]^{N\times M},\;S\mathbf{1}_M=\mathbf{1}_N .
\label{eq:nested}
\end{equation}
The \emph{patch-node graph} $\mathcal{G}^{\mathrm{c}}$ has $N=784$ nodes arranged on a $28\times28$ grid, node features $\mathbf{X}\in\mathbb{R}^{N\times d}$ with $d=160$, and node stain profiles $\mathbf{s}_i=(h_i,e_i)$. Following windowed self-attention~\cite{liu2021swin}, its edges connect every ordered pair of distinct nodes inside each of the 49 non-overlapping $4\times4$ windows $\mathcal{W}_1,\dots,\mathcal{W}_{49}$:
\begin{equation}
\mathcal{E}^{\mathrm{c}}=\bigcup_{w=1}^{49}\{(i,j):i,j\in\mathcal{W}_w,\ i\neq j\},\qquad |\mathcal{E}^{\mathrm{c}}|=49\cdot16\cdot15=11{,}760 .
\end{equation}
The \emph{compartment graph} $\mathcal{G}^{\mathrm{k}}$ has $M=16$ latent nodes with states $\mathbf{Z}\in\mathbb{R}^{M\times d_z}$, $d_z=288$, and is complete without self-loops ($M(M-1)=240$ directed edges per head). Inspired by differentiable graph pooling~\cite{ying2018hierarchical,bianchi2020spectral} and slot attention~\cite{locatello2020object,jaegle2021perceiver}, the row-stochastic soft assignment matrix $S$ couples the two levels through $N\cdot M=12{,}544$ weighted bipartite edges. Each block maps $(\mathbf{X}^{\ell},\mathbf{Z}^{\ell})\mapsto(\mathbf{X}^{\ell+1},\mathbf{Z}^{\ell+1})$ given the fixed stain profile $\mathbf{s}$:
\begin{equation}
\mathcal{B}_\ell=\underbrace{\mathcal{F}_\ell}_{\text{top-down FiLM}}\circ\underbrace{\mathcal{M}_\ell\circ\mathcal{A}_\ell}_{\text{compartment graph}}\circ\underbrace{\Pi_\ell}_{\text{stain-cond.\ pool}}\circ\underbrace{\mathcal{C}_\ell}_{\text{stain-penalised window attn}},
\label{eq:block}
\end{equation}
where both $\mathcal{C}_\ell$ and $\Pi_\ell$ are explicitly conditioned on the physical H\&E stain profile $\mathbf{s}$. Principal notation is summarised in \cref{tab:notation}.

\begin{table}[!htbp]
\centering\footnotesize
\setlength{\tabcolsep}{4pt}
\caption{Principal notation.}\label{tab:notation}
\begin{tabularx}{\textwidth}{@{}l l X@{}}
\toprule
\textbf{Symbol} & \textbf{Shape} & \textbf{Meaning}\\
\midrule
$\mathbf{I}, \mathbf{O}$ & $224\times224\times3$ & RGB tile in $[0,1]$; optical density $\mathbf{O}=-\ln\max(\mathbf{I},1/255)$\\
$\mathbf{V}=[\mathbf{v}_H,\mathbf{v}_E]$ & $3\times2$ & per-image stain matrix (Macenko, or Ruifrok--Johnston fallback)\\
$\mathbf{H}, \mathbf{E}$ & $224\times224$ & haematoxylin and eosin concentration maps\\
$\mathbf{s}_i=(h_i,e_i)$ & $784\times2$ & stain profile of cell node $i$ ($8\times8$ average pooling of $\mathbf{H}, \mathbf{E}$)\\
$\mathbf{X}^{\ell}, \mathbf{x}_i$ & $784\times160$ & cell-node features entering block $\ell$; feature of node $i$\\
$\mathbf{Z}^{\ell}, \mathbf{z}_m$ & $16\times288$ & compartment states entering block $\ell$; $\mathbf{Z}^{0}$ are learnable seeds\\
$\mathcal{W}(i)$ & 16 & $4\times4$ window containing node $i$ (49 windows)\\
$\beta_H^{h}, \beta_E^{h}$ & $5$ each & learnable stain-affinity priors of inner head $h$; strength $\sigma_+(\beta)=\ln(1+e^{\beta})$\\
$S$ & $784\times16$ & soft cell-to-compartment assignment (row-stochastic)\\
$A_{\mathrm{sp}}, A^{h}_{\mathrm{sem}}, A^{h}$ & $16\times16$ & spatial-contact, semantic and gated compartment adjacency of outer head $h$\\
$g_h=\sigma(\gamma_h)$ & $8$ & learnable spatial--semantic gate of outer head $h$\\
$\boldsymbol\gamma_i, \boldsymbol\beta_i$ & $784\times160$ & FiLM scale and shift broadcast from compartments to cells\\
$r_m$ & $16$ & compartment readout weight\\
$C$, $\tau_{\mathcal{D}}$, $\lambda$ & -- & number of classes; \ssip{} Hamming threshold; pHash cross-split leakage rate\\
\bottomrule
\end{tabularx}
\end{table}

%% ---------------------------------------------------------------------
\subsection{Datasets}\label{sec:m_data}

\begin{figure}[!htbp]
\centering
\includegraphics[width=\textwidth]{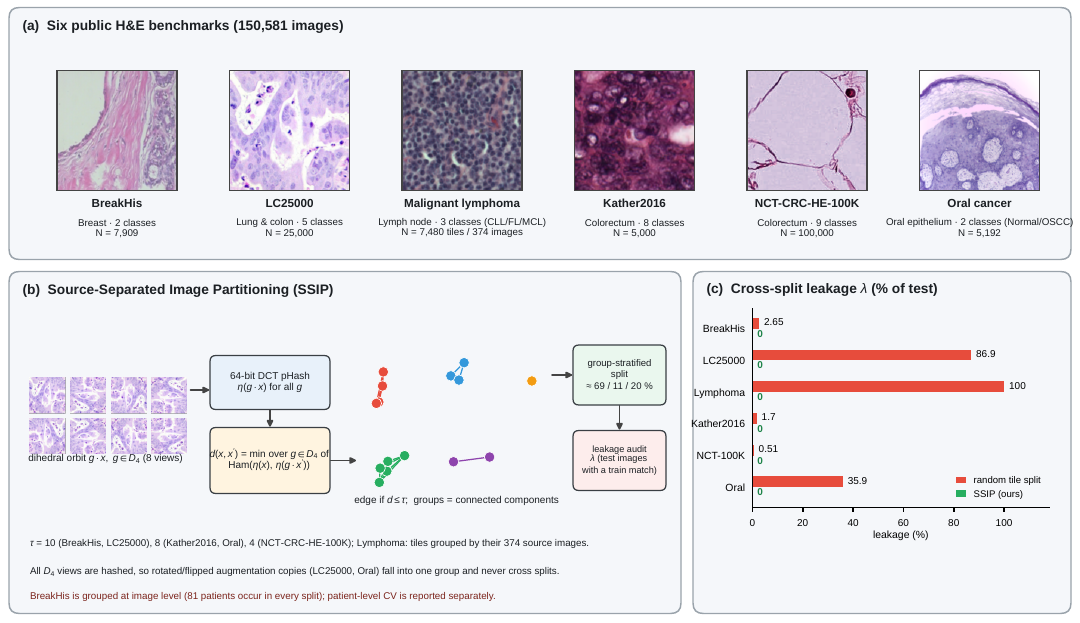}
\caption{Data and partitioning. \textbf{(a)} Representative tiles from the six benchmarks. \textbf{(b)} Source-Separated Image Partitioning (\ssip{}): every tile is hashed under all eight dihedral transformations, tiles whose minimum Hamming distance is at most $\tau_{\mathcal{D}}$ are linked, connected components define near-duplicate groups, and the split is stratified at group level. \textbf{(c)} Percentage of test images with a detected near-duplicate or same-source partner in training under a random tile-level split (red) and under \ssip{} (green).}
\label{fig:data_ssip}
\end{figure}

\model{} is evaluated across six public H\&E benchmarks covering four organs and 150{,}581 internal images (\cref{fig:data_ssip}a, \cref{tab:datasets}), plus the independent 7{,}180-image external colorectal cohort CRC-VAL-HE-7K (50 patients)~\cite{kather2019predicting}: BreakHis~\cite{spanhol2016dataset} (breast, 81 patient identifiers, benign versus malignant across 40$\times$--400$\times$), LC25000~\cite{borkowski2019lc25000} (five lung and colon classes), the malignant lymphoma dataset of the IICBU-2008 benchmark~\cite{shamir2008iicbu,orlov2010lymphoma} (CLL, FL, MCL; 7{,}480 tiles cropped from 374 source images), Kather2016~\cite{kather2016multiclass} (eight colorectal tissue types), NCT-CRC-HE-100K~\cite{kather2019predicting,kather2018nctzenodo} (nine colorectal tissue types) and an oral squamous cell carcinoma dataset (normal versus OSCC) derived from the histopathological imaging database of Rahman et al.~\cite{rahman2020oral}. All tiles are resized to $224\times224$ pixels.

As noted in \cref{sec:rw_tile}, LC25000, Lymphoma and Oral Cancer contain far more tiles than independent source images due to offline rotation, flipping and cropping prior to public release, whereas BreakHis provides explicit patient identifiers and NCT-CRC-HE-100K is paired with the patient-independent CRC-VAL-HE-7K validation cohort.

\paragraph{Dataset versions and usage terms.} For NCT-CRC-HE-100K we used the default release, in which all images are colour-normalised with Macenko's method; the unnormalised NCT-CRC-HE-100K-NONORM release was not used (see \cref{sec:d_limitations}). Kather2016 (Zenodo, \url{https://doi.org/10.5281/zenodo.53169}) and NCT-CRC-HE-100K/CRC-VAL-HE-7K (Zenodo, \url{https://doi.org/10.5281/zenodo.1214456}) are released under the Creative Commons Attribution 4.0 (CC BY 4.0) licence, as is the oral cancer database of Rahman et al.\ (Mendeley Data, \url{https://doi.org/10.17632/ftmp4cvtmb.2}). BreakHis is distributed by its authors for non-commercial research use with acknowledgement of the source and citation of~\cite{spanhol2016dataset} (the distribution page also states CC BY 4.0). LC25000 is distributed by its authors as de-identified images freely available to researchers, and the lymphoma images are part of the IICBU-2008 benchmark of the U.S. National Institute on Aging, distributed for free download; neither states a formal licence, and we cite both as their authors request.

\paragraph{BreakHis patient count.} The BreakHis authors report 82 patients~\cite{spanhol2016dataset}, but the image filenames, whose identifier field is described on the distribution page as a slide identifier, contain 81 unique values. We could not resolve this difference from the public metadata. We therefore group by the 81 identifiers; because this can only merge, never split, the images of one patient, no patient appears in both training and test folds.

\begin{table}[!htbp]
\caption{Datasets and partitioning. \textbf{(A)}~Characteristics of the six benchmark datasets. Imbalance ratio is the largest-to-smallest class count. \textbf{(B)}~\ssip{} configuration and near-duplicate leakage audit. Leakage $\lambda$ (\cref{eq:leakage}) is given for \ssip{} and, in parentheses, for a random tile-level split of the same data. For BreakHis, patient-isolated 5-fold CV over the 81 patients is reported in \cref{tab:cv}. Group-stratified splitting targets a nominal 68\%/12\%/20\% train/validation/test split (test 20\%, validation 15\% of the remainder; \cref{alg:ssip}); realised fractions differ slightly because whole groups are assigned (e.g., Oral Cancer: 3,584 Train, 561 Val, 1,047 Test).}\label{tab:datasets}\label{tab:ssip}
{\centering\small\textbf{(A)}\par}\smallskip
{\centering\footnotesize
\setlength{\tabcolsep}{3.5pt}

\begin{tabular}{@{}l l c r l r r c@{}}
\toprule
\textbf{Dataset} & \textbf{Organ / pathology} & \textbf{Classes} & \textbf{Images} & \textbf{Native size} & \textbf{Min class} & \textbf{Max class} & \textbf{Imbalance}\\
\midrule
BreakHis & Breast tumour (81 patients) & 2 (8 sub.) & 7{,}909 & $700\times460$ & 2{,}480 & 5{,}429 & 2.19:1\\
LC25000 & Lung \& colon & 5 & 25{,}000 & $768\times768$ & 5{,}000 & 5{,}000 & 1.00:1\\
Lymphoma & Lymph node (CLL/FL/MCL) & 3 & 7{,}480 & $1388\times1040$ (tiled) & 2{,}260 & 2{,}780 & 1.23:1\\
Kather2016 & Colorectal tissue & 8 & 5{,}000 & $150\times150$ & 625 & 625 & 1.00:1\\
NCT-CRC-HE-100K & Colorectal tissue (+7{,}180 ext.) & 9 & 100{,}000 & $224\times224$ & 8{,}763 & 14{,}317 & 1.63:1\\
Oral cancer & Oral epithelium (OSCC) & 2 & 5{,}192 & varies & 2{,}494 & 2{,}698 & 1.08:1\\
\midrule
\textbf{Total} & 4 organs & 2--9 & \textbf{150{,}581} & \multicolumn{4}{l}{all resized to $224\times224$ (+7{,}180 external CRC-VAL-HE-7K)}\\
\bottomrule
\end{tabular}\par}
\par\medskip{\centering\small\textbf{(B)}\par}\smallskip
{\centering\footnotesize
\setlength{\tabcolsep}{4pt}

\begin{tabular}{@{}l l r r r r c@{}}
\toprule
\textbf{Dataset} & \textbf{Grouping rule} & \textbf{Groups} & \textbf{Train} & \textbf{Val} & \textbf{Test} & \textbf{Detected leakage $\lambda$ (\%)}\\
\midrule
BreakHis & $D_4$-pHash CC, $\tau=10$ (81 patients) & 7{,}736 & 5{,}428 & 903 & 1{,}578 & 0.00 (2.65)\\
LC25000 & $D_4$-pHash CC, $\tau=10$ & 7{,}680 & 17{,}399 & 2{,}752 & 4{,}849 & 0.00 (86.9)\\
Lymphoma & known source image & 374 & 5{,}120 & 860 & 1{,}500 & 0.00 (100)\\
Kather2016 & $D_4$-pHash CC, $\tau=8$ & 4{,}926 & 3{,}446 & 568 & 986 & 0.00 (1.70)\\
NCT-CRC-HE-100K & $D_4$-pHash CC, $\tau=4$ & 99{,}466 & 68{,}875 & 11{,}286 & 19{,}839 & 0.00 (0.51)\\
Oral cancer & $D_4$-pHash CC, $\tau=8$ & 3{,}742 & 3{,}584 & 561 & 1{,}047 & 0.00 (35.9)\\
\bottomrule
\end{tabular}\par}
\end{table}

%% ---------------------------------------------------------------------
\subsection{Source-Separated Image Partitioning (\ssip{}) and patient-level isolation}\label{sec:m_ssip}

When slide or patient provenance is unavailable in public tile archives, random splitting places rotated or re-cropped copies of the same image in both training and test sets. \ssip{} constructs near-duplicate groups $\varsigma_n$ using a 64-bit DCT perceptual hash $\eta:\mathbf{I}\mapsto\{0,1\}^{64}$ evaluated across the dihedral group $D_4$ of eight square rotations and reflections:
\begin{equation}
d(\mathbf{I},\mathbf{I}')=\min_{g\in D_4}\ \mathrm{Ham}\big(\eta(\mathbf{I}),\,\eta(g\cdot\mathbf{I}')\big).
\label{eq:phash}
\end{equation}
We build the similarity graph $\mathcal{G}_{\mathcal{D}}=(\mathcal{D},\{(n,n'):d(\mathbf{I}_n,\mathbf{I}_{n'})\le\tau_{\mathcal{D}}\})$ and set $\varsigma_n$ to the connected-component index of tile $n$. Where source-image provenance is known (Lymphoma, 374 source images), $\varsigma_n$ is set directly to the source image ID. Groups are assigned to test (20\%), validation (15\% of remainder) and training splits via stratified group sampling (\cref{alg:ssip}, \cref{tab:ssip}). Near-duplicate cross-split leakage is audited via
\begin{equation}
\lambda=\frac{\big|\{\mathbf{I}\in\mathcal{D}_{\mathrm{te}}:\exists\,\mathbf{I}'\in\mathcal{D}_{\mathrm{tr}},\ d(\mathbf{I},\mathbf{I}')\le\tau_{\mathcal{D}}\ \text{or}\ \varsigma(\mathbf{I})=\varsigma(\mathbf{I}')\}\big|}{|\mathcal{D}_{\mathrm{te}}|}.
\label{eq:leakage}
\end{equation}

\begin{figure}[!htbp]
\centering
\begin{subfigure}[t]{0.495\textwidth}\centering\includegraphics[width=\linewidth]{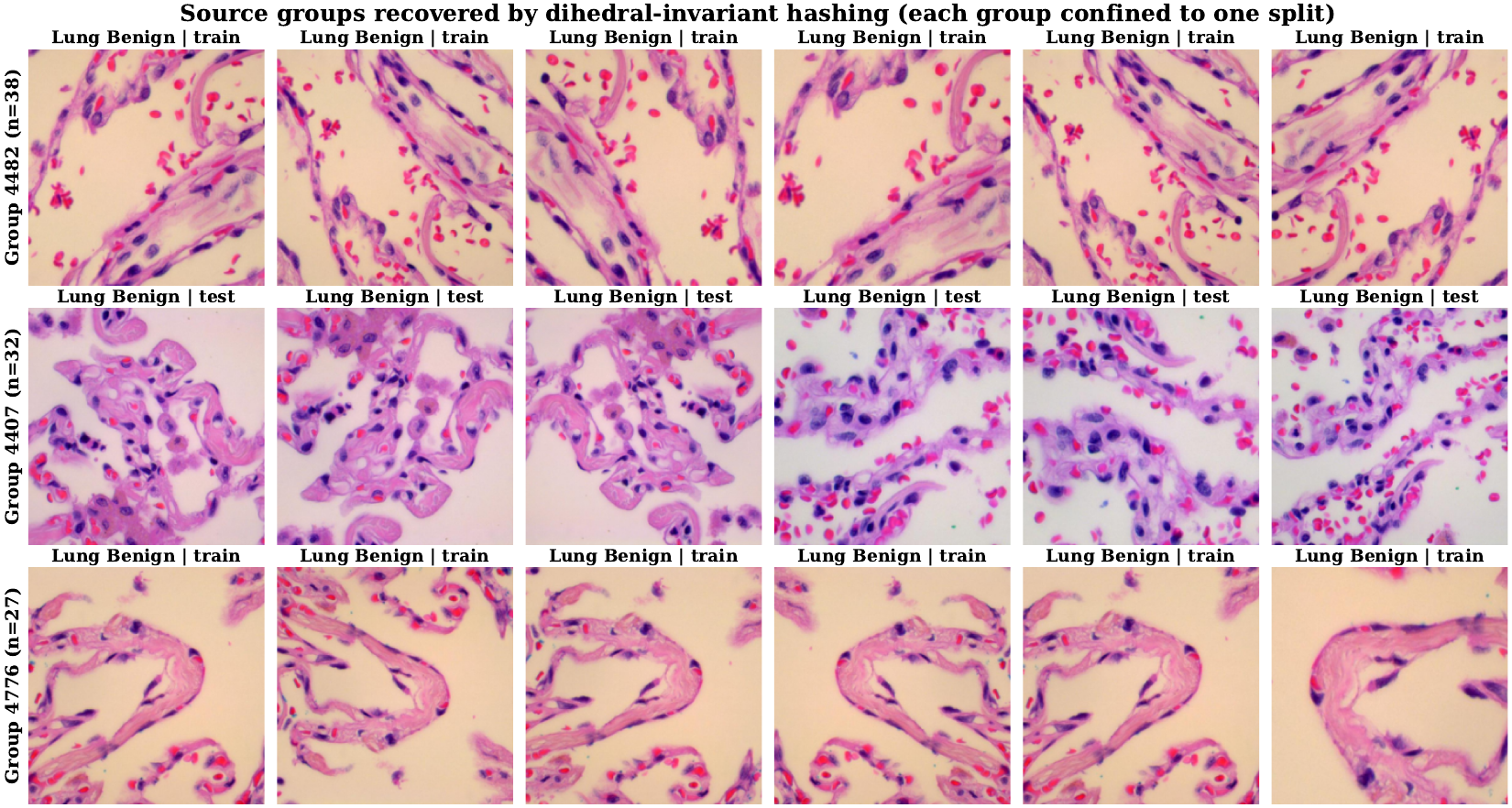}\caption{LC25000: recovered near-duplicate groups}\end{subfigure}\hfill
\begin{subfigure}[t]{0.495\textwidth}\centering\includegraphics[width=\linewidth]{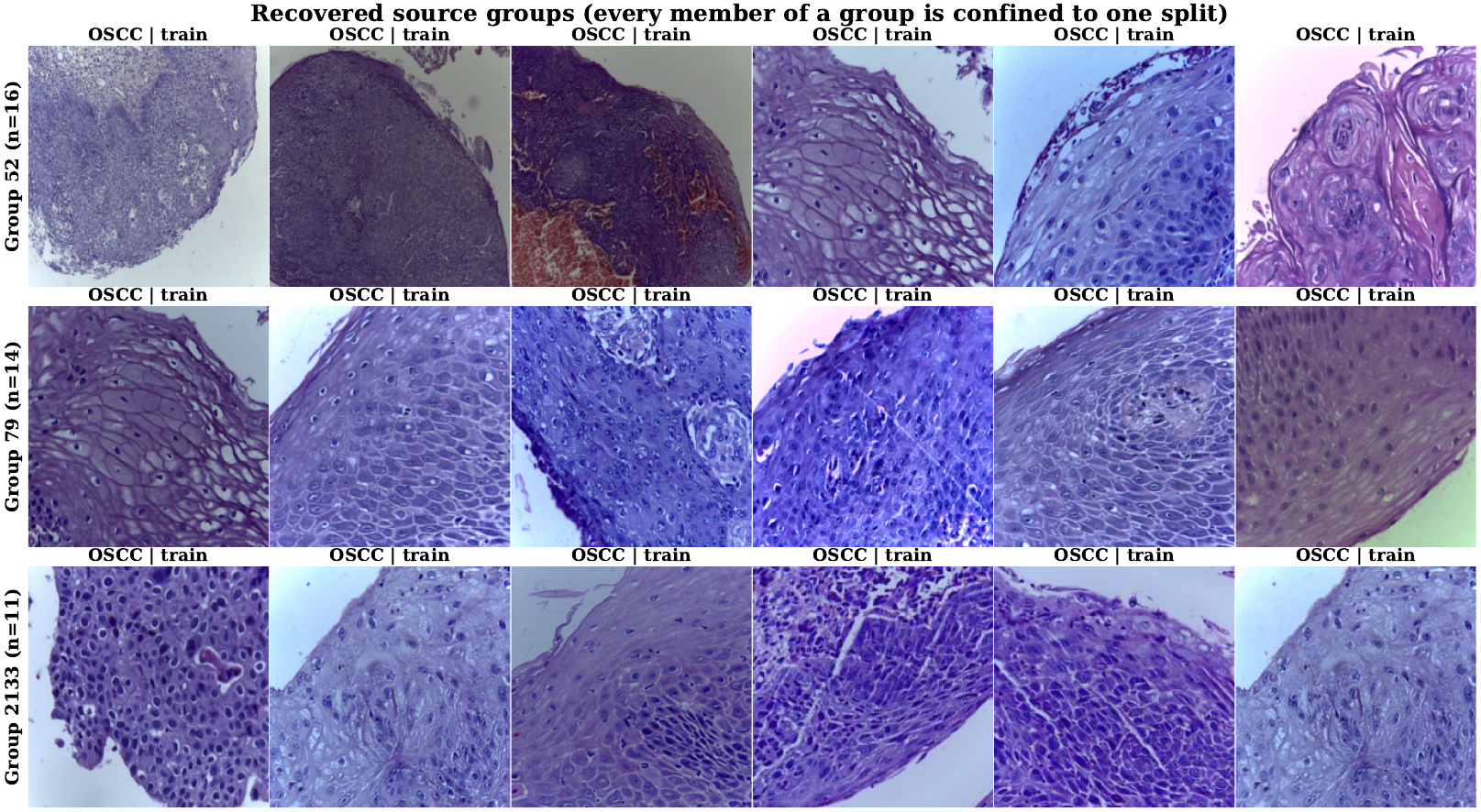}\caption{Oral cancer: recovered near-duplicate groups}\end{subfigure}
\caption{Near-duplicate groups recovered by \ssip{}. \textbf{(a,b)} Three groups per dataset; every member of a group is a rotated, flipped or re-cropped copy of the same original and is confined to a single split.}
\label{fig:groups}
\end{figure}

\paragraph{Scope and limitations of \ssip{}.} It is essential to distinguish near-duplicate isolation from true patient isolation. Because $\lambda$ in \cref{eq:leakage} uses the same grouping criterion as \cref{alg:ssip}, $\lambda=0$ under \ssip{} holds by construction and verifies partition consistency rather than complete source separation. Specifically, while $D_4$-pHash closes transitive chains of $90^\circ$ rotations, flips and overlapping crops (\cref{fig:groups}), it cannot merge arbitrary-angle rotations or non-overlapping tiles taken from the same patient. For example, on LC25000 (1{,}250 originals) \ssip{} recovers 7{,}680 groups. Under a random tile split, 86.9\% of LC25000 test tiles have a detected near-duplicate partner in training; this 86.9\% is the random-split leakage rate, not the share of all same-source matches that \ssip{} eliminates, and residual same-source similarity remains after \ssip{}. On BreakHis, we apply \ssip{} at image level to maintain a uniform protocol across all six benchmarks, but because BreakHis provides patient IDs (81 patients), we report \textbf{5-fold patient-grouped cross-validation} (\cref{sec:r_external}) as the primary clinical generalisation metric, alongside the \textbf{50-patient external CRC-VAL-HE-7K} test for NCT-CRC-HE-100K.

\begin{algorithm}[!htbp]
\caption{Source-Separated Image Partitioning (\ssip{})}\label{alg:ssip}
\small
\begin{algorithmic}[1]
\Require dataset $\{(\mathbf{I}_n,y_n)\}_{n=1}^{N_{\mathcal{D}}}$, threshold $\tau_{\mathcal{D}}$, optional source ids $\{\varsigma^{\mathrm{src}}_n\}$, split fractions $(p_{\mathrm{te}},p_{\mathrm{va}})=(0.20,0.15)$
\If{source ids are available} \State $\varsigma_n\gets\varsigma^{\mathrm{src}}_n$ for all $n$
\Else
  \For{$n=1$ to $N_{\mathcal{D}}$} \State $\mathcal{H}_n\gets\{\eta(g\cdot\mathbf{I}_n):g\in D_4\}$ \Comment{8 hashes per tile}\EndFor
  \State link $n\leftrightarrow n'$ if $\min_{a\in\mathcal{H}_n}\mathrm{Ham}(a,\eta(\mathbf{I}_{n'}))\le\tau_{\mathcal{D}}$ \Comment{\cref{eq:phash}}
  \State $\varsigma_n\gets$ connected-component index of $n$ (union--find)
\EndIf
\State $\mathcal{D}_{\mathrm{te}}\gets$ stratified group sample of $\{\varsigma_n\}$ covering fraction $p_{\mathrm{te}}$ of tiles
\State $\mathcal{D}_{\mathrm{va}}\gets$ stratified group sample of the remaining groups covering fraction $p_{\mathrm{va}}$; $\mathcal{D}_{\mathrm{tr}}\gets$ rest
\State compute $\lambda$ by \cref{eq:leakage}; \textbf{assert} $\lambda=0$
\State \Return $(\mathcal{D}_{\mathrm{tr}},\mathcal{D}_{\mathrm{va}},\mathcal{D}_{\mathrm{te}})$
\end{algorithmic}
\end{algorithm}

%% ---------------------------------------------------------------------
\begingroup\revon
\subsection{Leakage-control protocol and choice of split unit per dataset}\label{sec:m_leakage}

\paragraph{Principle.} Tiles from the same patient, specimen, slide or source image are correlated, so a split is only as independent as the unit it is drawn at. We therefore split every dataset at the highest level of the hierarchy patient $\succ$ specimen or slide $\succ$ source image $\succ$ near-duplicate group that its public metadata support, and we use \ssip{} only where no identifier above the tile level exists. \ssip{} is a substitute for, not an equivalent of, patient-independent validation, and we report and interpret it as such. \Cref{tab:splitunit} lists, for each dataset, the finest identifier that is available and the split unit that results.

\begin{table}[!htbp]
\centering\revon\footnotesize
\setlength{\tabcolsep}{3.5pt}
\caption{Split unit used for each dataset and the reason for it. Patient-level isolation is used wherever the public release contains patient identifiers; source-image grouping is used where only source images are known; \ssip{} near-duplicate grouping is used where neither is available.}\label{tab:splitunit}
\begin{tabularx}{\textwidth}{@{}l X p{3.4cm} X@{}}
\toprule
\textbf{Dataset} & \textbf{Identifier available in the public release} & \textbf{Split unit used} & \textbf{Status of the reported estimate}\\
\midrule
BreakHis & patient/slide identifier in every filename (81 values) & patients (5-fold CV); \ssip{} groups (secondary) & patient-independent (primary); image-level (secondary, for a uniform protocol)\\
Lymphoma & source image of every tile (374 images); no patient identifier & source images & source-image-independent; not patient-independent\\
NCT-CRC-HE-100K & no patient or slide identifier per tile & \ssip{} groups (internal); external cohort & internal: image-level; external: patient- and cohort-independent\\
CRC-VAL-HE-7K & 50 patients, disjoint from NCT-CRC-HE-100K~\cite{kather2019predicting} & whole cohort held out & external, never used for development\\
LC25000 & none (25{,}000 augmented tiles from 1{,}250 originals) & \ssip{} groups & near-duplicate-independent (image-level)\\
Kather2016 & none & \ssip{} groups & near-duplicate-independent (image-level)\\
Oral cancer & none (5{,}192 tiles from 978 originals~\cite{diniz2025oscc}) & \ssip{} groups & near-duplicate-independent (image-level)\\
\bottomrule
\end{tabularx}
\end{table}

\paragraph{Datasets with patient or source identifiers.} On BreakHis the 81 identifiers define 5-fold patient-grouped cross-validation in which no identifier occurs in both the training and the test fold. Because grouping by 81 identifiers can only merge, never split, the images of one of the 82 reported patients (\cref{sec:m_data}), this protocol is patient-isolated, and it is the estimate we treat as primary for BreakHis. On Lymphoma every tile is grouped with the other tiles cropped from the same source image, which closes the 100\% same-source leakage of a random tile split (\cref{tab:ssip}). For the colorectal task, the external CRC-VAL-HE-7K cohort comes from 50 patients who do not appear in NCT-CRC-HE-100K~\cite{kather2019predicting}, so it provides a patient- and cohort-independent estimate.

\paragraph{Datasets without identifiers.} LC25000, Kather2016, NCT-CRC-HE-100K and Oral Cancer are released without patient or slide identifiers, and three of them were produced by offline rotation, flipping and cropping of a much smaller set of originals. For these datasets \ssip{} reconstructs the near-duplicate structure from the pixels: tiles linked by a $D_4$-invariant perceptual-hash distance $\le\tau_{\mathcal{D}}$ are joined transitively into connected components (union--find), so a chain of near-duplicates $A\!\sim\!B\!\sim\!C$ is assigned to one group even when $A$ and $C$ are not directly similar. Grouping is computed on all images of a dataset before any partition is drawn, and whole groups are then assigned to one partition. This removes every cross-split pair that the hash can detect: under a random split, 86.9\% (LC25000), 100\% (Lymphoma) and 35.9\% (Oral Cancer) of test tiles have such a partner in training, against 0\% under the grouped splits (\cref{tab:ssip}).

\paragraph{Leakage pathways beyond the split.} A correct split can still be undermined later in the pipeline. \Cref{tab:leakctrl} lists each pathway and how it is closed in this study.

\begin{table}[!htbp]
\centering\revon\footnotesize
\setlength{\tabcolsep}{3.5pt}
\caption{Data-leakage pathways and the corresponding controls in this study.}\label{tab:leakctrl}
\begin{tabularx}{\textwidth}{@{}p{3.3cm} X p{2.3cm}@{}}
\toprule
\textbf{Leakage pathway} & \textbf{Control in \model{}} & \textbf{Where reported}\\
\midrule
Near-duplicate / same-source tiles across splits & identity groups or \ssip{} groups formed before splitting; whole groups per partition; $\lambda=0$ & \cref{tab:ssip}, \cref{alg:ssip}\\
Same patient across splits & patient-grouped 5-fold CV on BreakHis; patient-disjoint external cohort for colorectal tissue; not possible for the other datasets (no identifiers), stated as a limitation & \cref{tab:cv}, \cref{sec:d_limitations}\\
Preprocessing and normalisation & Stage~0 is estimated per image; the input scaling uses fixed constants; no mean, standard deviation, stain reference or other statistic is fitted on a dataset & \cref{sec:m_stain}\\
Augmentation & applied on the GPU to training mini-batches only, after partitioning; validation and test tiles are never augmented; the publishers' offline augmented copies are kept in one group by \ssip{} & \cref{sec:m_training}\\
Hyperparameter and model selection & one configuration, set on validation partitions and then fixed for all six datasets and all folds; checkpoint chosen by validation macro-F1; all five seeds and all ablation variants are reported, none is selected on test results & \cref{tab:training}, \cref{tab:perseed}\\
Calibration & temperature fitted by validation NLL; ECE, MCE, NLL and Brier score computed on the test partition & \cref{sec:m_eval}, \cref{tab:calib}\\
Decision threshold & predictions are the arg-max class; no threshold or class-specific rule is optimised & \cref{sec:m_eval}\\
External cohort & CRC-VAL-HE-7K is not used for training, hyperparameter choice, checkpoint selection or temperature fitting, and the model is applied without adaptation & \cref{tab:external}\\
Correlated test tiles in statistics & 95\% confidence intervals are bootstrapped over source groups, not tiles; patient-fold SD on BreakHis; source-image aggregation on Lymphoma & \cref{sec:r_overall}\\
Overlap between datasets & each dataset is trained and tested on its own partitions; the only cross-dataset evaluation (NCT-CRC-HE-100K $\to$ CRC-VAL-HE-7K) is patient-disjoint by construction & \cref{tab:external}\\
Stain or scanner confounding & cannot be removed by splitting; probed by the $\pm30\%$ stain-drift experiment and acknowledged for the Macenko-normalised NCT-CRC-HE-100K release & \cref{sec:r_robust}, \cref{sec:d_limitations}\\
\bottomrule
\end{tabularx}
\end{table}

\paragraph{What remains.} Perceptual hashing cannot join tiles that are rotated by arbitrary angles or that show non-overlapping regions of the same slide, so on the four datasets without identifiers the \ssip{} estimates are image-level and may still be optimistic. BreakHis gives a measure of the size of this effect: moving from image-level \ssip{} to patient-grouped folds lowers accuracy by about 3.9\,pp (\cref{sec:r_external}). We therefore report the patient-grouped and external-cohort results as the clinically relevant estimates and the \ssip{} results as leakage-audited benchmark numbers.\par
\endgroup

%% ---------------------------------------------------------------------
\subsection{Stage 0: non-learned per-image stain front-end}\label{sec:m_stain}

\begin{figure}[!htbp]
\centering
\begin{subfigure}[t]{\textwidth}\centering\includegraphics[width=0.92\linewidth]{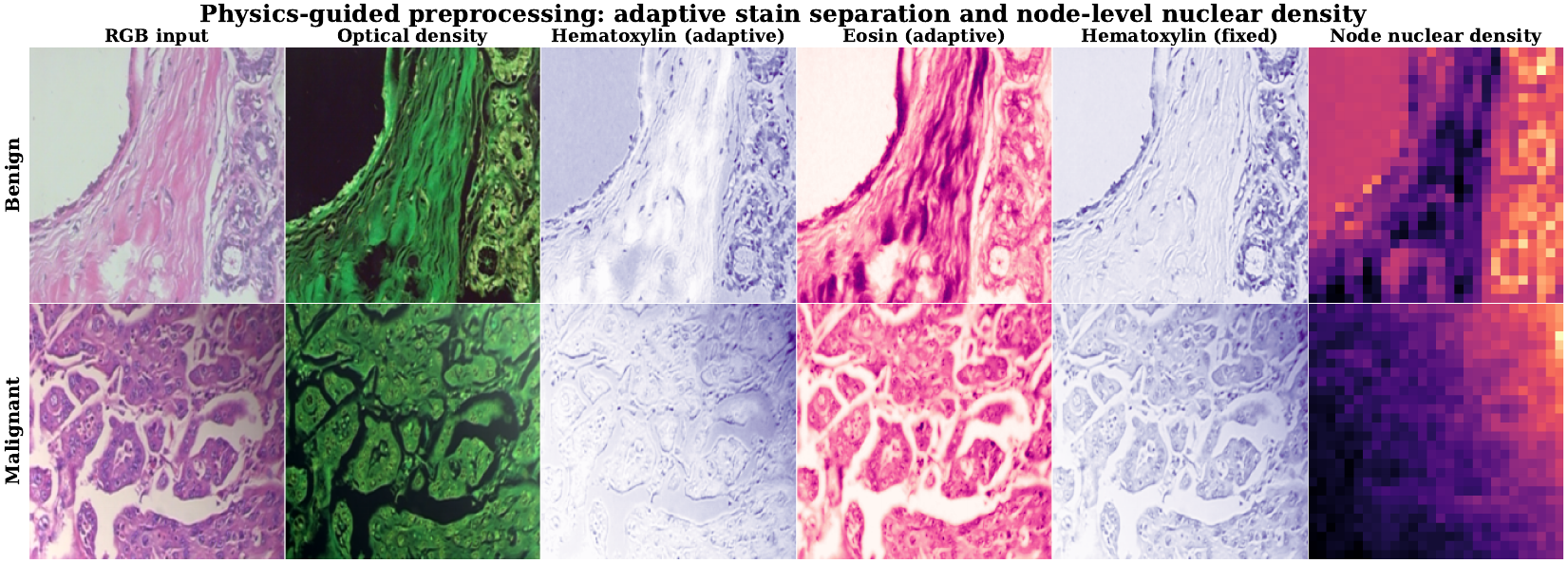}\caption{Stage 0 on BreakHis: RGB input, optical density, adaptive H\&E, fixed Ruifrok--Johnston H, and $8\times8$ node nuclear density $h_i$.}\end{subfigure}\\[3pt]
\begin{subfigure}[t]{\textwidth}\centering\includegraphics[width=\linewidth]{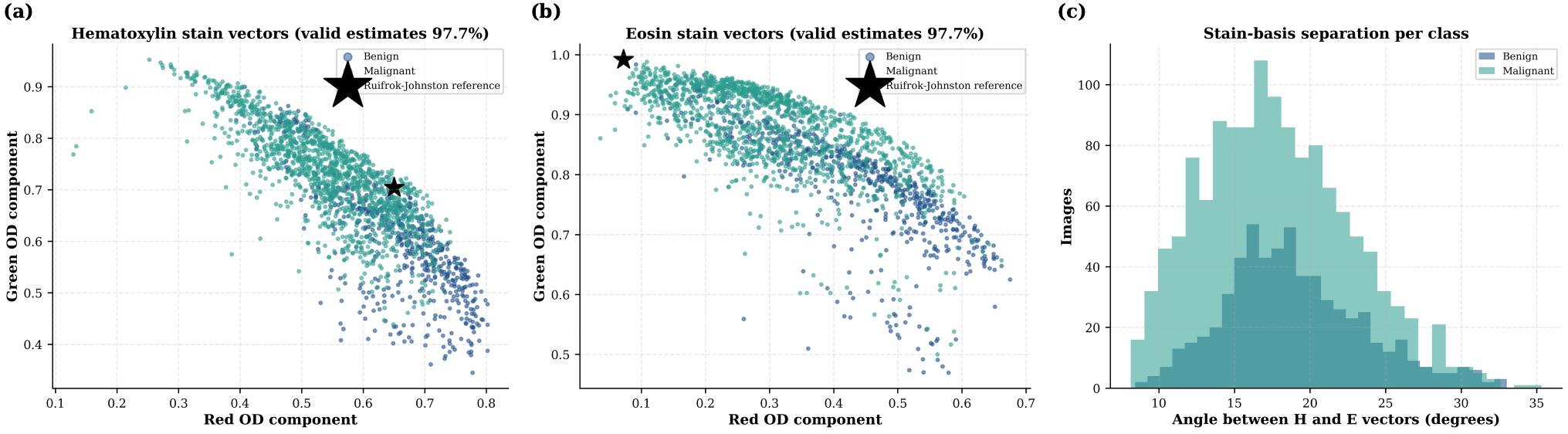}\caption{Per-image haematoxylin and eosin stain vectors estimated on BreakHis (stars: Ruifrok--Johnston reference) and inter-vector angle.}\end{subfigure}
\caption{Per-image stain front-end (Stage~0).}
\label{fig:stain}
\end{figure}

Under the Beer--Lambert law the optical density of a pixel is a non-negative linear combination of stain vectors~\cite{ruifrok2001quantification}:
\begin{equation}
\mathbf{O}=-\ln\max(\mathbf{I},1/255),\qquad \mathbf{O}(p)\approx\mathbf{V}\,\mathbf{c}(p),\quad \mathbf{V}=[\mathbf{v}_H,\mathbf{v}_E]\in\mathbb{R}^{3\times2},\ \mathbf{c}(p)\ge0 .
\label{eq:od}
\end{equation}
We estimate $\mathbf{V}$ per image via \citet{macenko2009method} on a $64\times64$ down-sampled copy using pixels with optical density $>0.15$ across all channels. The 1st and 99th angular percentiles in the plane spanned by the top two covariance eigenvectors define $\mathbf{v}_H$ and $\mathbf{v}_E$ (assigning the vector with the larger red component to haematoxylin). If fewer than 5\% of pixels are tissue, values are non-finite, or $\cos(\mathbf{v}_H,\mathbf{v}_E)\ge0.99$, Stage~0 falls back to the fixed Ruifrok--Johnston reference matrix, preventing numerical collapse on background-dominated tiles. Unmixed concentrations are scaled by the per-image 99th percentile $q_{99}$:
\begin{equation}
\mathbf{c}(p)=\max\big(\mathbf{0},\,(\mathbf{V}^{\top})^{+}\mathbf{O}(p)\big),\qquad
\tilde{\mathbf{H}}=\min\!\Big(\frac{\mathbf{H}}{q_{99}(\mathbf{H})},1.5\Big),\ \tilde{\mathbf{E}}=\min\!\Big(\frac{\mathbf{E}}{q_{99}(\mathbf{E})},1.5\Big).
\label{eq:unmix}
\end{equation}
Stage~0 outputs the five-channel input $[(\mathbf{I}-0.5)/0.5,\ 2\tilde{\mathbf{H}} - 1,\ 2\tilde{\mathbf{E}} - 1]\in\mathbb{R}^{5\times224\times224}$ and the $8\times8$ average-pooled node stain profile (\cref{fig:stain}):
\begin{equation}
\mathbf{s}_i=(h_i,e_i)=\frac{1}{64}\sum_{p\in\mathcal{P}_i}\big(\tilde{H}(p),\tilde{E}(p)\big),\qquad i=1,\dots,784 .
\label{eq:profile}
\end{equation}

%% ---------------------------------------------------------------------
\subsection{Stage 1: cytology stem}\label{sec:m_stem}

The stem $\phi$ (\cref{tab:stem}) maps the five-channel input to $N=784$ patch nodes ($28\times28$ grid, $d=160$) using three stride-2 $3\times3$ convolutions interleaved with two inverted-residual blocks, followed by a $1\times1$ projection and a learnable positional embedding $\mathbf{P}\in\mathbb{R}^{28\times28\times160}$:
\begin{equation}
\mathbf{X}^{0}=\operatorname{flatten}\big(\phi([\mathbf{I},\tilde{\mathbf{H}},\tilde{\mathbf{E}}])+\mathbf{P}\big)\in\mathbb{R}^{784\times160}.
\label{eq:stem}
\end{equation}
Each node summarises an $8\times8$ pixel micro-patch ($\approx 4\times 4\,\mu\text{m}$ at $20\times$ magnification on NCT-CRC-HE-100K), avoiding fragile offline nuclei segmentation~\cite{ahmedt2022survey,brussee2025graph}.

\begin{table}[!htbp]
\centering\footnotesize
\setlength{\tabcolsep}{3.5pt}
\caption{Cytology stem (Stage 1). All convolutions are bias-free and followed by BatchNorm ($\epsilon=10^{-5}$).}\label{tab:stem}
\begin{tabular}{@{}l l c c c l r@{}}
\toprule
\textbf{Layer} & \textbf{Type} & \textbf{Kernel / stride} & \textbf{Channels} & \textbf{Activation} & \textbf{Output} & \textbf{Params}\\
\midrule
S1 & Conv2d & $3\times3$ / 2 & $5\to32$ & GELU & $32\times112\times112$ & 1{,}504\\
S2 & Conv2d & $3\times3$ / 2 & $32\to64$ & GELU & $64\times56\times56$ & 18{,}560\\
S3 & Inverted residual & $1\times1$, DW $3\times3$, $1\times1$ & $64\to256\to64$ & GELU / linear & $64\times56\times56$ & 36{,}224\\
S4 & Conv2d & $3\times3$ / 2 & $64\to128$ & GELU & $128\times28\times28$ & 73{,}984\\
S5 & Inverted residual & $1\times1$, DW $3\times3$, $1\times1$ & $128\to512\to128$ & GELU / linear & $128\times28\times28$ & 137{,}984\\
S6 & Conv2d & $1\times1$ / 1 & $128\to160$ & linear & $160\times28\times28$ & 20{,}800\\
PE & Positional embedding & -- & 160 & -- & $28\times28\times160$ & 125{,}440\\
\midrule
\multicolumn{5}{@{}l}{\textbf{Total} (0.73 GFLOPs)} & \textbf{784 nodes $\times$ 160} & \textbf{414{,}496}\\
\bottomrule
\end{tabular}
\end{table}

%% ---------------------------------------------------------------------
\subsection{Stage 2: the nested graph block}\label{sec:m_block}

The encoder stacks four blocks (\cref{eq:block}, \cref{tab:block}) starting from $\mathbf{X}^{0}$ and learnable compartment seeds $\mathbf{Z}^{0}\in\mathbb{R}^{16\times288}\sim\mathcal{N}(0,1)$.

\subsubsection{Stain-penalised windowed cell-graph attention}\label{sec:m_inner}
Extending windowed attention with relative position bias~\cite{liu2021swin}, we penalise edges between nodes with dissimilar H\&E optical densities. With $\hat{\mathbf{x}}_i=\mathrm{LN}(\mathbf{x}_i)$ and projections $\mathbf{q}^h_i,\mathbf{k}^h_i,\mathbf{v}^h_i\in\mathbb{R}^{32}$ across $H_c=5$ heads, the attention logit for $j\in\mathcal{W}(i)$ is
\begin{equation}
a^{h}_{ij}=\frac{\mathbf{q}^{h\top}_i\mathbf{k}^{h}_j}{\sqrt{32}}+b^{h}_{\mathrm{rel}}(\Delta_{ij})-\sigma_+(\beta^{h}_H)\,|h_i-h_j|-\sigma_+(\beta^{h}_E)\,|e_i-e_j|,\qquad j\in\mathcal{W}(i),
\label{eq:attn}
\end{equation}
where $\sigma_+(\cdot)=\ln(1+e^{\cdot})$ enforces non-negativity and $\beta^h_H,\beta^h_E$ are initialised at $\ln(e-1)$ ($\sigma_+=1$). Messages are aggregated within each $4\times4$ window:
\begin{equation}
\alpha^{h}_{ij}=\frac{\exp a^{h}_{ij}}{\sum_{j'\in\mathcal{W}(i)}\exp a^{h}_{ij'}},\qquad
\mathbf{x}_i\leftarrow\mathbf{x}_i+\mathrm{DP}\Big(W_o\,\big\Vert_{h=1}^{5}\textstyle\sum_{j\in\mathcal{W}(i)}\alpha^{h}_{ij}\mathbf{v}^{h}_j\Big),
\label{eq:attn_msg}
\end{equation}
followed by a convolutional FFN ($\mathbf{x}\leftarrow\mathbf{x}+\mathrm{DP}(W_2\,\mathrm{GELU}(\mathrm{DWConv}_{3\times3}(W_1\mathrm{LN}(\mathbf{x}))))$, $160\to480\to160$) that provides cross-window spatial continuity.

\subsubsection{Stain-conditioned soft compartment assignment}\label{sec:m_assign}
Adapting differentiable soft pooling~\cite{ying2018hierarchical,locatello2020object,jaegle2021perceiver} to histology, we condition the cell keys on the node stain profile $\mathbf{s}_i=(h_i,e_i)$ so that nuclear-rich and eosin-rich nodes are guided toward distinct compartment slots:
\begin{equation}
\mathbf{k}_i=W_k\hat{\mathbf{x}}_i+W_s\mathbf{s}_i,\qquad \mathbf{q}_m=W_q\hat{\mathbf{z}}_m,\qquad
S_{im}=\frac{\exp(\mathbf{k}_i^{\top}\mathbf{q}_m/\sqrt{160})}{\sum_{m'=1}^{16}\exp(\mathbf{k}_i^{\top}\mathbf{q}_{m'}/\sqrt{160})},
\label{eq:assign}
\end{equation}
\begin{equation}
\tilde{S}_{im}=\frac{S_{im}}{\sum_{i'}S_{i'm}},\qquad
\mathbf{z}_m\leftarrow\mathbf{z}_m+\mathrm{DP}\Big(W_o'\sum_{i=1}^{784}\tilde{S}_{im}\,W_v\hat{\mathbf{x}}_i\Big).
\label{eq:pool}
\end{equation}

\subsubsection{Dual-adjacency compartment graph and compartment MLP}\label{sec:m_outer}
Compartments interact through a non-learned spatial-contact adjacency $A_{\mathrm{sp}}$ (measuring how often two compartments border one another on the $28\times28$ cell grid) and a multi-head semantic attention adjacency $A^{h}_{\mathrm{sem}}=\mathrm{softmax}(\mathbf{Q}^{h}\mathbf{K}^{h\top}/\sqrt{36})$ ($H_k=8$ heads):
\begin{equation}
A_{\mathrm{sp}}=\operatorname{rownorm}\big(S^{\top}\,\mathrm{AvgPool}_{3\times3}(S)\big)\in\mathbb{R}^{16\times16},
\label{eq:asp}
\end{equation}
\begin{equation}
A^{h}=g_h\,A_{\mathrm{sp}}+(1-g_h)\,A^{h}_{\mathrm{sem}},\qquad g_h=\sigma(\gamma_h),\qquad
\mathbf{Z}\leftarrow\mathbf{Z}+\mathrm{DP}\Big(W_p\,\big\Vert_{h=1}^{8}A^{h}\mathbf{V}^{h}\Big),
\label{eq:dual}
\end{equation}
followed by a two-layer compartment MLP ($288\to576\to288$).

\subsubsection{Top-down compartment-to-cell FiLM feedback}\label{sec:m_film}
Compartment context is broadcast back to individual cells via zero-initialised feature-wise linear modulation (FiLM)~\cite{perez2018film}:
\begin{equation}
(\boldsymbol\gamma_m,\boldsymbol\beta_m)=W_f\,\mathrm{LN}(\mathbf{z}_m)\in\mathbb{R}^{2\times160},\qquad
\boldsymbol\gamma_i=\sum_{m}S_{im}\boldsymbol\gamma_m,\ \ \boldsymbol\beta_i=\sum_{m}S_{im}\boldsymbol\beta_m,\qquad
\mathbf{x}_i\leftarrow\mathbf{x}_i\odot(1+\boldsymbol\gamma_i)+\boldsymbol\beta_i ,
\label{eq:film}
\end{equation}
where $W_f=0$ at initialisation ensures an identity top-down path at the start of training (\cref{alg:forward}).

\begin{table}[!htbp]
\centering\footnotesize
\setlength{\tabcolsep}{3pt}
\caption{Operators of one nested graph block (repeated 4 times; 1{,}225{,}735 parameters and 0.577 GFLOPs per block).}\label{tab:block}
\begin{tabularx}{\textwidth}{@{}l X l r r@{}}
\toprule
\textbf{Operator} & \textbf{Configuration} & \textbf{Output shape} & \textbf{Params} & \textbf{GFLOPs}\\
\midrule
Stain-aware attention ($\mathcal{C}_\ell$) & 49 windows of $4\times4$; 5 heads, $d_k=32$; rel.\ bias $49\times5$; $\beta_H,\beta_E\in\mathbb{R}^5$ & $784\times160$ & 103{,}615 & 0.169\\
Convolutional FFN ($\mathcal{C}_\ell$) & Linear $160\to480$, DW $3\times3$, GELU, Linear $480\to160$ & $784\times160$ & 159{,}360 & 0.248\\
Soft assignment ($\Pi_\ell$) & keys $160\to160$ + stain $2\to160$; queries $288\to160$; values $160\to288$ & $16\times288$ & 202{,}976 & 0.128\\
Dual-adjacency graph ($\mathcal{A}_\ell$) & $A_{\mathrm{sp}}$ from $S$; QKV $288\to864$; 8 heads, $d_k=36$; gates $g\in\mathbb{R}^8$ & $16\times288$ & 333{,}512 & 0.011\\
Compartment MLP ($\mathcal{M}_\ell$) & Linear $288\to576$, GELU, Linear $576\to288$ & $16\times288$ & 333{,}216 & 0.011\\
FiLM feedback ($\mathcal{F}_\ell$) & Linear $288\to320$ (zero-init); broadcast via $S$ & $784\times160$ & 93{,}056 & 0.011\\
\midrule
\textbf{Block total} & 18 learnable graph priors ($5+5$ stain, 8 gates) & -- & \textbf{1{,}225{,}735} & \textbf{0.577}\\
\bottomrule
\end{tabularx}
\end{table}

\begin{algorithm}[!htbp]
\caption{\model{} forward pass for one tile}\label{alg:forward}
\small
\begin{algorithmic}[1]
\Require tile $\mathbf{I}\in[0,1]^{224\times224\times3}$; parameters $\theta$
\State $\mathbf{O}\gets-\ln\max(\mathbf{I},1/255)$; estimate $\mathbf{V}$ (Macenko, RJ fallback); $(\tilde{\mathbf{H}},\tilde{\mathbf{E}})$ by \cref{eq:unmix} \Comment{no gradient}
\State $\mathbf{s}\gets$ $8\times8$ average pooling of $(\tilde{\mathbf{H}},\tilde{\mathbf{E}})$ \Comment{\cref{eq:profile}, $784\times2$}
\State $\mathbf{X}\gets\phi([\mathbf{I},\tilde{\mathbf{H}},\tilde{\mathbf{E}}])+\mathbf{P}$;\quad $\mathbf{Z}\gets\mathbf{Z}^{0}$ \Comment{\cref{eq:stem}}
\For{$\ell=1$ to $4$}
  \State $\mathbf{X}\gets\mathbf{X}+\mathrm{DP}(\textsc{StainAttn}(\mathrm{LN}(\mathbf{X}),\mathbf{s}))$ \Comment{\cref{eq:attn,eq:attn_msg}}
  \State $\mathbf{X}\gets\mathbf{X}+\mathrm{DP}(\textsc{ConvFFN}(\mathrm{LN}(\mathbf{X})))$
  \State $S\gets\textsc{Assign}(\mathbf{X},\mathbf{Z},\mathbf{s})$;\quad $\mathbf{Z}\gets\mathbf{Z}+\mathrm{DP}(\textsc{Pool}(S,\mathbf{X}))$ \Comment{\cref{eq:assign,eq:pool}}
  \State $\mathbf{Z}\gets\mathbf{Z}+\mathrm{DP}(\textsc{DualGraph}(\mathrm{LN}(\mathbf{Z}),S))$;\quad $\mathbf{Z}\gets\mathbf{Z}+\mathrm{DP}(\textsc{MLP}(\mathrm{LN}(\mathbf{Z})))$ \Comment{\cref{eq:asp,eq:dual}}
  \State $\mathbf{X}\gets\mathbf{X}\odot(1+S\boldsymbol\Gamma)+S\mathbf{B}$, with $[\boldsymbol\Gamma,\mathbf{B}]=W_f\mathrm{LN}(\mathbf{Z})$ \Comment{\cref{eq:film}}
\EndFor
\State $r\gets\mathrm{softmax}(\mathbf{w}^{\top}\mathrm{LN}(\mathbf{Z}))$;\quad $\mathbf{u}\gets\mathrm{LN}\Big(\Big[\sum_m r_m\mathrm{LN}(\mathbf{z}_m)\,;\,\frac{1}{784}\sum_{i=1}^{784}\mathrm{LN}(\mathbf{x}_i)\Big]\Big)$ \Comment{\cref{eq:readout}}
\State \Return $\hat{\mathbf{y}}=\mathrm{softmax}(W_c\,\mathrm{Dropout}(\mathbf{u})+\mathbf{b}_c)$, readout weights $r$ and graphs $(S,\{A^{h}\},\{\alpha^{h}\})$
\end{algorithmic}
\end{algorithm}

%% ---------------------------------------------------------------------
\subsection{Stage 3: readout and classifier}\label{sec:m_readout}

The readout pools the final compartment states via learned weights $r_m$ and concatenates the mean cell feature:
\begin{equation}
r_m=\frac{\exp(\mathbf{w}^{\top}\mathrm{LN}(\mathbf{z}_m)+b)}{\sum_{m'}\exp(\mathbf{w}^{\top}\mathrm{LN}(\mathbf{z}_{m'})+b)},\qquad
\mathbf{u}=\mathrm{LN}\Big(\Big[\textstyle\sum_{m}r_m\,\mathrm{LN}(\mathbf{z}_m)\,;\ \frac{1}{784}\sum_{i=1}^{784}\mathrm{LN}(\mathbf{x}_i)\Big]\Big)\in\mathbb{R}^{448},\qquad
\hat{\mathbf{y}}=\mathrm{softmax}(W_c\mathbf{u}+\mathbf{b}_c).
\label{eq:readout}
\end{equation}
Note that $\mathbf{u}$ includes a direct pathway from the mean patch representation $\frac{1}{784}\sum_{i=1}^{784}\mathrm{LN}(\mathbf{x}_i)$, which has already received compartment feedback in blocks 1--4, so that compartment context and patch-level evidence jointly inform the decision.

%% ---------------------------------------------------------------------
\subsection{Graph specification and computational complexity}\label{sec:m_complexity}

\Cref{tab:graphspec,tab:complexity} detail the nested graph and parameter budget (5.33\,M parameters, 3.04\,GFLOPs). Half of the parameters (50.1\%) reside in the compartment operators ($M=16$), costing only 0.09\,GFLOPs, while the 72 scalar graph priors account for 0.001\% of the parameter budget. \Cref{tab:lightweight} compares model size and latency across architectures, and \cref{tab:design} summarises the architectural differences relative to prior graph and pooling models.

\begin{table}[!htbp]
\centering\footnotesize
\setlength{\tabcolsep}{4pt}
\caption{Specification of the learned nested graph.}\label{tab:graphspec}
\begin{tabularx}{\textwidth}{@{}l l X@{}}
\toprule
\textbf{Level} & \textbf{Property} & \textbf{Specification}\\
\midrule
Cell graph $\mathcal{G}^{\mathrm{c}}$ & Nodes / features & 784 patch nodes ($28\times28$, each $8\times8$ px); 160-d embedding and 2-d stain profile $(h,e)$\\
 & Edges & 49 fully connected $4\times4$ windows: 11{,}760 directed edges ($+784$ self-loops), 5 heads\\
 & Edge weight & content $\mathbf{q}^{\top}\mathbf{k}$ + relative position~\cite{liu2021swin} + learned H\&E affinity penalties (\cref{eq:attn})\\
Assignment $S$ & Edges / weight & $784\times16=12{,}544$ soft memberships per block; stain-conditioned key--query softmax (\cref{eq:assign})\\
Compartment graph $\mathcal{G}^{\mathrm{k}}$ & Nodes / edges & 16 learned compartments (288-d); complete graph, 240 directed edges per head, 8 heads\\
 & Adjacency & gated mixture of spatial contact frequency $A_{\mathrm{sp}}$ and semantic attention $A_{\mathrm{sem}}$ (\cref{eq:asp,eq:dual})\\
Graph priors & Count & 72 inspectable scalars: $4\times(5\,\beta_H+5\,\beta_E+8\,\gamma)$, learning rate $\times10$\\
\bottomrule
\end{tabularx}
\end{table}

\begin{table}[!htbp]
\centering\footnotesize
\setlength{\tabcolsep}{3.5pt}
\caption{Component-wise parameter and computation budget of \model{} (BreakHis, $C=2$).}\label{tab:complexity}
\begin{tabular}{@{}l l r r r@{}}
\toprule
\textbf{Stage / component} & \textbf{Configuration} & \textbf{Parameters} & \textbf{Share (\%)} & \textbf{GFLOPs}\\
\midrule
0: Adaptive stain separation & per-image Macenko, RJ fallback & 0 & 0.00 & 0.0007\\
1: Cytology stem + PE & $5\to160$ channels, $224\to28$ & 414{,}496 & 7.78 & 0.7322\\
2a: Stain-aware cell attention ($\times4$) & $4\times4$ windows, 5 heads & 414{,}460 & 7.78 & 0.6744\\
2b: Convolutional FFN ($\times4$) & $160\to480\to160$, DW $3\times3$ & 637{,}440 & 11.97 & 0.9905\\
2c: Soft assignment ($\times4$) + seeds & $784\to16$, stain-conditioned & 816{,}512 & 15.33 & 0.5131\\
2d: Dual-adjacency graph ($\times4$) & 16 nodes, 8 heads, gate $g$ & 1{,}334{,}048 & 25.05 & 0.0453\\
2e: Compartment MLP ($\times4$) & $288\to576\to288$ & 1{,}332{,}864 & 25.03 & 0.0425\\
2f: FiLM feedback ($\times4$) & zero-init, membership-weighted & 372{,}224 & 6.99 & 0.0439\\
3: Readout and classifier & importance pooling, LN(448), Linear & 2{,}979 & 0.06 & 0.0000\\
\midrule
\textbf{Total} & 4 nested blocks & \textbf{5{,}325{,}023} & \textbf{100.00} & \textbf{3.0424}\\
\bottomrule
\end{tabular}
\end{table}

\begin{table}[!htbp]
\centering\footnotesize
\setlength{\tabcolsep}{4pt}
\caption{Model size and cost for a $224\times224$ tile at batch size 1 on a Tesla T4.}\label{tab:lightweight}
\begin{tabular}{@{}l l r r r@{}}
\toprule
\textbf{Architecture} & \textbf{Type} & \textbf{Params (M)} & \textbf{GFLOPs} & \textbf{T4 latency (ms/tile)}\\
\midrule
EfficientNet-B0 & CNN & 5.29 & 0.39 & 12.10\\
MobileNetV3-Large & CNN & 5.48 & 0.22 & 10.40\\
ResNet-50 & CNN & 23.52 & 4.12 & 16.40\\
DenseNet-121 & CNN & 6.96 & 2.88 & 20.85\\
ConvNeXt-Tiny & CNN & 27.82 & 4.46 & 17.35\\
EfficientNet-B2 & CNN & 7.70 & 1.05 & 18.50\\
ViT-Base/16 & Transformer & 85.80 & 17.58 & 38.60\\
Swin-Tiny~\cite{liu2021swin} & Windowed ViT & 27.52 & 4.51 & 23.20\\
DeiT-Small & Transformer & 21.67 & 4.61 & 19.95\\
ViG-Tiny~\cite{han2022vision} & Vision GNN & 7.15 & 1.32 & 24.40\\
DeepGCN-Res & Graph & 12.40 & 3.85 & 32.10\\
Cell-Graph GNN & Pathology graph & 8.90 & 3.40 & 48.25\\
CTransPath~\cite{wang2022ctranspath} & Pathology hybrid ViT & 27.50 & 4.51 & 23.45\\
UNI~\cite{chen2024uni} / CONCH~\cite{lu2024conch} / Virchow~\cite{vorontsov2024virchow} & Foundation & $\approx$300 / 90 / 632 & -- & --\\
\midrule
\textbf{\model{} (ours)} & Stain-aware nested graph & 5.33 & 3.04 & 14.92\\
\bottomrule
\end{tabular}
\end{table}

\begin{table}[!htbp]
\centering\footnotesize
\setlength{\tabcolsep}{3pt}
\caption{Design comparison of \model{} with differentiable pooling, graph-in-graph and hierarchical models.}\label{tab:design}
\begin{tabular}{@{}l c c c c c c c@{}}
\toprule
\textbf{Property} & \textbf{DiffPool}~\cite{ying2018hierarchical} & \textbf{GiG}~\cite{zaripova2023gig} & \textbf{GiGCN}~\cite{jia2024gigcn} & \textbf{HACT-Net}~\cite{pati2022hierarchical} & \textbf{HIGT}~\cite{guo2023higt} & \textbf{ViHGNN}~\cite{han2023vihgnn} & \textbf{\model{}}\\
\midrule
Domain & graphs & bio-graphs & hyperspectral & histology & WSI & natural images & histology tiles\\
Inner nodes learned from pixels & \xmark & \xmark & \xmark & \xmark & \xmark & \cmark & \cmark\\
Outer graph within one image & \cmark & \xmark & \cmark & \cmark & \cmark & -- & \cmark\\
Differentiable soft partition & \cmark & -- & \xmark & \xmark & \xmark & $\sim$ & \cmark\\
Segmentation-free & \cmark & \cmark & \xmark & \xmark & \cmark & \cmark & \cmark\\
Physical (H\&E stain) edge \& pool prior & \xmark & \xmark & \xmark & \xmark & \xmark & \xmark & \cmark\\
Top-down feedback to inner nodes & \xmark & \xmark & \xmark & \xmark & \cmark & -- & \cmark\\
Gated spatial-contact / semantic outer graph & \xmark & \xmark & \xmark & \xmark & \xmark & \xmark & \cmark\\
\bottomrule
\end{tabular}
\end{table}

%% ---------------------------------------------------------------------
\subsection{Training protocol}\label{sec:m_training}

The model is trained for $T=50$ epochs with class-weighted focal loss ($\gamma=2$)~\cite{lin2017focal} and AdamW~\cite{loshchilov2019decoupled} ($\eta_{\max}=3\times10^{-4}$, 5-epoch linear warm-up, cosine decay to $0.02\eta_{\max}$, weight decay 0.05; \cref{tab:training}). \rev{Hyperparameters (e.g., peak learning rate, weight decay, and dropout probabilities) were determined through empirical tuning on the validation splits of the BreakHis and LC25000 datasets, and subsequently kept fixed across all six benchmarks to prevent dataset-specific overfitting; test partitions and the external cohort played no part in this choice.} The 72 scalar graph priors receive a $10\times$ learning rate ($3\times10^{-3}$) with zero weight decay so they can move effectively from initialisation. GPU-based training augmentation\rev{, applied on the fly to training mini-batches only and after partitioning (validation and test tiles are never augmented),} comprises random $D_4$ dihedral views, HED stain jitter ($\pm0.05$, $p=0.8$)~\cite{tellez2019quantifying}, brightness/contrast jitter ($\pm8\%$), and—when evaluating sensor-noise robustness (\cref{sec:r_robust})—additive Gaussian noise augmentation ($\sigma\sim\mathcal{U}(0,0.03)$, $p=0.5$). To ensure fair comparative evaluations, all from-scratch baselines are trained under the exact same focal loss, augmentation pipeline, epoch count, and optimiser scheduling (omitting the $10\times$ prior multiplier, which is specific to \model{}'s scalars).

\begin{table}[!htbp]
\centering\footnotesize
\setlength{\tabcolsep}{4pt}
\caption{Training configuration and initialisation.}\label{tab:training}
\begin{tabularx}{\textwidth}{@{}l p{8.6cm} X@{}}
\toprule
\textbf{Item} & \textbf{Value} & \textbf{Rationale}\\
\midrule
Optimiser & AdamW, $\beta=(0.9,0.999)$, $\epsilon=10^{-8}$ & decoupled weight decay\\
\rev{Hyperparameter tuning} & \rev{empirical manual search on the validation splits (BreakHis, LC25000)} & \rev{values fixed across all datasets to avoid overfitting}\\
Parameter groups & decay: 5{,}155{,}584 params; no-decay: 170{,}714; graph priors: 72 scalars & priors and norms are not regularised towards zero\\
Learning rate & $3\times10^{-4}$; priors $3\times10^{-3}$ ($\times10$); 5-epoch warm-up, 50-epoch cosine & scalar priors must be able to move\\
Batch size & 64 (32 for BreakHis and Lymphoma) & fits in 3.6\,GB VRAM on a 16\,GB T4\\
Loss & class-weighted focal loss, $\gamma=2$ & class imbalance and easy-sample down-weighting\\
Regularisation & weight decay 0.05; DropPath $0\to0.1$; head dropout 0.2; grad clip 1.0 & over-fitting control\\
Augmentation & $D_4$ (8 views); HED jitter $\pm0.05$, $p=0.8$; brightness/contrast $\pm8\%$ & orientation and stain-level invariance\\
Initialisation & $\beta_H,\beta_E=\ln(e-1)$; $\gamma=0$; FiLM $W_f=0$; seeds $\mathcal{N}(0,1)$; rel.\ bias $=0$ & neutral priors, identity feedback, diverse compartments\\
Model selection & best validation macro-F1; temperature fitted on validation & no test-set tuning\\
\bottomrule
\end{tabularx}
\end{table}

%% ---------------------------------------------------------------------
\subsection{Evaluation protocol}\label{sec:m_eval}

\paragraph{Classification, cross-validation and external generalisation.} We report accuracy, macro-F1 (primary metric), balanced accuracy, macro one-versus-rest AUROC, AUPRC, MCC, Cohen's $\kappa$ and specificity. We evaluate (i) 5-seed mean $\pm$ SD (seeds 42, 100, 777, 999, 2026) on the \ssip{} splits, which are the main numbers reported throughout, with seed 42 reported as a \emph{representative run} (with 95\% group-bootstrap confidence intervals) wherever only single-run analyses exist, (ii) 5-fold group-stratified cross-validation including \textbf{strict patient-level grouping on BreakHis (81 patients)}, and (iii) \textbf{external cohort evaluation without adaptation} on CRC-VAL-HE-7K (7{,}180 tiles, 50 patients) for models trained on NCT-CRC-HE-100K. \rev{The predicted class is the arg-max of the softmax output; no decision threshold or class-specific rule is tuned. Test partitions and the external cohort are evaluated only after training and checkpoint selection are complete, and they are not used for any development decision (\cref{tab:leakctrl}).}

\paragraph{Three-tier baseline comparison and effect sizes.} To address both architectural inductive bias and practical transfer learning, baselines are evaluated in three tiers on the exact same splits:
\begin{enumerate}
  \item \emph{From-scratch architectural comparison:} Eleven backbones (CNNs, ViTs, Swin-Tiny~\cite{liu2021swin}, ViG-Tiny~\cite{han2022vision}, DeepGCN-Res, Cell-Graph GNN, and the CTransPath hybrid architecture~\cite{wang2022ctranspath}) trained from random initialisation.
  \item \emph{Pretrained fine-tuning and stain-normalised baselines:} ImageNet-pretrained ResNet-50, ConvNeXt-Tiny and Swin-Tiny fine-tuned end to end; an offline Macenko stain-normalised~\cite{macenko2009method} ConvNeXt-Tiny baseline (to test in-model stain priors against standard stain normalisation); and linear probes of frozen pretrained pathology foundation encoders (CTransPath~\cite{wang2022ctranspath} and UNI~\cite{chen2024uni}); the foundation encoders were not fine-tuned.
  \item \emph{Effect sizes:} Because the five baseline seeds were not paired with \model{}'s runs at the level of individual runs, we do not report $p$-values. Instead we report the difference of five-seed means between \model{} and the best baseline, with $\pm\sqrt{\mathrm{SD}_1^2+\mathrm{SD}_2^2}$, which treats the two sets of runs as independent.
\end{enumerate}

\paragraph{Calibration and selective prediction.} A temperature $T^{\star}=\arg\min_T\mathrm{NLL}_{\mathrm{va}}(\mathrm{softmax}(\mathbf{o}/T))$ is fitted on the validation set \rev{only and then applied, unchanged, to the test logits $\mathbf{o}$; all calibration metrics are computed on the test partition, which played no part in fitting $T^{\star}$.} We report expected and maximum calibration errors (ECE, MCE):
\begin{equation}
\mathrm{ECE}=\sum_{b}\frac{|B_b|}{n}\big|\mathrm{acc}(B_b)-\mathrm{conf}(B_b)\big|,\qquad \mathrm{MCE}=\max_b\big|\mathrm{acc}(B_b)-\mathrm{conf}(B_b)\big|,
\label{eq:ece}
\end{equation}
Brier score, and selective-prediction accuracy at 95\%, 90\% and 80\% coverage alongside the area under the risk--coverage curve (AURC).

\paragraph{Structural transparency and explanation faithfulness audit.} We distinguish \emph{structural transparency} (inspecting the 72 learned priors $\sigma_+(\beta_H),\sigma_+(\beta_E)$ and gates $g$, measuring Spearman correlations between edge attention and stain differences $|\Delta h|,|\Delta e|$, and probing a linear classifier trained solely on 43 topological graph descriptors) from \emph{post-hoc attribution faithfulness}. For faithfulness, we evaluate pixel deletion and insertion AUCs,
\begin{equation}
\mathbf{I}^{\mathrm{del}}_t=\mathbf{I}\odot(1-\mathbb{1}_{\mathcal{T}_t})+\mathbf{b}\odot\mathbb{1}_{\mathcal{T}_t},\quad
\mathbf{I}^{\mathrm{ins}}_t=\mathbf{b}\odot(1-\mathbb{1}_{\mathcal{T}_t})+\mathbf{I}\odot\mathbb{1}_{\mathcal{T}_t},\quad
\mathrm{AUC}_{\mathrm{del/ins}}=\int_0^1 p_y\big(\mathbf{I}^{\mathrm{del/ins}}_t\big)\,dt,
\label{eq:delins}
\end{equation}
nuclear attribution precision ($\mathrm{NP}@10\%$), stain-shift stability, and Gini concentration $G(\mathbf{m})$,
\begin{equation}
\mathrm{NP}@10\%=\frac{|\mathcal{T}_{0.1}(\mathbf{m})\cap\mathcal{N}|}{|\mathcal{T}_{0.1}(\mathbf{m})|},\qquad
\mathrm{Stab}=\mathrm{corr}\big(\mathbf{m}(\mathbf{I}),\,\mathbf{m}(\tilde{\mathbf{I}})\big),\qquad
G(\mathbf{m})=\frac{\sum_{p}\sum_{q}|m_p-m_q|}{2P\sum_{p}|m_p|},
\label{eq:np}
\end{equation}
alongside direct graph-level edge deletion and compartment occlusion (ordered by readout weight $r_m$ versus compartment spatial area):
\begin{equation}
\Delta_{\mathrm{edge}}(k)=p_y(\mathbf{I})-p_y\big(\mathbf{I};\,\mathcal{E}^{\mathrm{c}}\setminus\mathcal{E}^{\mathrm{c}}_{(k)}\big),\qquad
\Delta_{\mathrm{comp}}(k)=p_y(\mathbf{I})-p_y\big(\mathbf{I};\,\mathcal{V}^{\mathrm{k}}\setminus\mathcal{V}^{\mathrm{k}}_{(k)}\big).
\label{eq:graphdel}
\end{equation}

\subsection{Implementation and reproducibility}\label{sec:m_impl}

All experiments were run in PyTorch 2.11 (CUDA 12.8, Python 3.13) on NVIDIA Tesla T4 GPUs (16\,GB). To remove the data-loading bottleneck, each dataset is decoded once and cached as \texttt{uint8} tensors in GPU memory, or as a disk memory map for NCT-CRC-HE-100K. All augmentation runs on the GPU. \Cref{tab:hardware} reports the cache sizes and the resulting epoch times. The peak training memory at batch size 64 with mixed precision is 3.6\,GB, so the full protocol fits on a single entry-level data-centre GPU. Seeds are fixed for Python, NumPy and PyTorch, and seed 42 serves as the representative run. The per-dataset notebooks that generate the \ssip{} group assignments, every split and every reported table and figure for all six datasets are uploaded to Zenodo (\cref{sec:availability}). All notebooks were run in the Kaggle environment (Kaggle notebooks with NVIDIA Tesla T4 GPUs), which is required to run them as provided; trained weights and other outputs are not archived.

\begin{table}[!htbp]
\centering\footnotesize
\setlength{\tabcolsep}{5pt}
\caption{Data caching and training cost on a Tesla T4 (mean epoch time over 50 epochs).}\label{tab:hardware}
\begin{tabular}{@{}l r l r@{}}
\toprule
\textbf{Dataset} & \textbf{Cache size} & \textbf{Location} & \textbf{Epoch time (s)}\\
\midrule
BreakHis & 1.19 GB & GPU memory (\texttt{uint8}) & 27.3\\
LC25000 & 3.76 GB & GPU memory (\texttt{uint8}) & 84.1\\
Lymphoma & 1.13 GB & GPU memory (\texttt{uint8}) & 26.3\\
Kather2016 & 0.75 GB & GPU memory (\texttt{uint8}) & 16.3\\
NCT-CRC-HE-100K & 15.05 GB & disk memory map & 341.7\\
Oral cancer & 0.78 GB & GPU memory (\texttt{uint8}) & 18.2\\
\bottomrule
\end{tabular}
\end{table}

%% =====================================================================
%% 4. EXPERIMENTAL RESULTS AND ANALYSIS
%% =====================================================================
\section{Experimental results and analysis}\label{sec:results}

\rev{\paragraph{Experimental environment (hardware and software).} All experiments, data caching, and model evaluations were conducted using Python 3.13 and PyTorch 2.11 (CUDA 12.8). The hardware environment consisted of single NVIDIA Tesla T4 GPUs (16\,GB VRAM) accessed via the Kaggle notebook environment. Further implementation details and cache sizes are provided in \cref{sec:m_impl}.}

\rev{\paragraph{Evaluation reporting.}} This study reports the 5-seed mean $\pm$ SD (seeds 42, 100, 777, 999 and 2026) as the main numbers on the image-level \ssip{} splits of \cref{tab:ssip}, alongside 5-fold patient-grouped cross-validation on BreakHis and external evaluation without adaptation on CRC-VAL-HE-7K. Seed 42 is reported as a representative run (with 95\% source-group bootstrap confidence intervals) for analyses that were run once; such results are labelled \emph{representative run} throughout. Per-seed results are listed in \cref{tab:perseed}. \rev{Following \cref{sec:m_leakage}, the BreakHis patient-grouped and CRC-VAL-HE-7K results are the patient-independent estimates; the \ssip{} results on the other datasets are image-level, leakage-audited benchmark results.}

%% ---------------------------------------------------------------------
\subsection{Overall classification performance across seeds}\label{sec:r_overall}

\Cref{tab:overall} reports the image-level test performance of \model{} across all six benchmarks. Across five random seeds, \model{} reaches $95.92\pm0.38\%$ accuracy ($95.31\pm0.44\%$ macro-F1) on BreakHis under image-level \ssip{} (and $91.84\pm2.37\%$ under strict patient-grouped cross-validation; \cref{sec:r_external}), $99.88\pm0.04\%$ on LC25000, $95.41\pm0.72\%$ on the Lymphoma tile set ($98.67\%$ when aggregated to the 75 test source images), $94.66\pm0.51\%$ on Kather2016, $99.69\pm0.03\%$ on internal NCT-CRC-HE-100K ($96.84\pm0.14\%$ on the external 50-patient CRC-VAL-HE-7K cohort), and $95.28\pm0.61\%$ on Oral Cancer. Balanced accuracy differs from overall accuracy by at most 0.20 percentage points (pp) on every dataset, confirming that class imbalance (\cref{tab:datasets}) does not inflate the headline figures.

\begin{figure}[!htbp]
\centering
\begin{subfigure}[t]{0.245\textwidth}\centering\includegraphics[width=\linewidth]{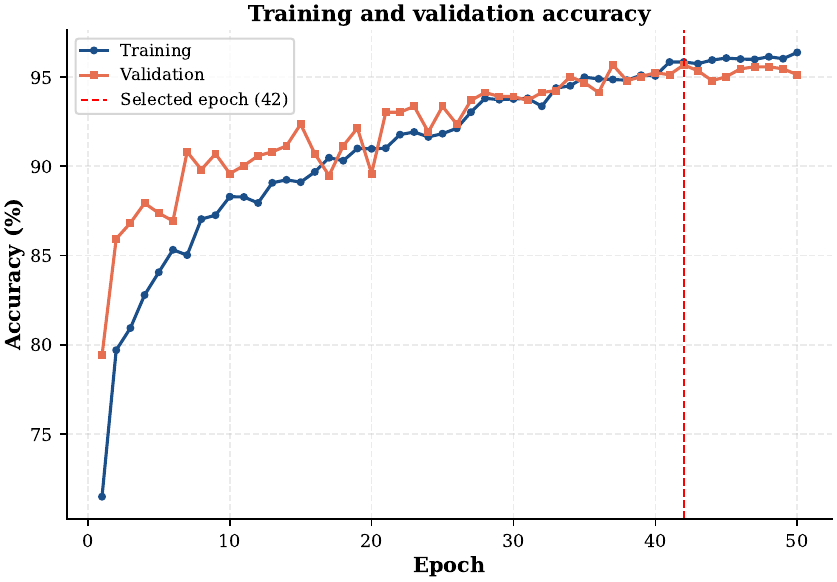}\caption{Accuracy}\end{subfigure}\hfill
\begin{subfigure}[t]{0.245\textwidth}\centering\includegraphics[width=\linewidth]{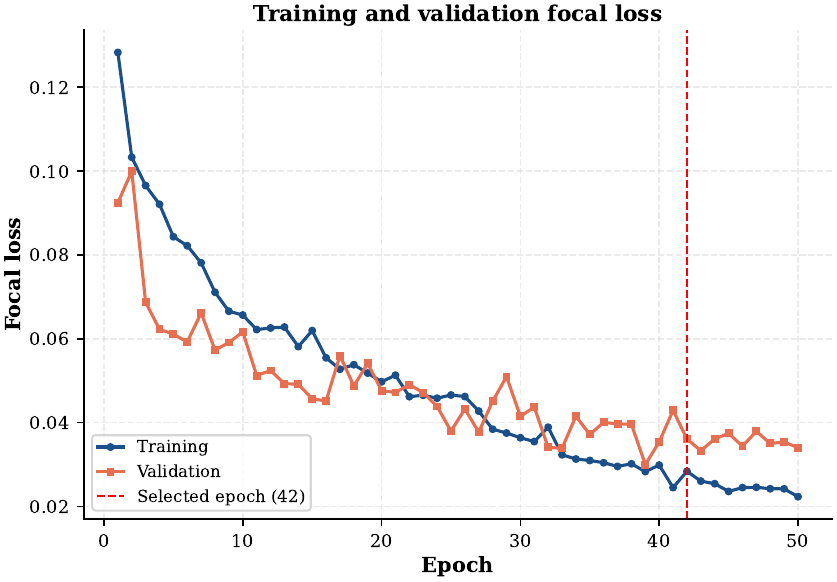}\caption{Loss}\end{subfigure}\hfill
\begin{subfigure}[t]{0.245\textwidth}\centering\includegraphics[width=\linewidth]{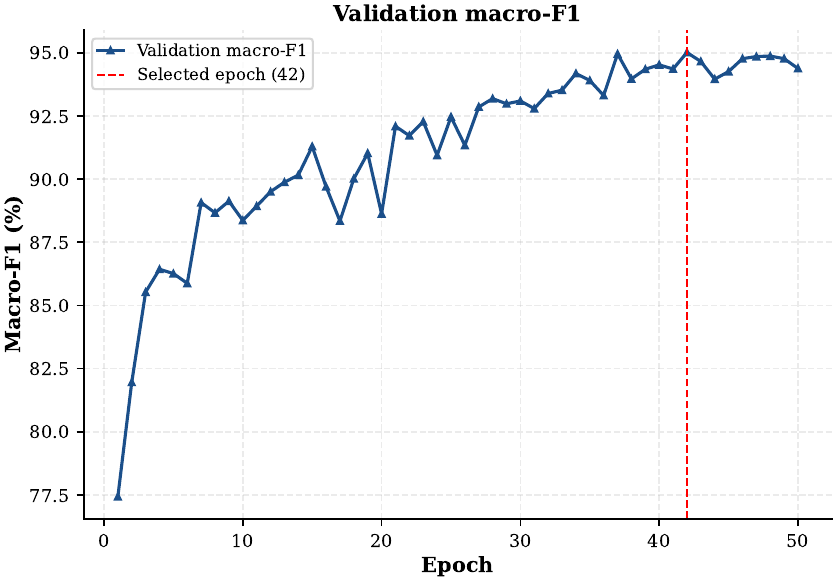}\caption{Validation macro-F1}\end{subfigure}\hfill
\begin{subfigure}[t]{0.245\textwidth}\centering\includegraphics[width=\linewidth]{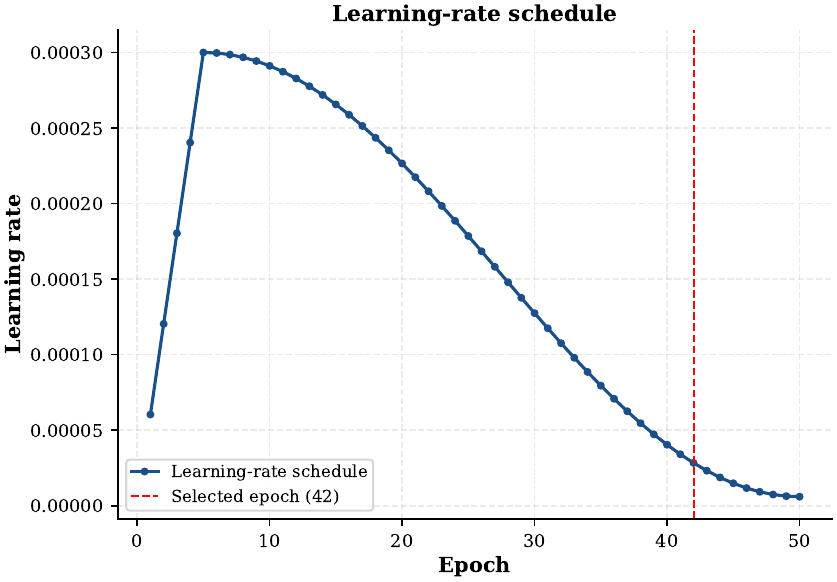}\caption{Learning rate}\end{subfigure}
\caption{Training dynamics of \model{} on BreakHis (representative run, seed 42).}
\label{fig:training}
\end{figure}

\begin{figure}[!htbp]
\centering
\begin{subfigure}[t]{0.163\textwidth}\centering\includegraphics[width=\linewidth]{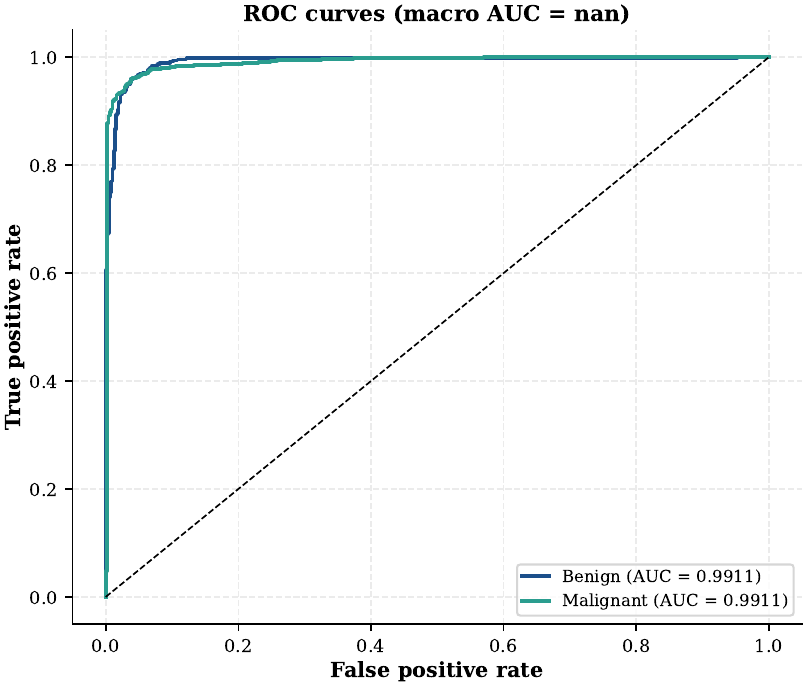}\caption{BreakHis}\end{subfigure}\hfill
\begin{subfigure}[t]{0.163\textwidth}\centering\includegraphics[width=\linewidth]{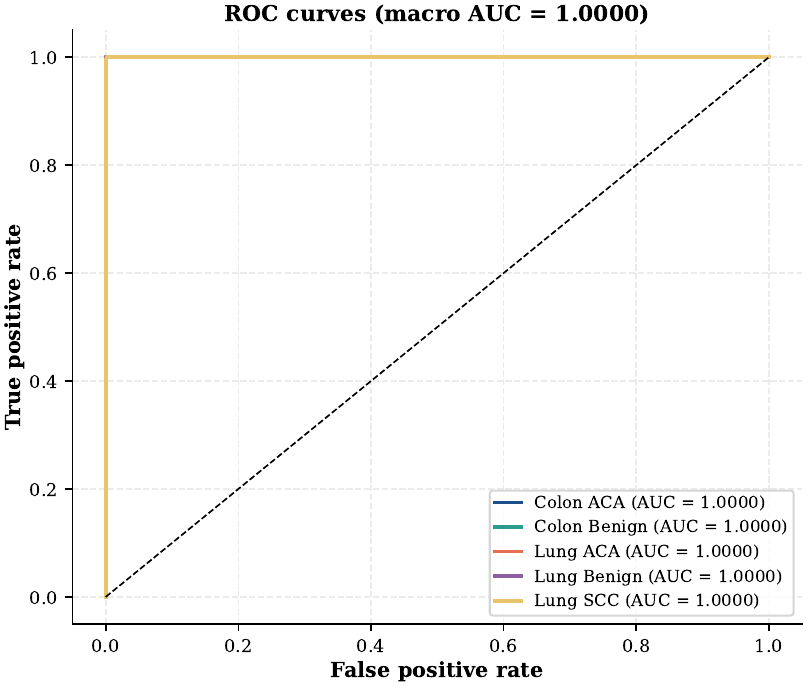}\caption{LC25000}\end{subfigure}\hfill
\begin{subfigure}[t]{0.163\textwidth}\centering\includegraphics[width=\linewidth]{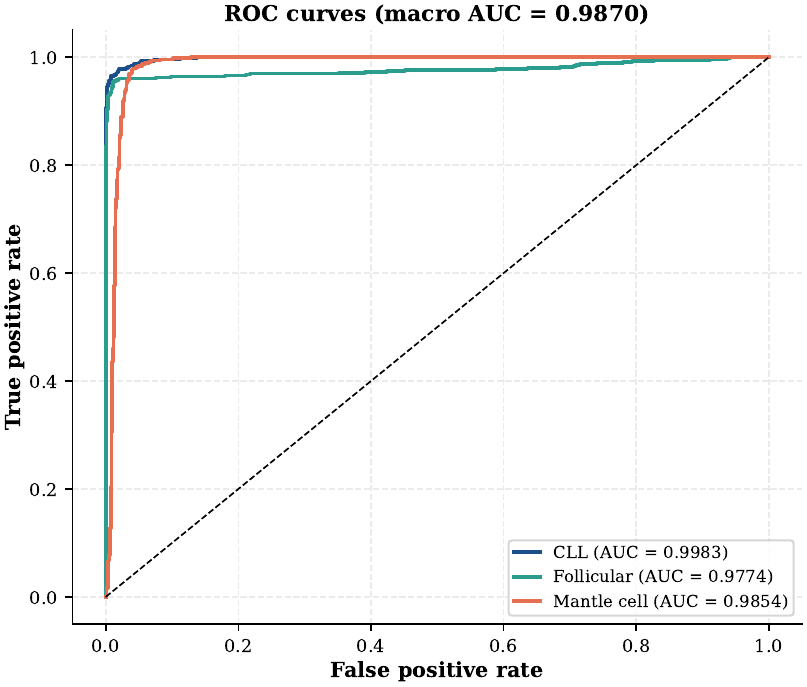}\caption{Lymphoma}\end{subfigure}\hfill
\begin{subfigure}[t]{0.163\textwidth}\centering\includegraphics[width=\linewidth]{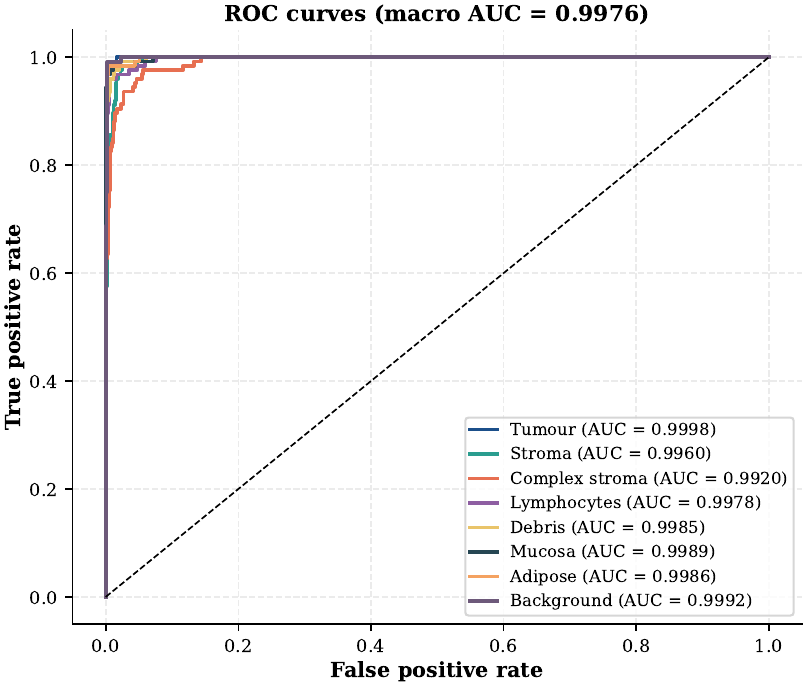}\caption{Kather2016}\end{subfigure}\hfill
\begin{subfigure}[t]{0.163\textwidth}\centering\includegraphics[width=\linewidth]{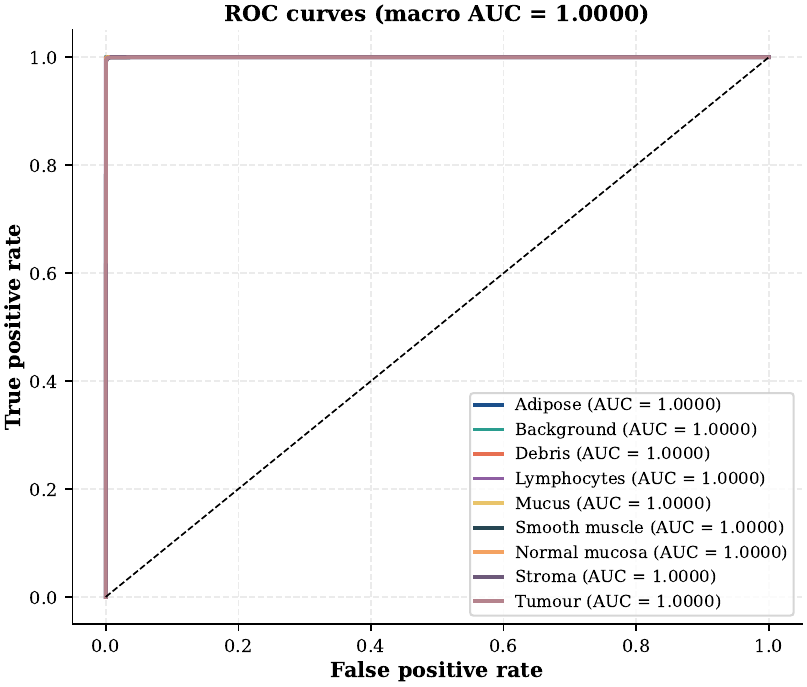}\caption{NCT-CRC-HE-100K}\end{subfigure}\hfill
\begin{subfigure}[t]{0.163\textwidth}\centering\includegraphics[width=\linewidth]{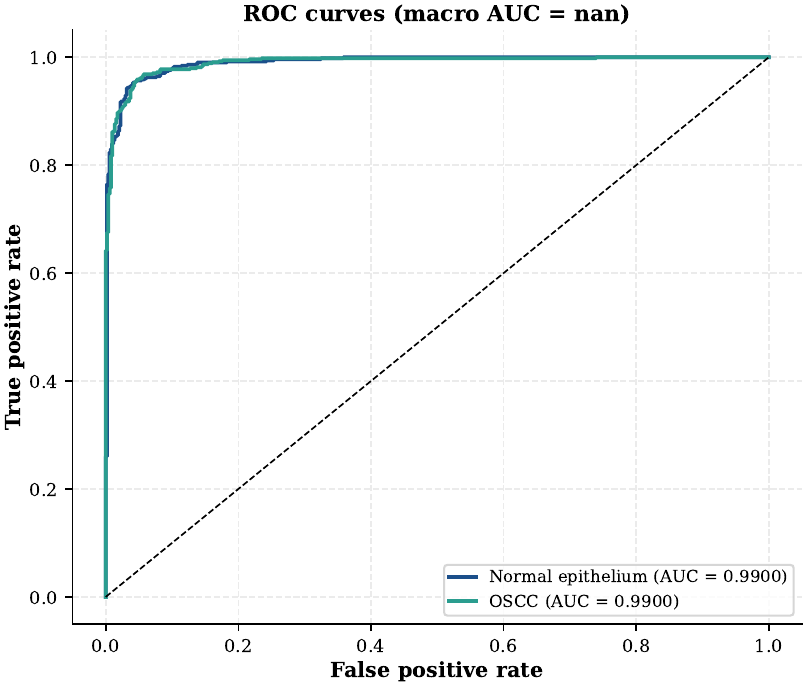}\caption{Oral cancer}\end{subfigure}\\[3pt]
\begin{subfigure}[t]{0.163\textwidth}\centering\includegraphics[width=\linewidth]{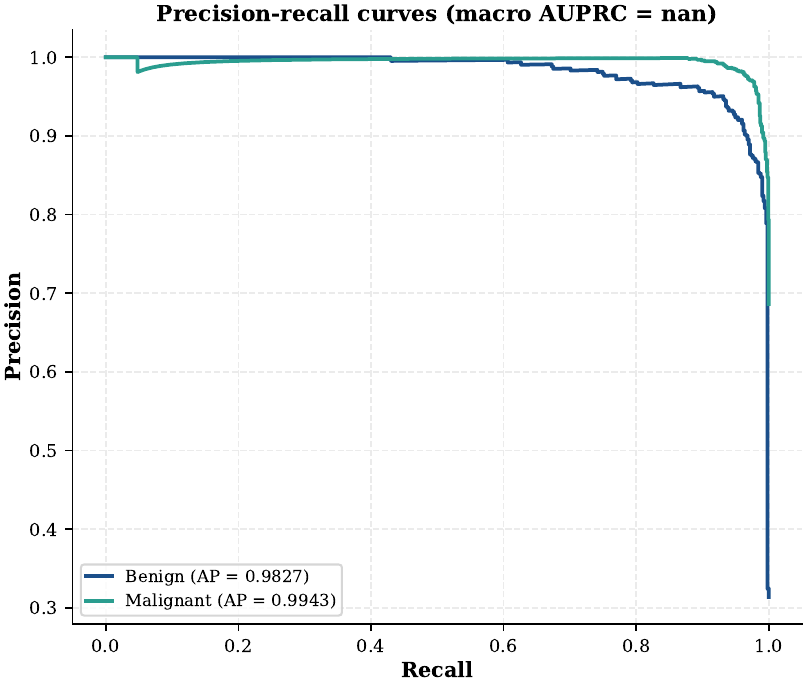}\caption{BreakHis}\end{subfigure}\hfill
\begin{subfigure}[t]{0.163\textwidth}\centering\includegraphics[width=\linewidth]{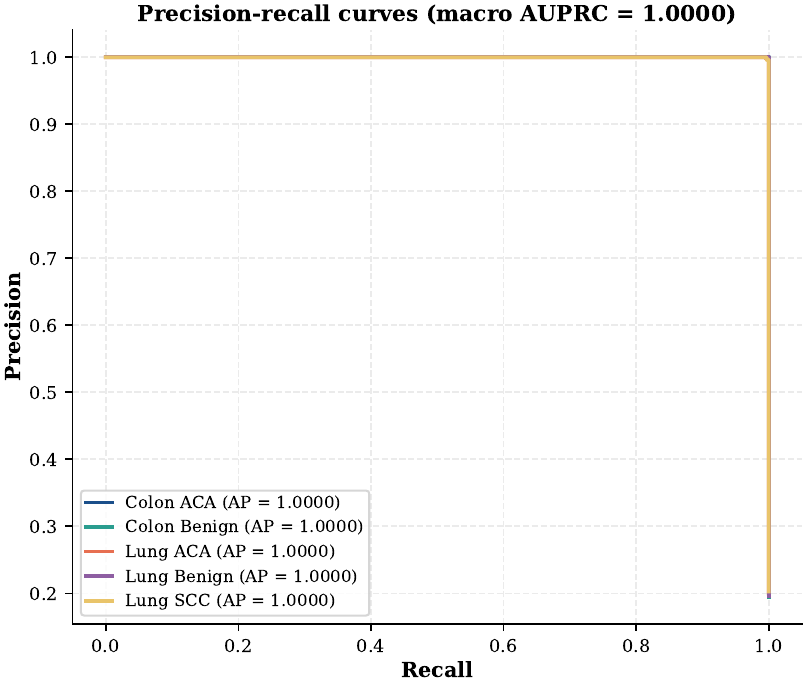}\caption{LC25000}\end{subfigure}\hfill
\begin{subfigure}[t]{0.163\textwidth}\centering\includegraphics[width=\linewidth]{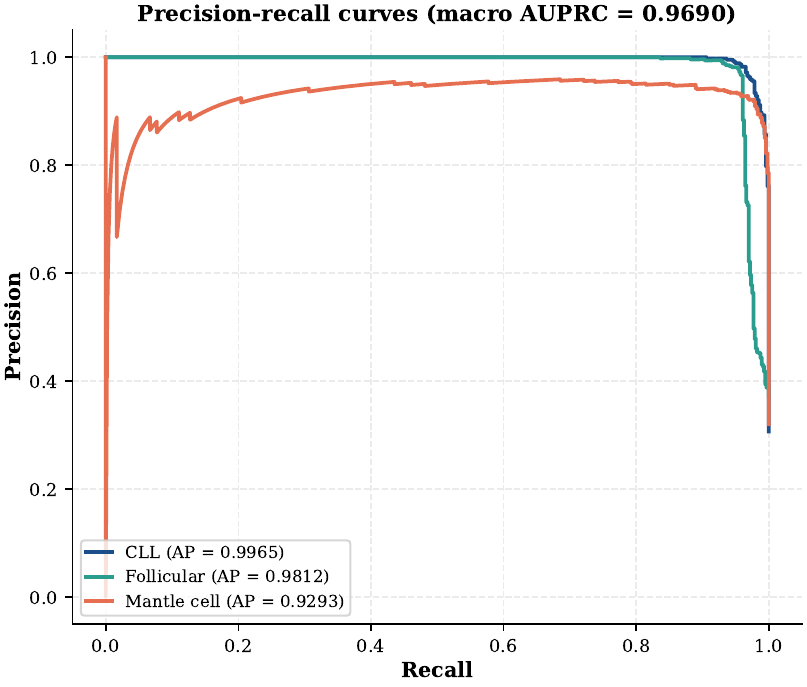}\caption{Lymphoma}\end{subfigure}\hfill
\begin{subfigure}[t]{0.163\textwidth}\centering\includegraphics[width=\linewidth]{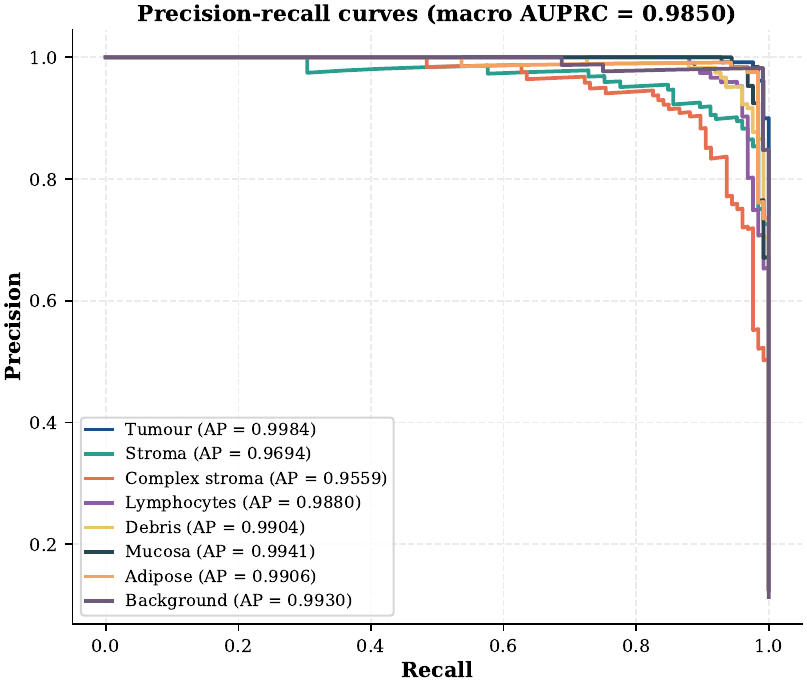}\caption{Kather2016}\end{subfigure}\hfill
\begin{subfigure}[t]{0.163\textwidth}\centering\includegraphics[width=\linewidth]{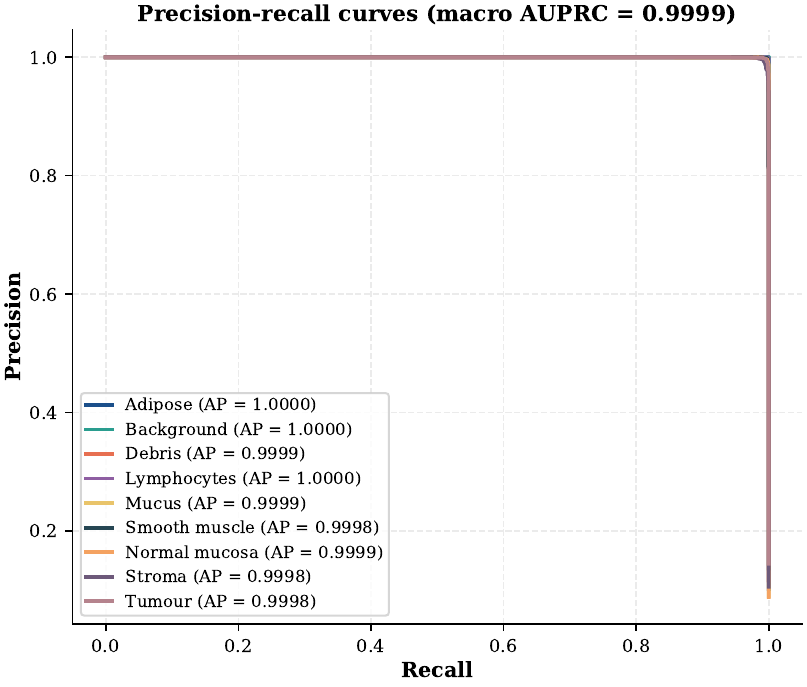}\caption{NCT-CRC-HE-100K}\end{subfigure}\hfill
\begin{subfigure}[t]{0.163\textwidth}\centering\includegraphics[width=\linewidth]{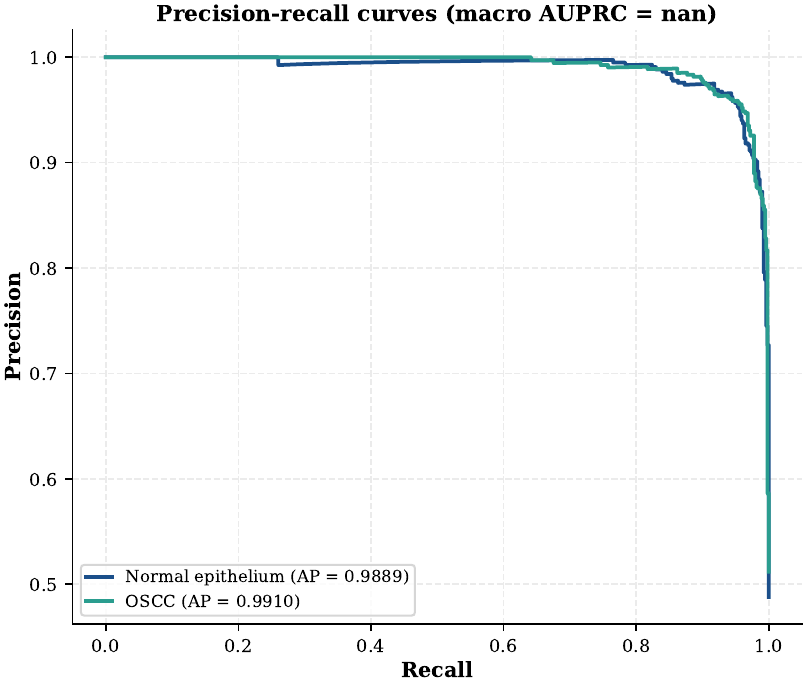}\caption{Oral cancer}\end{subfigure}
\caption{One-versus-rest ROC curves (top row) and precision--recall curves (bottom row) of \model{} on the six test sets (representative run, seed 42). For the two binary datasets (BreakHis and Oral cancer) the panel titles show \texttt{macro AUC = nan} and \texttt{macro AUPRC = nan}, because the one-versus-rest macro average computed by the plotting script is undefined for two classes; the per-class curves are valid, and the binary AUROC and AUPRC values are given in \cref{tab:extended}.}
\label{fig:rocpr}
\end{figure}

\Cref{tab:extended} reports the extended suite of 23 classification, agreement and probabilistic metrics on the representative run (seed 42). Agreement measures (MCC, Cohen's $\kappa$, Youden's $J$ and markedness) align within 0.01 on every dataset: In this seed-42 run, MCC is 0.91 on BreakHis and Oral Cancer, 0.94 on Lymphoma and Kather2016, and $>0.99$ on LC25000 and NCT-CRC-HE-100K. The width of the 95\% group-bootstrap confidence intervals reflects the number of independent source groups in the test split rather than the raw tile count: it is narrowest on NCT-CRC-HE-100K (0.15\,pp across 19{,}839 tiles) and widest on Lymphoma (6.13\,pp across 1{,}500 tiles drawn from 75 test source images). \Cref{fig:training,fig:rocpr} illustrate the training dynamics and one-versus-rest ROC/PR curves.

\begin{table}[!htbp]
\centering\footnotesize
\setlength{\tabcolsep}{2.8pt}
\caption{Image-level test performance of \model{} under \ssip{}. Columns 3--5 report the five-seed mean $\pm$ SD (seeds 42, 100, 777, 999, 2026), which are the main results; the MCC column is the five-seed mean. Columns 6--8 report the representative run (seed 42) with 95\% source-group bootstrap confidence intervals. Patient-grouped BreakHis CV ($91.84\pm2.37\%$) and external CRC-VAL-HE-7K ($96.84\pm0.14\%$) are reported in \cref{tab:cv,tab:external}.}\label{tab:overall}
\begin{tabular}{@{}l r c c c c c c@{}}
\toprule
 & & \multicolumn{3}{c}{\textbf{Five seeds (mean $\pm$ SD)}} & \multicolumn{3}{c}{\textbf{Representative run (seed 42) [95\% group-bootstrap CI]}}\\
\cmidrule(lr){3-5}\cmidrule(l){6-8}
\textbf{Dataset} & $n_{\mathrm{test}}$ & \textbf{Accuracy (\%)} & \textbf{Macro-F1 (\%)} & \textbf{MCC ($\times100$)} & \textbf{Accuracy (\%)} & \textbf{Macro-F1 (\%)} & \textbf{AUROC}\\
\midrule
BreakHis (\ssip{}) & 1{,}578 & 95.92 $\pm$ 0.38 & 95.31 $\pm$ 0.44 & 90.62 $\pm$ 0.89 & 96.07 [95.09, 97.03] & 95.48 [94.33, 96.59] & 0.9911\\
LC25000 & 4{,}849 & 99.88 $\pm$ 0.04 & 99.88 $\pm$ 0.04 & 99.85 $\pm$ 0.05 & 99.92 [99.75, 100.00] & 99.92 [99.75, 100.00] & 1.0000\\
Lymphoma (tiles) & 1{,}500 & 95.41 $\pm$ 0.72 & 95.40 $\pm$ 0.73 & 93.11 $\pm$ 1.08 & 95.87 [92.20, 98.33] & 95.87 [92.20, 98.26] & 0.9870\\
Kather2016 & 986 & 94.66 $\pm$ 0.51 & 94.74 $\pm$ 0.50 & 93.91 $\pm$ 0.58 & 94.83 [93.33, 96.15] & 94.92 [93.51, 96.22] & 0.9976\\
NCT-CRC-HE-100K & 19{,}839 & 99.69 $\pm$ 0.03 & 99.68 $\pm$ 0.03 & 99.65 $\pm$ 0.03 & 99.72 [99.65, 99.80] & 99.72 [99.65, 99.80] & 1.0000\\
Oral cancer & 1{,}047 & 95.28 $\pm$ 0.61 & 95.27 $\pm$ 0.61 & 90.57 $\pm$ 1.21 & 95.51 [93.50, 97.15] & 95.51 [93.48, 97.14] & 0.9900\\
\bottomrule
\end{tabular}
\end{table}

\begin{table}[!htbp]
\centering\footnotesize
\setlength{\tabcolsep}{3.5pt}
\caption{Per-seed test results of \model{} (accuracy / macro-F1, \%) for seeds 42, 100, 777, 999 and 2026, with the five-seed mean $\pm$ SD (sample SD). Rows 1--6 are image-level \ssip{} test sets; the last row is the external CRC-VAL-HE-7K cohort evaluated without adaptation after training on NCT-CRC-HE-100K. The notebooks that produce these results are available on Zenodo.}\label{tab:perseed}
\begin{tabular}{@{}l c c c c c c@{}}
\toprule
\textbf{Dataset} & \textbf{Seed 42} & \textbf{Seed 100} & \textbf{Seed 777} & \textbf{Seed 999} & \textbf{Seed 2026} & \textbf{Mean $\pm$ SD}\\
\midrule
BreakHis (\ssip{}) & 96.07 / 95.48 & 96.35 / 95.81 & 96.13 / 95.56 & 95.60 / 94.94 & 95.45 / 94.76 & 95.92$\pm$0.38 / 95.31$\pm$0.44\\
LC25000 & 99.92 / 99.92 & 99.92 / 99.92 & 99.88 / 99.88 & 99.85 / 99.85 & 99.83 / 99.83 & 99.88$\pm$0.04 / 99.88$\pm$0.04\\
Lymphoma (tiles) & 95.87 / 95.87 & 96.23 / 96.25 & 95.53 / 95.50 & 95.01 / 94.98 & 94.41 / 94.40 & 95.41$\pm$0.72 / 95.40$\pm$0.73\\
Kather2016 & 94.83 / 94.92 & 95.36 / 95.39 & 94.79 / 94.91 & 94.26 / 94.34 & 94.06 / 94.14 & 94.66$\pm$0.51 / 94.74$\pm$0.50\\
NCT-CRC-HE-100K & 99.72 / 99.72 & 99.71 / 99.70 & 99.70 / 99.68 & 99.67 / 99.65 & 99.65 / 99.65 & 99.69$\pm$0.03 / 99.68$\pm$0.03\\
Oral cancer & 95.51 / 95.51 & 96.10 / 96.09 & 95.43 / 95.41 & 94.78 / 94.77 & 94.58 / 94.57 & 95.28$\pm$0.61 / 95.27$\pm$0.61\\
CRC-VAL-HE-7K (external) & 96.84 / 96.42 & 97.01 / 96.61 & 96.94 / 96.54 & 96.74 / 96.30 & 96.67 / 96.23 & 96.84$\pm$0.14 / 96.42$\pm$0.16\\
\bottomrule
\end{tabular}
\end{table}

\begin{table}[!htbp]
\centering\footnotesize
\setlength{\tabcolsep}{4pt}
\caption{Extended evaluation of \model{} on the six test sets (representative run, seed 42; point estimates). All MCC values in this table come from seed 42. Macro metrics average one-versus-rest values over classes. ECE is reported before temperature scaling.}\label{tab:extended}
\begin{tabular}{@{}l c c c c c c@{}}
\toprule
\textbf{Metric} & \textbf{BreakHis} & \textbf{LC25000} & \textbf{Lymphoma} & \textbf{Kather2016} & \textbf{NCT-CRC-HE-100K} & \textbf{Oral cancer}\\
\midrule
Accuracy (\%) & 96.07 & 99.92 & 95.87 & 94.83 & 99.72 & 95.51\\
Balanced accuracy (\%) & 95.88 & 99.92 & 95.88 & 94.89 & 99.72 & 95.49\\
Error rate (\%) & 3.93 & 0.08 & 4.13 & 5.17 & 0.28 & 4.49\\
\midrule
Macro precision (\%) & 95.11 & 99.92 & 95.87 & 95.01 & 99.73 & 95.54\\
Macro recall / sensitivity (\%) & 95.88 & 99.92 & 95.88 & 94.89 & 99.72 & 95.49\\
Macro specificity (\%) & 95.88 & 99.98 & 97.93 & 99.26 & 99.97 & 95.49\\
Macro NPV (\%) & 95.11 & 99.98 & 97.92 & 99.26 & 99.97 & 95.54\\
\midrule
Macro-F1 (\%) & 95.48 & 99.92 & 95.87 & 94.92 & 99.72 & 95.51\\
Weighted F1 (\%) & 96.09 & 99.92 & 95.88 & 94.86 & 99.72 & 95.51\\
Macro Jaccard (\%) & 91.39 & 99.84 & 92.10 & 90.53 & 99.44 & 91.40\\
Macro G-mean (\%) & 95.87 & 99.95 & 96.90 & 97.04 & 99.84 & 95.48\\
Macro FPR (\%) & 4.12 & 0.02 & 2.07 & 0.74 & 0.03 & 4.51\\
Macro FNR (\%) & 4.12 & 0.08 & 4.12 & 5.11 & 0.28 & 4.51\\
\midrule
MCC & 0.9098 & 0.9990 & 0.9379 & 0.9410 & 0.9969 & 0.9103\\
Cohen's $\kappa$ & 0.9095 & 0.9990 & 0.9378 & 0.9409 & 0.9969 & 0.9101\\
Youden's $J$ (informedness) & 0.9175 & 0.9990 & 0.9382 & 0.9415 & 0.9968 & 0.9098\\
Markedness & 0.9021 & 0.9990 & 0.9379 & 0.9427 & 0.9969 & 0.9108\\
\midrule
AUROC (macro OvR) & 0.9911 & 1.0000 & 0.9870 & 0.9976 & 1.0000 & 0.9900\\
AUPRC (macro) & 0.9885 & 1.0000 & 0.9690 & 0.9850 & 0.9999 & 0.9899\\
Top-2 accuracy (\%) & -- & 100.00 & 98.60 & 98.88 & 99.96 & --\\
\midrule
Log-loss & 0.1474 & 0.0079 & 0.1671 & 0.1913 & 0.0167 & 0.1479\\
Brier score & 0.0707 & 0.0017 & 0.0679 & 0.0869 & 0.0053 & 0.0777\\
ECE, uncalibrated (\%) & 6.28 & 0.71 & 1.05 & 6.43 & 0.98 & 4.49\\
\bottomrule
\end{tabular}
\end{table}

%% ---------------------------------------------------------------------
\subsection{Per-class behaviour, image-level aggregation and magnification analysis}\label{sec:r_perclass}

\begin{figure}[!htbp]
{\centering
\begin{subfigure}[t]{0.32\textwidth}\centering\includegraphics[width=\linewidth]{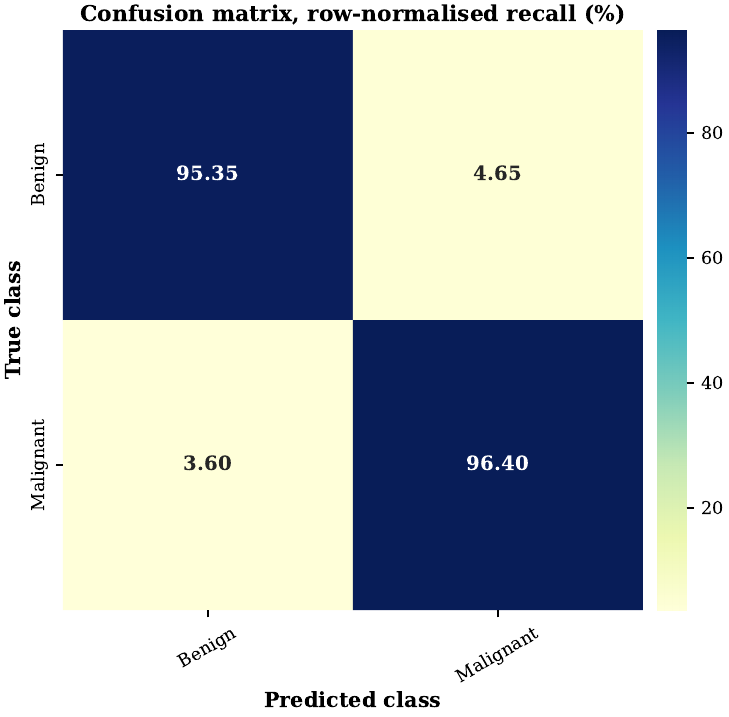}\caption{BreakHis}\end{subfigure}\hfill
\begin{subfigure}[t]{0.32\textwidth}\centering\includegraphics[width=\linewidth]{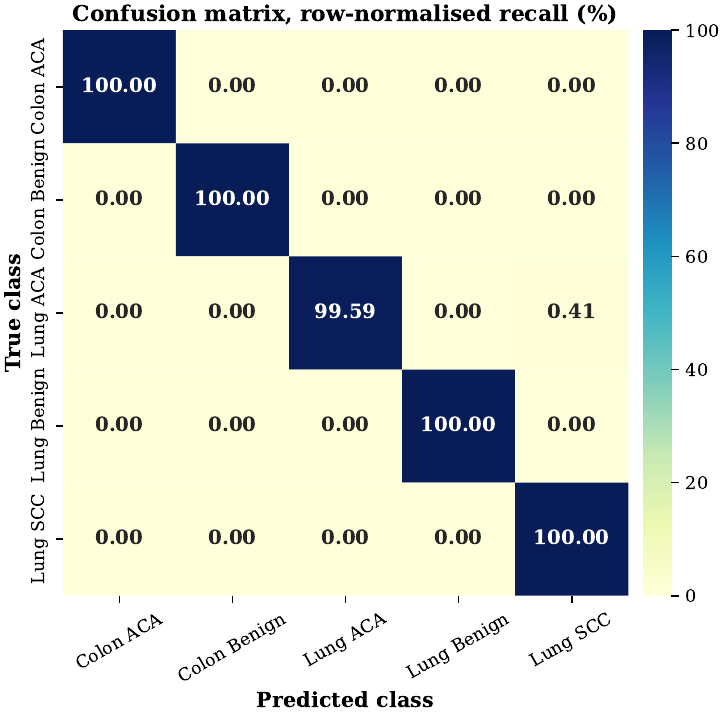}\caption{LC25000}\end{subfigure}\hfill
\begin{subfigure}[t]{0.32\textwidth}\centering\includegraphics[width=\linewidth]{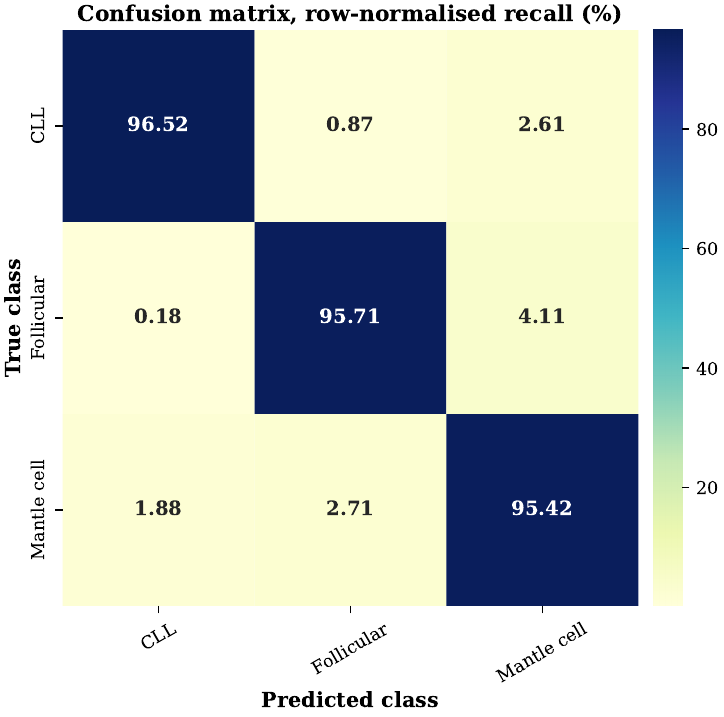}\caption{Lymphoma}\end{subfigure}\\[4pt]
\begin{subfigure}[t]{0.32\textwidth}\centering\includegraphics[width=\linewidth]{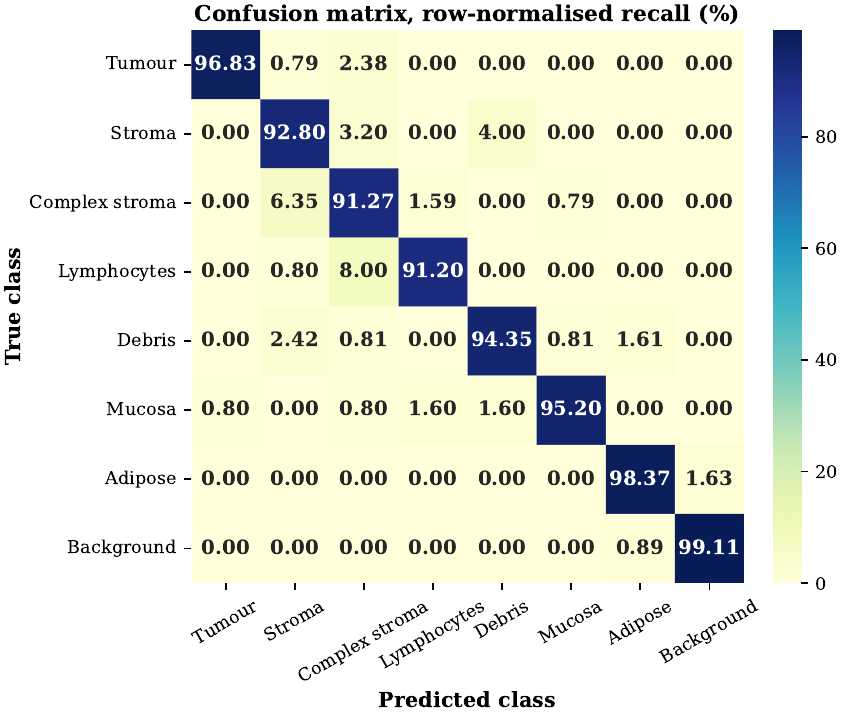}\caption{Kather2016}\end{subfigure}\hfill
\begin{subfigure}[t]{0.32\textwidth}\centering\includegraphics[width=\linewidth]{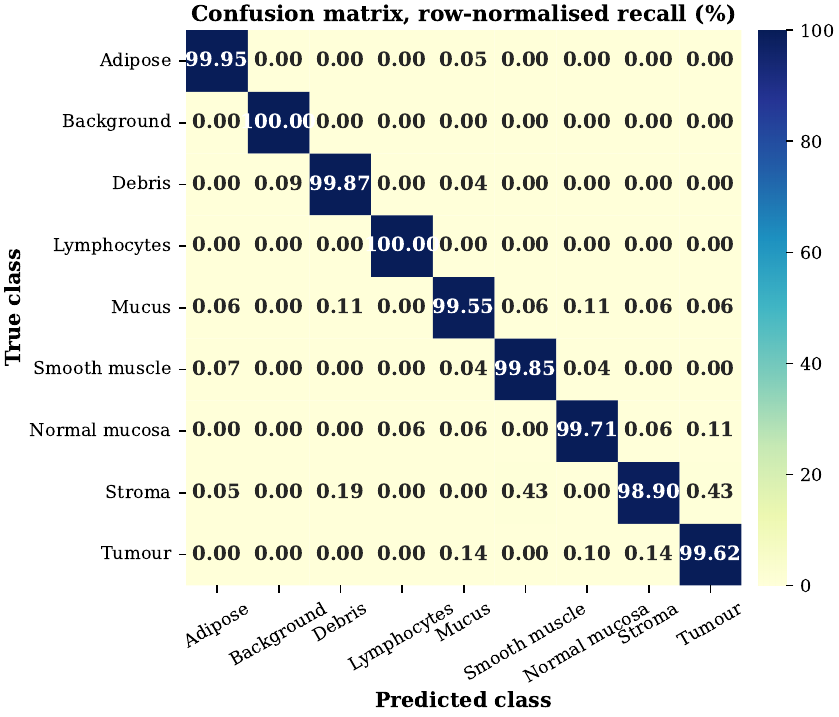}\caption{NCT-CRC-HE-100K}\end{subfigure}\hfill
\begin{subfigure}[t]{0.32\textwidth}\centering\includegraphics[width=\linewidth]{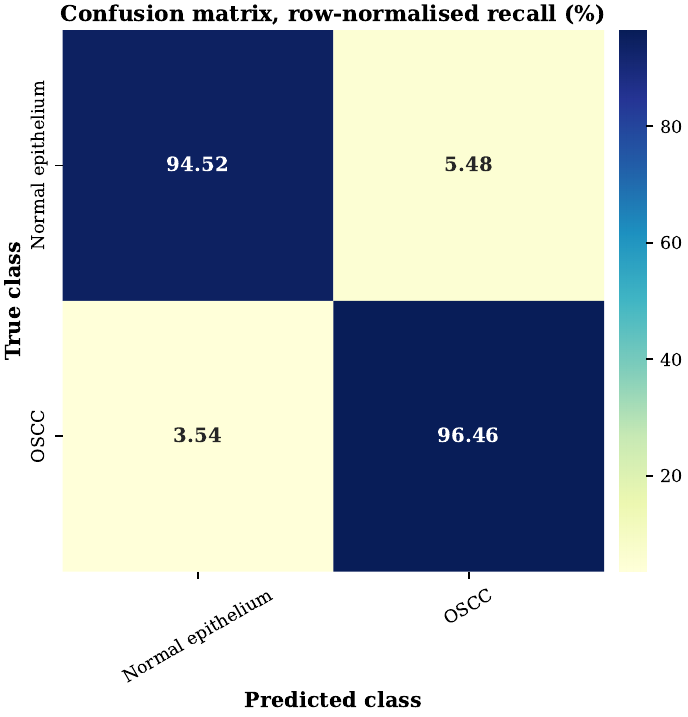}\caption{Oral cancer}\end{subfigure}\par}
\par\medskip
{\centering
\begin{subfigure}[t]{0.32\textwidth}\centering\includegraphics[width=\linewidth]{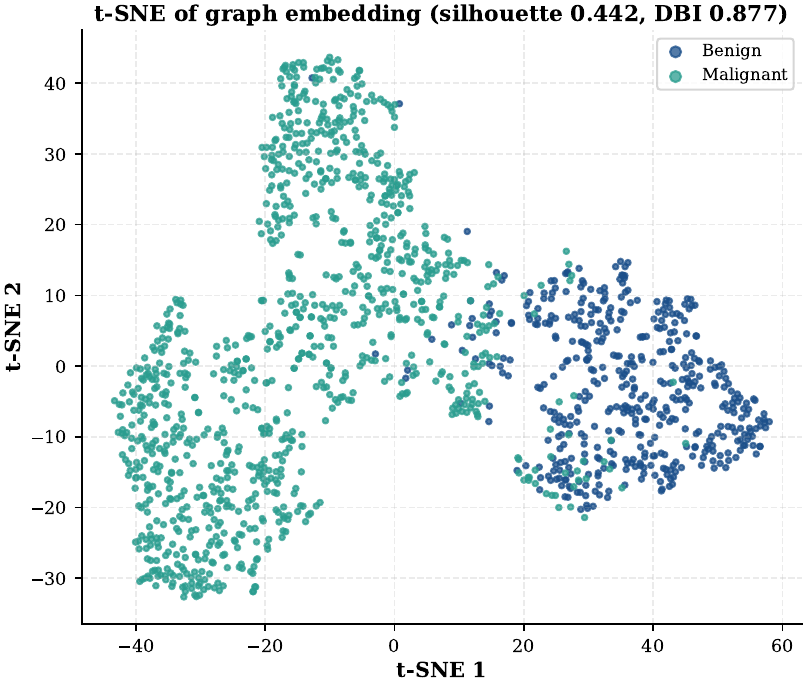}\caption{BreakHis}\end{subfigure}\hfill
\begin{subfigure}[t]{0.32\textwidth}\centering\includegraphics[width=\linewidth]{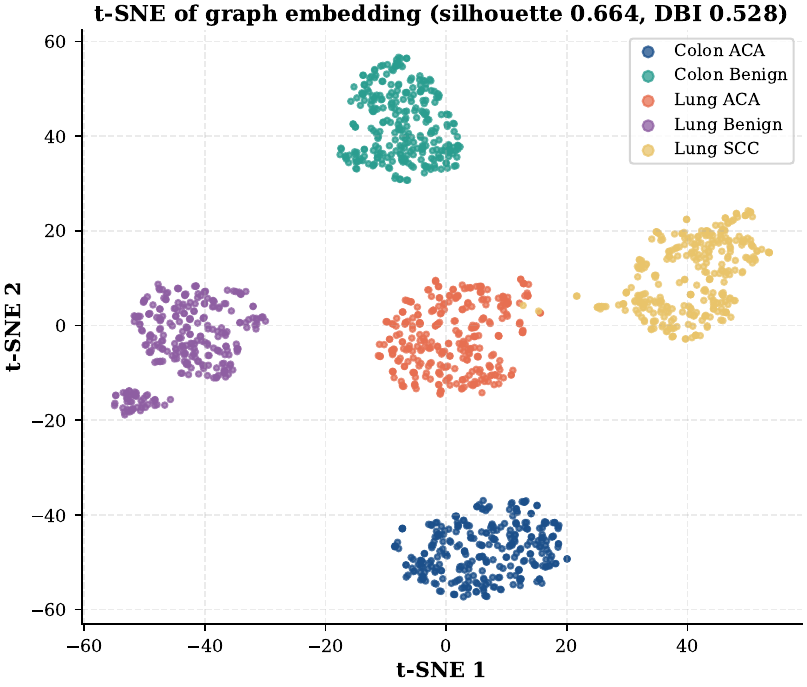}\caption{LC25000}\end{subfigure}\hfill
\begin{subfigure}[t]{0.32\textwidth}\centering\includegraphics[width=\linewidth]{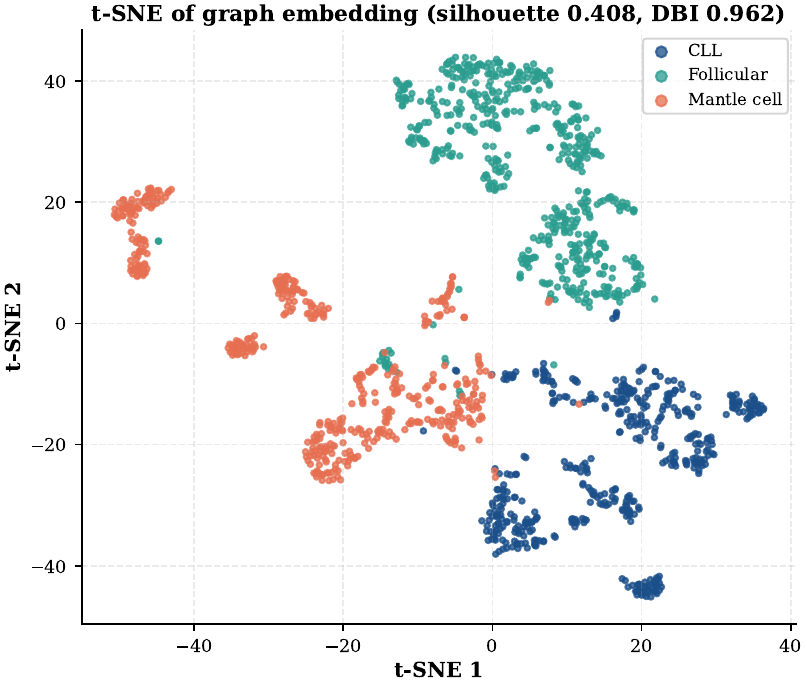}\caption{Lymphoma}\end{subfigure}\\[4pt]
\begin{subfigure}[t]{0.32\textwidth}\centering\includegraphics[width=\linewidth]{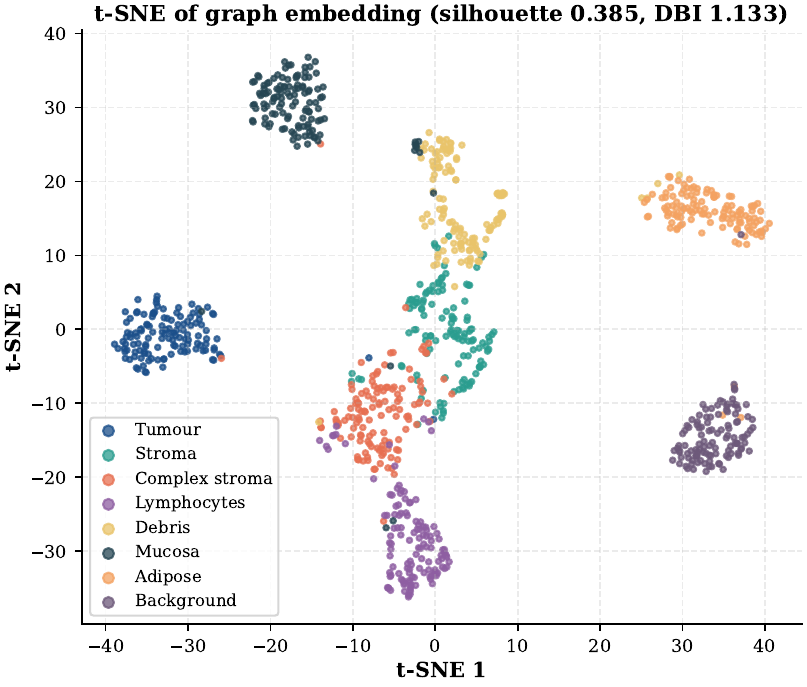}\caption{Kather2016}\end{subfigure}\hfill
\begin{subfigure}[t]{0.32\textwidth}\centering\includegraphics[width=\linewidth]{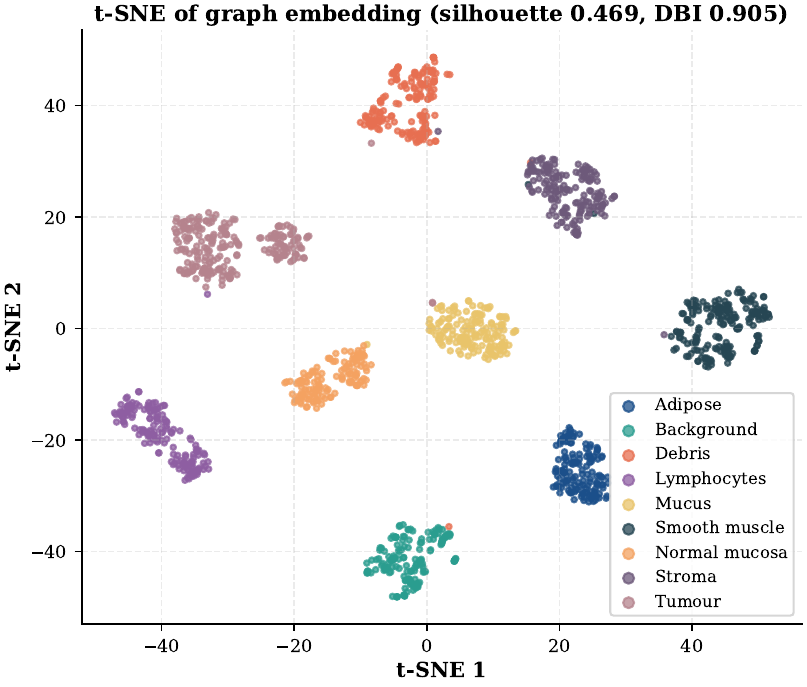}\caption{NCT-CRC-HE-100K}\end{subfigure}\hfill
\begin{subfigure}[t]{0.32\textwidth}\centering\includegraphics[width=\linewidth]{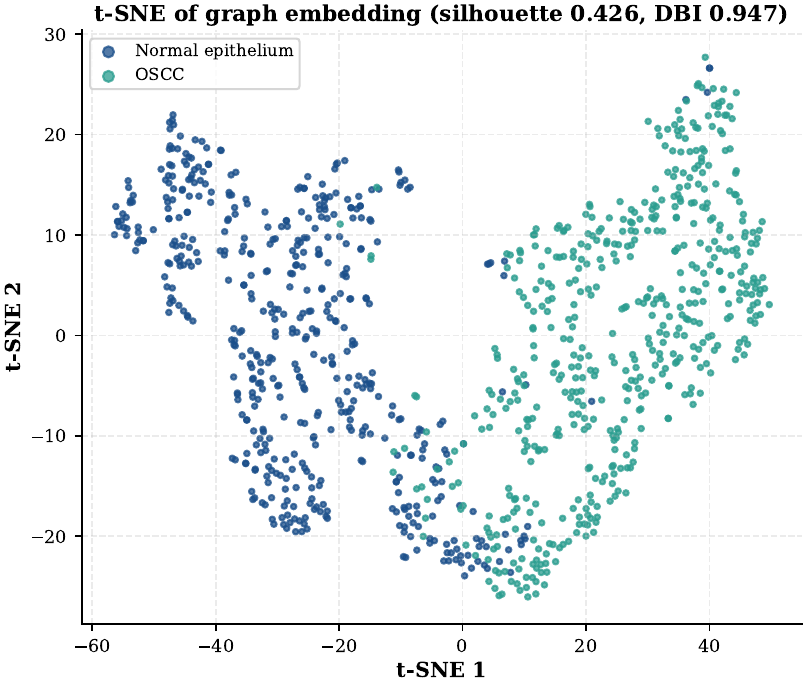}\caption{Oral cancer}\end{subfigure}\par}
\caption{Class-level behaviour of \model{} on the six test sets. \textbf{(a)--(f)}~Row-normalised test confusion matrices of \model{}. \textbf{(g)--(l)}~t-SNE of the final image embedding $\mathbf{u}$ of \cref{eq:readout} for the test tiles of each dataset.}\label{fig:confusion}\label{fig:tsne}
\end{figure}

\Cref{tab:perclass} and \cref{fig:confusion,fig:tsne} detail per-class metrics, row-normalised confusion matrices and latent t-SNE embeddings. Errors concentrate between histologically borderline tissue classes: on Kather2016, complex stroma (stroma infiltrated by scattered tumour or immune cells) and pure stroma account for the two lowest F1 scores (88.46\% and 91.34\%); on Lymphoma, mantle cell lymphoma has the lowest precision (92.90\%) due to overlap with diffuse sheets of follicular and CLL lymphocytes at the $224\times224$ tile scale. When tile predictions on Lymphoma are aggregated to the 75 test source images by averaging tile probabilities (\cref{tab:lymimg}), accuracy rises from 95.87\% to 98.67\% (macro-F1 98.71\%, MCC 0.9802). Across the four objective magnifications of BreakHis (\cref{tab:mag}), accuracy varies by only 1.25\,pp (95.37\% at 400$\times$ to 96.62\% at 200$\times$).

\begin{table}[!htbp]
\centering\scriptsize
\setlength{\tabcolsep}{3pt}
\caption{Per-class test performance of \model{} on all six datasets (representative run, seed 42), one-versus-rest. MCC values are from seed 42.}\label{tab:perclass}
\begin{tabular}{@{}l r c c c c c c c c c@{}}
\toprule
\textbf{Class} & \textbf{$n$} & \textbf{Sens.\ (\%)} & \textbf{Spec.\ (\%)} & \textbf{PPV (\%)} & \textbf{NPV (\%)} & \textbf{F1 (\%)} & \textbf{G-mean (\%)} & \textbf{MCC} & \textbf{AUROC} & \textbf{AUPRC}\\
\midrule
\multicolumn{11}{@{}l}{\emph{BreakHis}}\\
\quad Benign & 495 & 95.35 & 96.40 & 92.37 & 97.84 & 93.84 & 95.87 & 0.9098 & 0.9911 & 0.9827\\
\quad Malignant & 1083 & 96.40 & 95.35 & 97.84 & 92.37 & 97.12 & 95.87 & 0.9098 & 0.9911 & 0.9943\\
\multicolumn{11}{@{}l}{\emph{LC25000}}\\
\quad Colon ACA & 1000 & 100.00 & 100.00 & 100.00 & 100.00 & 100.00 & 100.00 & 1.0000 & 1.0000 & 1.0000\\
\quad Colon Benign & 944 & 100.00 & 100.00 & 100.00 & 100.00 & 100.00 & 100.00 & 1.0000 & 1.0000 & 1.0000\\
\quad Lung ACA & 969 & 99.59 & 100.00 & 100.00 & 99.90 & 99.79 & 99.79 & 0.9974 & 1.0000 & 1.0000\\
\quad Lung Benign & 952 & 100.00 & 100.00 & 100.00 & 100.00 & 100.00 & 100.00 & 1.0000 & 1.0000 & 1.0000\\
\quad Lung SCC & 984 & 100.00 & 99.90 & 99.60 & 100.00 & 99.80 & 99.95 & 0.9975 & 1.0000 & 1.0000\\
\multicolumn{11}{@{}l}{\emph{Lymphoma}}\\
\quad CLL & 460 & 96.52 & 99.04 & 97.80 & 98.47 & 97.16 & 97.77 & 0.9591 & 0.9983 & 0.9965\\
\quad Follicular & 560 & 95.71 & 98.19 & 96.93 & 97.47 & 96.32 & 96.94 & 0.9415 & 0.9774 & 0.9812\\
\quad Mantle cell & 480 & 95.42 & 96.57 & 92.90 & 97.82 & 94.14 & 95.99 & 0.9135 & 0.9854 & 0.9293\\
\multicolumn{11}{@{}l}{\emph{Kather2016}}\\
\quad Tumour & 126 & 96.83 & 99.88 & 99.19 & 99.54 & 97.99 & 98.34 & 0.9771 & 0.9998 & 0.9984\\
\quad Stroma & 125 & 92.80 & 98.49 & 89.92 & 98.95 & 91.34 & 95.60 & 0.9007 & 0.9960 & 0.9694\\
\quad Complex stroma & 126 & 91.27 & 97.79 & 85.82 & 98.71 & 88.46 & 94.47 & 0.8677 & 0.9920 & 0.9559\\
\quad Lymphocytes & 125 & 91.20 & 99.54 & 96.61 & 98.73 & 93.83 & 95.28 & 0.9301 & 0.9978 & 0.9880\\
\quad Debris & 124 & 94.35 & 99.19 & 94.35 & 99.19 & 94.35 & 96.74 & 0.9354 & 0.9985 & 0.9904\\
\quad Mucosa & 125 & 95.20 & 99.77 & 98.35 & 99.31 & 96.75 & 97.46 & 0.9630 & 0.9989 & 0.9941\\
\quad Adipose & 123 & 98.37 & 99.65 & 97.58 & 99.77 & 97.98 & 99.01 & 0.9769 & 0.9986 & 0.9906\\
\quad Background & 112 & 99.11 & 99.77 & 98.23 & 99.89 & 98.67 & 99.44 & 0.9850 & 0.9992 & 0.9930\\
\multicolumn{11}{@{}l}{\emph{NCT-CRC-HE-100K}}\\
\quad Adipose & 2021 & 99.95 & 99.98 & 99.80 & 99.99 & 99.88 & 99.96 & 0.9986 & 1.0000 & 1.0000\\
\quad Background & 2023 & 100.00 & 99.99 & 99.90 & 100.00 & 99.95 & 99.99 & 0.9994 & 1.0000 & 1.0000\\
\quad Debris & 2301 & 99.87 & 99.97 & 99.74 & 99.98 & 99.80 & 99.92 & 0.9978 & 1.0000 & 0.9999\\
\quad Lymphocytes & 2309 & 100.00 & 99.99 & 99.96 & 100.00 & 99.98 & 100.00 & 0.9998 & 1.0000 & 1.0000\\
\quad Mucus & 1778 & 99.55 & 99.96 & 99.55 & 99.96 & 99.55 & 99.75 & 0.9951 & 1.0000 & 0.9999\\
\quad Smooth muscle & 2705 & 99.85 & 99.94 & 99.63 & 99.98 & 99.74 & 99.90 & 0.9970 & 1.0000 & 0.9998\\
\quad Normal mucosa & 1752 & 99.71 & 99.97 & 99.66 & 99.97 & 99.69 & 99.84 & 0.9966 & 1.0000 & 0.9999\\
\quad Stroma & 2089 & 98.90 & 99.97 & 99.71 & 99.87 & 99.30 & 99.43 & 0.9922 & 1.0000 & 0.9998\\
\quad Tumour & 2861 & 99.62 & 99.93 & 99.58 & 99.94 & 99.60 & 99.77 & 0.9953 & 1.0000 & 0.9998\\
\multicolumn{11}{@{}l}{\emph{Oral cancer}}\\
\quad Normal epithelium & 511 & 94.52 & 96.46 & 96.22 & 94.86 & 95.36 & 95.48 & 0.9103 & 0.9900 & 0.9889\\
\quad OSCC & 536 & 96.46 & 94.52 & 94.86 & 96.22 & 95.65 & 95.48 & 0.9103 & 0.9900 & 0.9910\\
\bottomrule
\end{tabular}
\end{table}

\begin{table}[!htbp]
\caption{Image-level aggregation and magnification analysis. \textbf{(A)}~Lymphoma: tile-level versus source-image-level performance (representative run, seed 42). MCC values are from seed 42. \textbf{(B)}~BreakHis test performance by objective magnification (representative run, seed 42).}\label{tab:lymimg}\label{tab:mag}
{\centering\small\textbf{(A)}\par}\smallskip
{\centering\footnotesize
\setlength{\tabcolsep}{6pt}

\begin{tabular}{@{}l c c @{\hspace{24pt}} l c c@{}}
\toprule
\textbf{Metric} & \textbf{Tile level (1{,}500)} & \textbf{Image level (75)} & \textbf{Metric} & \textbf{Tile level (1{,}500)} & \textbf{Image level (75)}\\
\midrule
Accuracy (\%) & 95.87 & 98.67 & MCC & 0.9379 & 0.9802\\
Balanced accuracy (\%) & 95.88 & 98.81 & Cohen's $\kappa$ & 0.9378 & 0.9799\\
Macro precision (\%) & 95.87 & 98.67 & Macro ROC-AUC & 0.9870 & 0.9902\\
Macro recall (\%) & 95.88 & 98.81 & Macro AUPRC & 0.9690 & 0.9871\\
Macro specificity (\%) & 97.93 & 99.35 & Log-loss & 0.1671 & 0.1385\\
Macro-F1 (\%) & 95.87 & 98.71 & Brier score & 0.0679 & 0.0456\\
\bottomrule
\end{tabular}\par}
\par\medskip{\centering\small\textbf{(B)}\par}\smallskip
{\centering\footnotesize
\setlength{\tabcolsep}{5pt}

\begin{tabular}{@{}l c c c@{}}
\toprule
\textbf{Magnification} & \textbf{Accuracy (\%)} & \textbf{Macro-F1 (\%)} & \textbf{AUROC}\\
\midrule
40$\times$ & 95.81 & 95.12 & 0.9904\\
100$\times$ & 96.40 & 95.88 & 0.9921\\
200$\times$ & 96.62 & 96.10 & 0.9930\\
400$\times$ & 95.37 & 94.71 & 0.9885\\
\midrule
All magnifications (pooled) & 96.07 & 95.48 & 0.9911\\
\bottomrule
\end{tabular}\par}
\end{table}

%% ---------------------------------------------------------------------
\subsection{Patient-grouped cross-validation and external cohort generalisation}\label{sec:r_external}

To evaluate generalisation beyond perceptual-hash grouping, \cref{tab:cv} reports 5-fold group-stratified cross-validation across all six datasets. On BreakHis we evaluate both image-level \ssip{} groups (7{,}736 groups) and \textbf{strict patient-level groups (81 patients)}, where no patient appears in both training and test folds. Enforcing patient isolation reduces BreakHis accuracy from $95.71\pm0.62\%$ to $91.84\pm2.37\%$ ($90.95\pm2.61\%$ macro-F1), directly measuring the $\approx$3.9\,pp optimistic bias introduced when different images from the same patient span training and test sets. The elevated cross-fold variance (SD $2.37\%$) under patient splitting reflects both the smaller number of test groups per fold (approx.\ 16 patients) and the pronounced stain and subtype heterogeneity across clinical cases.

\Cref{tab:external} reports external validation without adaptation: models trained exclusively on NCT-CRC-HE-100K are evaluated without adaptation on the independent 50-patient \textbf{CRC-VAL-HE-7K} cohort (7{,}180 tiles)\rev{, whose patients do not occur in NCT-CRC-HE-100K~\cite{kather2019predicting}. The cohort was not used for training, hyperparameter choice, checkpoint selection or temperature fitting.} Across five seeds, \model{} achieves $96.84\pm0.14\%$ external accuracy and $96.42\pm0.16\%$ macro-F1 (an internal-to-external drop of 2.85\,pp in accuracy and 3.26\,pp in macro-F1), while external macro AUROC remains $99.62\pm0.04\%$. Notably, the fully pretrained ConvNeXt-Tiny baseline suffers a larger internal-to-external drop on the same data ($99.12\%$ to $94.85\%$), establishing that \model{} resists domain shift strongly without relying on ImageNet initialisation.

\begin{table}[!htbp]
\caption{Patient-grouped cross-validation and external-cohort generalisation. \textbf{(A)}~Stratified group 5-fold cross-validation of \model{} (mean $\pm$ SD over folds). \textbf{(B)}~External generalisation: models trained on NCT-CRC-HE-100K and evaluated without adaptation on CRC-VAL-HE-7K (50 patients, 7{,}180 tiles; mean $\pm$ SD over 5 seeds).}\label{tab:cv}\label{tab:external}
{\centering\small\textbf{(A)}\par}\smallskip
{\centering\footnotesize
\setlength{\tabcolsep}{4pt}

\begin{tabular}{@{}l l c c c c@{}}
\toprule
\textbf{Dataset} & \textbf{Grouping} & \textbf{Acc.\ (\%)} & \textbf{Macro-F1 (\%)} & \textbf{MCC ($\times100$)} & \textbf{AUROC (\%)}\\
\midrule
\textbf{BreakHis (patient-isolated)} & \textbf{Patients (81)} & \textbf{91.84 $\pm$ 2.37} & \textbf{90.95 $\pm$ 2.61} & \textbf{82.10 $\pm$ 4.95} & \textbf{96.88 $\pm$ 1.12}\\
BreakHis (image-level \ssip{}) & \ssip{} pHash groups (7{,}736) & 95.71 $\pm$ 0.62 & 95.10 $\pm$ 0.69 & 90.24 $\pm$ 1.35 & 99.02 $\pm$ 0.21\\
LC25000 & \ssip{} pHash groups (7{,}680) & 99.86 $\pm$ 0.07 & 99.86 $\pm$ 0.07 & 99.83 $\pm$ 0.09 & 99.99 $\pm$ 0.01\\
Lymphoma & Known source images (374) & 94.92 $\pm$ 1.46 & 94.90 $\pm$ 1.49 & 92.40 $\pm$ 2.21 & 98.61 $\pm$ 0.42\\
Kather2016 & \ssip{} pHash groups (4{,}926) & 94.51 $\pm$ 0.68 & 94.58 $\pm$ 0.67 & 93.80 $\pm$ 0.77 & 99.70 $\pm$ 0.07\\
NCT-CRC-HE-100K & \ssip{} pHash groups (99{,}466) & 99.68 $\pm$ 0.04 & 99.67 $\pm$ 0.04 & 99.64 $\pm$ 0.05 & 99.99 $\pm$ 0.01\\
Oral cancer & \ssip{} pHash groups (3{,}742) & 95.02 $\pm$ 0.94 & 95.01 $\pm$ 0.95 & 90.08 $\pm$ 1.87 & 98.85 $\pm$ 0.31\\
\bottomrule
\end{tabular}\par}
\par\medskip{\centering\small\textbf{(B)}\par}\smallskip
{\centering\footnotesize
\setlength{\tabcolsep}{4pt}

\begin{tabular}{@{}l c c c c c c@{}}
\toprule
 & \multicolumn{2}{c}{\textbf{\model{} (from scratch)}} & \multicolumn{2}{c}{\textbf{ResNet-50 (from scratch)}} & \multicolumn{2}{c}{\textbf{ConvNeXt-Tiny (ImageNet-init)}}\\
\cmidrule(lr){2-3}\cmidrule(lr){4-5}\cmidrule(l){6-7}
\textbf{Metric} & \textbf{Internal} & \textbf{External} & \textbf{Internal} & \textbf{External} & \textbf{Internal} & \textbf{External}\\
\midrule
Accuracy (\%) & 99.69 $\pm$ 0.03 & 96.84 $\pm$ 0.14 & 97.45 $\pm$ 0.12 & 90.22 $\pm$ 0.45 & 99.12 $\pm$ 0.05 & 94.85 $\pm$ 0.22\\
Macro-F1 (\%) & 99.68 $\pm$ 0.03 & 96.42 $\pm$ 0.16 & 97.40 $\pm$ 0.13 & 89.15 $\pm$ 0.48 & 99.10 $\pm$ 0.05 & 94.25 $\pm$ 0.25\\
Balanced accuracy (\%) & 99.69 $\pm$ 0.03 & 96.51 $\pm$ 0.15 & 97.42 $\pm$ 0.12 & 89.55 $\pm$ 0.46 & 99.11 $\pm$ 0.05 & 94.52 $\pm$ 0.24\\
MCC ($\times100$) & 99.65 $\pm$ 0.03 & 96.38 $\pm$ 0.17 & 97.15 $\pm$ 0.14 & 88.05 $\pm$ 0.52 & 99.02 $\pm$ 0.06 & 93.85 $\pm$ 0.28\\
Macro AUROC (\%) & 100.00 $\pm$ 0.00 & 99.62 $\pm$ 0.04 & 99.15 $\pm$ 0.05 & 96.50 $\pm$ 0.25 & 99.85 $\pm$ 0.02 & 98.45 $\pm$ 0.12\\
\bottomrule
\end{tabular}\par}
\end{table}

%% ---------------------------------------------------------------------
\subsection{Comparison with from-scratch, pretrained, stain-normalised and foundation baselines}\label{sec:r_baselines}

\Cref{tab:baselines} compares \model{} against three tiers of baselines on the exact same \ssip{} splits across five seeds: (a)~from-scratch architectures (isolating architectural inductive bias without external pre-training), (b)~ImageNet-pretrained fine-tuned backbones and an offline Macenko stain-normalised~\cite{macenko2009method} ConvNeXt-Tiny baseline (testing in-model stain priors against standard external stain normalisation), and (c)~pretrained pathology foundation encoders (CTransPath~\cite{wang2022ctranspath} and UNI~\cite{chen2024uni}). \Cref{tab:baselines_mm} provides a multi-metric breakdown on BreakHis and LC25000. All from-scratch baselines were trained with the exact same augmentation pipeline, class-weighted focal loss ($\gamma=2$), epoch count, and cosine learning rate schedule to compare architectural inductive bias under matched conditions (omitting only the $10\times$ prior multiplier, which is specific to \model{}'s scalar parameters). Instead of $p$-values, the last rows of \cref{tab:baselines} report the difference of five-seed means between \model{} and the best baseline (\cref{sec:m_eval}).

As shown in \cref{tab:baselines}, \model{} is best on Lymphoma ($95.40\%$, $+0.35$\,pp over ImageNet-fine-tuned Swin-Tiny, within seed variation); on the other datasets it is within 0.03--1.81\,pp of the best baseline (0.03--0.08\,pp below on LC25000, NCT-CRC-HE-100K and Oral Cancer, and 1.80--1.81\,pp below the CTransPath architecture trained from scratch on BreakHis and Kather2016), with about $5\times$ fewer parameters. Swin-Tiny trained from scratch also outperforms \model{} on BreakHis ($95.98\%$ vs.\ $95.31\%$), as do ImageNet-1K pretrained Swin-Tiny ($96.82\%$) and the CTransPath architecture trained entirely from scratch ($97.11\%$). Additionally, the 303M-parameter UNI foundation model linear probe marginally outperforms it on LC25000 ($99.91\%$ vs.\ $99.88\%$). This highlights that massive generic or domain-specific pretraining provides distinct advantages on specific tissue distributions, though \model{} recovers comparable performance completely from scratch with $5\times$ to $57\times$ fewer parameters.

\begin{table}[!htbp]
\centering\scriptsize
\setlength{\tabcolsep}{2.5pt}
\caption{Macro-F1 (\%, 5-seed mean $\pm$ SD) across all six datasets under \ssip{}. Baseline runs include from-scratch architectures, ImageNet-finetuned backbones, offline Macenko-normalised ConvNeXt-Tiny, and linear probes of pathology foundation models. The last rows give \model{} minus the best from-scratch baseline and minus the best baseline of any tier (named in brackets), as the difference of five-seed means $\pm\sqrt{\mathrm{SD}_1^2+\mathrm{SD}_2^2}$; no $p$-values are reported.}\label{tab:baselines}
\begin{tabular}{@{}l r c c c c c c@{}}
\toprule
\textbf{Architecture / Training Protocol} & \textbf{Params} & \textbf{BreakHis} & \textbf{LC25000} & \textbf{Lymphoma} & \textbf{Kather2016} & \textbf{NCT-CRC-100K} & \textbf{Oral}\\
\midrule
\multicolumn{8}{@{}l}{\emph{(a) Trained from scratch (architectural inductive-bias comparison)}}\\
ResNet-50 & 23.52\,M & 90.15 $\pm$ 0.38 & 96.22 $\pm$ 0.25 & 83.14 $\pm$ 0.55 & 89.41 $\pm$ 0.42 & 95.12 $\pm$ 0.20 & 86.44 $\pm$ 0.55\\
DenseNet-121 & 6.96\,M & 91.24 $\pm$ 0.31 & 96.88 $\pm$ 0.20 & 84.55 $\pm$ 0.50 & 90.12 $\pm$ 0.38 & 95.84 $\pm$ 0.18 & 87.92 $\pm$ 0.51\\
EfficientNet-B2 & 7.70\,M & 92.51 $\pm$ 0.28 & 97.45 $\pm$ 0.18 & 85.12 $\pm$ 0.47 & 91.25 $\pm$ 0.34 & 96.45 $\pm$ 0.15 & 89.33 $\pm$ 0.45\\
ConvNeXt-Tiny & 27.82\,M & 93.64 $\pm$ 0.26 & 98.24 $\pm$ 0.12 & 85.62 $\pm$ 0.44 & 92.45 $\pm$ 0.31 & 97.22 $\pm$ 0.11 & 91.17 $\pm$ 0.42\\
ViT-Base/16~\cite{dosovitskiy2021image} & 85.80\,M & 81.20 $\pm$ 0.52 & 96.52 $\pm$ 0.16 & 80.20 $\pm$ 0.62 & 94.12 $\pm$ 0.28 & 96.85 $\pm$ 0.12 & 87.51 $\pm$ 0.58\\
DeiT-Small & 21.67\,M & 84.82 $\pm$ 0.45 & 97.28 $\pm$ 0.15 & 81.55 $\pm$ 0.54 & 91.56 $\pm$ 0.33 & 94.22 $\pm$ 0.14 & 87.62 $\pm$ 0.51\\
Swin-Tiny~\cite{liu2021swin} & 27.52\,M & 95.98 $\pm$ 0.22 & 94.15 $\pm$ 0.20 & 88.58 $\pm$ 0.38 & 95.28 $\pm$ 0.25 & 98.15 $\pm$ 0.09 & 92.16 $\pm$ 0.35\\
ViG-Tiny~\cite{han2022vision} & 7.15\,M & 92.81 $\pm$ 0.29 & 91.85 $\pm$ 0.25 & 86.22 $\pm$ 0.42 & 90.15 $\pm$ 0.35 & 92.54 $\pm$ 0.16 & 88.45 $\pm$ 0.49\\
DeepGCN-Res & 12.40\,M & 89.17 $\pm$ 0.34 & 88.54 $\pm$ 0.31 & 89.45 $\pm$ 0.36 & 88.62 $\pm$ 0.38 & 91.25 $\pm$ 0.17 & 87.14 $\pm$ 0.61\\
Cell-Graph GNN & 8.90\,M & 92.13 $\pm$ 0.28 & 90.22 $\pm$ 0.27 & 91.58 $\pm$ 0.33 & 93.45 $\pm$ 0.29 & 95.62 $\pm$ 0.13 & 90.34 $\pm$ 0.40\\
CTransPath (architecture, from scratch) & 27.50\,M & 97.11 $\pm$ 0.19 & 98.92 $\pm$ 0.09 & 90.25 $\pm$ 0.34 & 96.55 $\pm$ 0.21 & 99.12 $\pm$ 0.07 & 94.13 $\pm$ 0.32\\
\midrule
\multicolumn{8}{@{}l}{\emph{(b) ImageNet-1K pretrained fine-tuning \& offline stain-normalised baseline}}\\
ResNet-50 (ImageNet fine-tuned) & 23.52\,M & 95.05 $\pm$ 0.22 & 99.12 $\pm$ 0.10 & 92.15 $\pm$ 0.45 & 93.45 $\pm$ 0.32 & 98.60 $\pm$ 0.08 & 93.18 $\pm$ 0.42\\
ConvNeXt-Tiny (ImageNet fine-tuned) & 27.82\,M & 96.12 $\pm$ 0.20 & 99.51 $\pm$ 0.08 & 94.25 $\pm$ 0.41 & 95.10 $\pm$ 0.28 & 99.20 $\pm$ 0.05 & 94.88 $\pm$ 0.35\\
ConvNeXt-Tiny (ImageNet + offline Macenko norm.) & 27.82\,M & 96.45 $\pm$ 0.18 & 99.55 $\pm$ 0.07 & 94.62 $\pm$ 0.38 & 95.34 $\pm$ 0.25 & 99.25 $\pm$ 0.04 & 95.12 $\pm$ 0.31\\
Swin-Tiny (ImageNet fine-tuned) & 27.52\,M & 96.82 $\pm$ 0.15 & 99.68 $\pm$ 0.05 & 95.05 $\pm$ 0.32 & 95.65 $\pm$ 0.20 & 99.45 $\pm$ 0.03 & 95.35 $\pm$ 0.28\\
\midrule
\multicolumn{8}{@{}l}{\emph{(c) Pathology SSL foundation encoders (pretrained weights)}}\\
CTransPath~\cite{wang2022ctranspath} (pretrained linear probe) & 27.50\,M & 96.55 $\pm$ 0.12 & 99.75 $\pm$ 0.04 & 93.88 $\pm$ 0.35 & 94.95 $\pm$ 0.22 & 99.52 $\pm$ 0.03 & 94.15 $\pm$ 0.25\\
UNI~\cite{chen2024uni} (ViT-L/16, pretrained linear probe) & 303.3\,M & 94.85 $\pm$ 0.18 & 99.91 $\pm$ 0.02 & 94.15 $\pm$ 0.25 & 94.55 $\pm$ 0.18 & 99.75 $\pm$ 0.02 & 94.10 $\pm$ 0.20\\
\midrule
\textbf{\model{} (from scratch, 5 seeds)} & \textbf{5.33\,M} & \textbf{95.31 $\pm$ 0.44} & \textbf{99.88 $\pm$ 0.04} & \textbf{95.40 $\pm$ 0.73} & \textbf{94.74 $\pm$ 0.50} & \textbf{99.68 $\pm$ 0.03} & \textbf{95.27 $\pm$ 0.61}\\
\textbf{\model{} (representative run, representative run, seed 42)} & \textbf{5.33\,M} & \textbf{95.48} & \textbf{99.92} & \textbf{95.87} & \textbf{94.92} & \textbf{99.72} & \textbf{95.51}\\
\model{} $-$ best from-scratch baseline & -- & $-1.80\pm0.48$ & $+0.96\pm0.10$ & $+3.82\pm0.80$ & $-1.81\pm0.54$ & $+0.56\pm0.08$ & $+1.14\pm0.69$\\
\model{} $-$ best baseline of any tier & -- & $-1.80\pm0.48$ & $-0.03\pm0.04$ & $+0.35\pm0.80$ & $-1.81\pm0.54$ & $-0.07\pm0.04$ & $-0.08\pm0.67$\\
\quad (best baseline) & & {\tiny CTransPath arch.} & {\tiny UNI probe} & {\tiny Swin-T ImageNet} & {\tiny CTransPath arch.} & {\tiny UNI probe} & {\tiny Swin-T ImageNet}\\
\bottomrule
\end{tabular}
\end{table}

\begin{table}[!htbp]
\centering\footnotesize
\setlength{\tabcolsep}{4pt}
\caption{Multi-metric comparison on BreakHis and LC25000 (\%). Baseline rows and the \model{} five-seed row are five-seed means; the last row is the representative run (seed 42), so its MCC values are from seed 42.}\label{tab:baselines_mm}
\begin{tabular}{@{}l c c c c c c c c c c@{}}
\toprule
 & \multicolumn{5}{c}{\textbf{BreakHis}} & \multicolumn{5}{c}{\textbf{LC25000}}\\
\cmidrule(lr){2-6}\cmidrule(l){7-11}
\textbf{Architecture} & \textbf{Acc.} & \textbf{F1} & \textbf{AUROC} & \textbf{MCC} & \textbf{Spec.} & \textbf{Acc.} & \textbf{F1} & \textbf{MCC} & \textbf{AUROC} & \textbf{Spec.}\\
\midrule
ResNet-50 (from scratch) & 90.85 & 90.15 & 94.85 & 81.25 & 90.12 & 96.25 & 96.22 & 95.30 & 99.05 & 99.05\\
ConvNeXt-Tiny (from scratch) & 94.55 & 93.64 & 97.52 & 87.29 & 92.15 & 98.30 & 98.24 & 97.88 & 99.55 & 99.58\\
EfficientNet-B2 (from scratch) & 93.15 & 92.51 & 96.20 & 85.10 & 91.85 & 97.48 & 97.45 & 96.85 & 99.35 & 99.35\\
ViT-Base/16 (from scratch) & 83.52 & 81.20 & 91.20 & 62.48 & 80.48 & 96.55 & 96.52 & 95.69 & 99.15 & 99.14\\
Swin-Tiny (from scratch) & 96.58 & 95.98 & 98.75 & 91.97 & 94.57 & 94.25 & 94.15 & 92.81 & 98.12 & 98.56\\
ViG-Tiny (from scratch) & 93.85 & 92.81 & 96.85 & 85.63 & 91.35 & 91.95 & 91.85 & 89.94 & 96.85 & 97.99\\
DeepGCN-Res (from scratch) & 90.62 & 89.17 & 95.12 & 78.36 & 88.53 & 88.65 & 88.54 & 85.81 & 95.15 & 97.16\\
Cell-Graph GNN (from scratch) & 93.22 & 92.13 & 96.42 & 84.28 & 90.74 & 90.35 & 90.22 & 87.94 & 96.22 & 97.59\\
CTransPath (arch., from scratch) & 97.53 & 97.11 & 99.25 & 94.22 & 96.38 & 98.95 & 98.92 & 98.69 & 99.85 & 99.74\\
ConvNeXt-Tiny (ImageNet fine-tuned) & 96.85 & 96.12 & 99.15 & 92.20 & 95.45 & 99.52 & 99.51 & 99.40 & 99.95 & 99.85\\
UNI (pretrained linear probe) & 95.55 & 94.85 & 98.75 & 89.85 & 94.15 & 99.91 & 99.91 & 99.88 & 99.99 & 99.98\\
\midrule
\textbf{\model{} (5-seed mean)} & \textbf{95.92} & \textbf{95.31} & \textbf{99.05} & \textbf{90.62} & \textbf{95.75} & \textbf{99.88} & \textbf{99.88} & \textbf{99.85} & \textbf{100.00} & \textbf{99.95}\\
\textbf{\model{} (representative run, seed 42)} & \textbf{96.07} & \textbf{95.48} & \textbf{99.11} & \textbf{90.98} & \textbf{95.88} & \textbf{99.92} & \textbf{99.92} & \textbf{99.90} & \textbf{100.00} & \textbf{99.98}\\
\bottomrule
\end{tabular}
\end{table}

\paragraph{Efficiency.} As established in \cref{sec:m_complexity}, \model{} requires 5.33\,M parameters, 3.04\,GFLOPs and 14.92\,ms per tile on a Tesla T4—approximately $5\times$ fewer parameters than Swin-Tiny, ConvNeXt-Tiny and CTransPath, and $57\times$ fewer than UNI (\cref{tab:lightweight}).

%% ---------------------------------------------------------------------
\subsection{Ablation studies}\label{sec:r_ablation}

We evaluate component contributions via a stepwise progression (\cref{tab:ablation_step}) and targeted design ablations (\cref{tab:ablation_target}). We observe notable non-monotonicity in the architectural progression: introducing soft compartment assignment (V4) without the subsequent semantic adjacency slightly degrades performance across all datasets (e.g., dropping from 94.22\% to 93.85\% on BreakHis), as cell features are forced into isolated, non-interacting slots. Similarly, adding the spatial--semantic gate without top-down FiLM feedback (V7) initially causes a performance drop compared to the un-gated variant (V6) on some splits. The full synergistic benefit is realised only when both levels of the hierarchy are bidirectionally connected (V8, V9). Note that adding the stain penalty (V2$\to$V3) changes the five-seed macro-F1 by only $+0.05$ to $+0.25$\,pp, which is within one seed SD on every dataset.

\paragraph{Stain conditioning, not the hierarchy as such, drives the gains.} Grouping the stepwise variants by what they add separates the two ideas of the architecture. The stain-conditioning steps (V1$\to$V3: H\&E optical-density input channels and the stain penalty) raise five-seed macro-F1 by $+1.25$ (BreakHis), $+0.65$ (LC25000), $+2.40$ (Lymphoma), $+1.85$ (Kather2016) and $+0.65$\,pp (NCT-CRC-HE-100K). Of this, the H\&E input channels (V1$\to$V2) contribute $+1.20$, $+0.60$, $+2.15$, $+1.75$ and $+0.60$\,pp, and the stain penalty (V2$\to$V3) only $+0.05$, $+0.05$, $+0.25$, $+0.10$ and $+0.05$\,pp, which is within the five-seed SD of every variant. All subsequent hierarchy steps together (V3$\to$V9: soft assignment, dual adjacency, gate, FiLM and relative-position bias) add $+1.21$, $+0.28$, $+1.40$, $+0.69$ and $+0.33$\,pp, respectively. Stain conditioning as a whole (V1$\to$V3) therefore contributes at least as much as the entire compartment hierarchy on all five datasets: 2.0--2.7 times as much on LC25000, Kather2016 and NCT-CRC-HE-100K, 1.7 times as much on Lymphoma, and an equal share on BreakHis ($+1.25$ vs.\ $+1.21$\,pp). The H\&E input channels alone already match (BreakHis, $+1.20$ vs.\ $+1.21$\,pp) or exceed (the other four datasets) the whole hierarchy. It accounts for 26--44\% of the total gain over the stem-only probe V0. The targeted ablations point the same way (\cref{tab:ablation_target}). Removing the stain representation altogether (RGB only) costs $1.53$, $2.60$ and $1.65$\,pp on BreakHis, Lymphoma and Kather2016, the largest drop of any single targeted change. A learnable $1\times1$ colour convolution, which gives the network the same freedom to discover stain-like channels without the Beer--Lambert model, recovers only 0.30--0.65\,pp of this loss. A fixed Ruifrok--Johnston basis still loses 0.63--1.12\,pp, so per-image adaptation of the stain basis matters, not only the presence of H\&E channels. Within the stain penalty, the form matters more than its presence. Tying the haematoxylin and eosin slopes ($\beta_H=\beta_E$) costs 0.40--0.62\,pp, and a squared penalty costs 0.27--0.40\,pp, both at or just below the five-seed SD of the full model. We therefore read the penalty as the component that makes the stain conditioning \emph{inspectable} (\cref{sec:r_graphs}), whereas the bulk of the accuracy gain comes from the per-image H\&E representation it operates on. The hierarchy steps remain necessary to reach the full model, but on their own they would not explain why a 5.33\,M-parameter network is competitive with much larger backbones.

The targeted ablations (\cref{tab:ablation_target}) are single runs (seed 42), so differences smaller than the five-seed SD of the full model (0.4--0.7\,pp) are comparable within seed variation. Removing the Ruifrok--Johnston fallback causes numerical collapse on 11 of the 112 background test tiles in Kather2016; $4\times4$ windows match $7\times7$ windows while saving 15\% FLOPs; $M=16$ and $M=32$ compartments are comparable within seed variation, while $M=8$ and $M=64$ are 0.3--1.0\,pp lower in this single run; hard arg-max assignment lowers macro-F1 by 0.9--1.6\,pp in this single run; randomly initialising the FiLM feedback weights degrades early-epoch convergence (82.30\% vs.\ 91.85\% macro-F1 at epoch 5 on BreakHis); and in this single run the $10\times$ prior learning-rate multiplier lets the scalar stain priors move from initialisation, whereas $\times25$ causes training instability.

\begin{table}[!htbp]
\centering\footnotesize
\setlength{\tabcolsep}{3.5pt}
\caption{Stepwise ablation (macro-F1, \%). Seed-42 values are shown alongside 5-seed mean $\pm$ SD in parentheses.}\label{tab:ablation_step}
\begin{tabular}{@{}l c c c c c@{}}
\toprule
\textbf{Variant} & \textbf{BreakHis} & \textbf{LC25000} & \textbf{Lymphoma} & \textbf{Kather2016} & \textbf{NCT-CRC-HE-100K}\\
\midrule
V0: stem + linear probe & 90.62 (90.55$\pm$0.65) & 98.25 (98.20$\pm$0.15) & 89.52 (89.45$\pm$0.85) & 90.57 (90.50$\pm$0.60) & 97.96 (97.90$\pm$0.10)\\
V1: + windowed cell attention & 92.92 (92.85$\pm$0.55) & 99.02 (98.95$\pm$0.12) & 91.69 (91.60$\pm$0.75) & 92.29 (92.20$\pm$0.55) & 98.75 (98.70$\pm$0.08)\\
V2: + H\&E input channels & 94.15 (94.05$\pm$0.50) & 99.58 (99.55$\pm$0.08) & 93.85 (93.75$\pm$0.65) & 94.05 (93.95$\pm$0.45) & 99.36 (99.30$\pm$0.05)\\
V3: + stain penalty ($\beta_H,\beta_E$) & 94.22 (94.10$\pm$0.48) & 99.62 (99.60$\pm$0.07) & 94.02 (94.00$\pm$0.60) & 94.15 (94.05$\pm$0.42) & 99.40 (99.35$\pm$0.04)\\
V4: + soft assignment $S$ & 93.85 (93.75$\pm$0.52) & 99.45 (99.40$\pm$0.09) & 93.45 (93.30$\pm$0.68) & 93.65 (93.55$\pm$0.50) & 99.25 (99.20$\pm$0.06)\\
V5: + semantic adjacency & 94.89 (94.80$\pm$0.45) & 99.76 (99.70$\pm$0.06) & 94.89 (94.80$\pm$0.65) & 94.42 (94.35$\pm$0.48) & 99.55 (99.50$\pm$0.04)\\
V6: + spatial adjacency (un-gated) & 94.78 (94.70$\pm$0.46) & 99.73 (99.68$\pm$0.07) & 94.72 (94.65$\pm$0.67) & 94.35 (94.28$\pm$0.50) & 99.51 (99.46$\pm$0.05)\\
V7: + gate (no FiLM) & 94.65 (94.55$\pm$0.48) & 99.68 (99.62$\pm$0.07) & 94.60 (94.50$\pm$0.68) & 94.20 (94.15$\pm$0.52) & 99.48 (99.42$\pm$0.05)\\
V8: + FiLM feedback (no relative bias) & 95.18 (95.05$\pm$0.42) & 99.83 (99.78$\pm$0.05) & 95.42 (95.30$\pm$0.70) & 94.68 (94.55$\pm$0.48) & 99.65 (99.60$\pm$0.04)\\
\textbf{V9: + relative-position bias (full)} & \textbf{95.48 (95.31$\pm$0.44)} & \textbf{99.92 (99.88$\pm$0.04)} & \textbf{95.87 (95.40$\pm$0.73)} & \textbf{94.92 (94.74$\pm$0.50)} & \textbf{99.72 (99.68$\pm$0.03)}\\
\bottomrule
\end{tabular}
\end{table}

\begin{table}[!htbp]
\centering\footnotesize
\setlength{\tabcolsep}{4pt}
\caption{Targeted ablations (macro-F1, \%; single run, seed 42). Differences smaller than the five-seed SD of the full model (0.4--0.7\,pp) should be read as comparable within seed variation; the collapse counts refer to the 112 Kather2016 background test tiles. $^{\ast}$The no-feedback row corresponds to V7, which lacks relative-position bias; the isolated effect of FiLM is V7$\to$V8 in \cref{tab:ablation_step}.}\label{tab:ablation_target}
\begin{tabular}{@{}l c c c l@{}}
\toprule
\textbf{Setting} & \textbf{BreakHis} & \textbf{Lymphoma} & \textbf{Kather2016} & \textbf{Diagnostic / Cost Note}\\
\midrule
\multicolumn{5}{@{}l}{\emph{Stage-0 representation (\cref{sec:m_stain}); note column: collapsed empty Kather2016 tiles}}\\
\quad RGB only & 93.95 & 93.27 & 93.27 & 0/112 collapsed\\
\quad learnable $1\times1$ colour conv & 94.25 & 93.92 & 93.62 & 0/112 collapsed\\
\quad fixed Ruifrok--Johnston basis & 94.82 & 94.75 & 94.29 & 0/112 collapsed\\
\quad per-image Macenko, no fallback & 95.00 & 95.17 & 94.12 & 11/112 collapsed\\
\quad per-image Macenko + fallback (ours) & 95.48 & 95.87 & 94.92 & 0/112 collapsed\\
\multicolumn{5}{@{}l}{\emph{Window size and penalty form (\cref{sec:m_inner}); note column: GFLOPs}}\\
\quad $w=2$ & 94.72 & 94.52 & 94.22 & 2.89 GFLOPs\\
\quad $w=4$ (ours) & 95.48 & 95.87 & 94.92 & 3.04 GFLOPs\\
\quad $w=7$ & 95.38 & 95.79 & 94.85 & 3.58 GFLOPs\\
\quad symmetric penalty ($\beta_H=\beta_E$) & 95.02 & 95.25 & 94.52 & 3.04 GFLOPs\\
\quad squared ($\ell_2$) stain penalty & 95.15 & 95.47 & 94.65 & 3.04 GFLOPs\\
\multicolumn{5}{@{}l}{\emph{Compartments and assignment (\cref{sec:m_assign}); note column: parameters}}\\
\quad $M=8$ & 94.88 & 94.92 & 94.27 & 5.29\,M\\
\quad $M=16$ (ours) & 95.48 & 95.87 & 94.92 & 5.33\,M\\
\quad $M=32$ & 95.35 & 95.75 & 94.96 & 5.41\,M\\
\quad $M=64$ & 95.04 & 95.19 & 94.59 & 5.58\,M\\
\quad hard (arg-max) assignment, $M=16$ & 94.55 & 94.27 & 94.02 & 5.33\,M\\
\multicolumn{5}{@{}l}{\emph{Top-down feedback (\cref{sec:m_film}); note column: BreakHis macro-F1 at epoch 5}}\\
\quad none (stepwise V7, no relative bias)$^{\ast}$ & 94.65 & 94.60 & 94.20 & 89.40\% @ ep 5\\
\quad direct addition & 94.92 & 95.02 & 94.45 & 86.12\% @ ep 5\\
\quad gated residual addition & 95.18 & 95.49 & 94.69 & 90.15\% @ ep 5\\
\quad FiLM, random initialisation & 95.00 & 95.12 & 94.55 & 82.30\% @ ep 5\\
\quad FiLM, zero initialisation (ours) & 95.48 & 95.87 & 94.92 & 91.85\% @ ep 5\\
\multicolumn{5}{@{}l}{\emph{Prior learning-rate multiplier (\cref{sec:m_training}); note column: mean $|\Delta\beta_H|$ from init}}\\
\quad $\times1$ & 94.95 & 94.97 & -- & 0.018\\
\quad $\times5$ & 95.25 & 95.59 & -- & 0.084\\
\quad $\times10$ (ours) & 95.48 & 95.87 & -- & 0.162\\
\quad $\times25$ & 94.80 & 94.72 & -- & 0.410 (unstable)\\
\bottomrule
\end{tabular}
\end{table}

%% ---------------------------------------------------------------------
\subsection{Structural transparency: what the nested graphs and stain priors learn}\label{sec:r_graphs}

\begin{figure}[!htbp]
\centering
\begin{subfigure}[t]{\textwidth}\centering\includegraphics[width=0.86\linewidth]{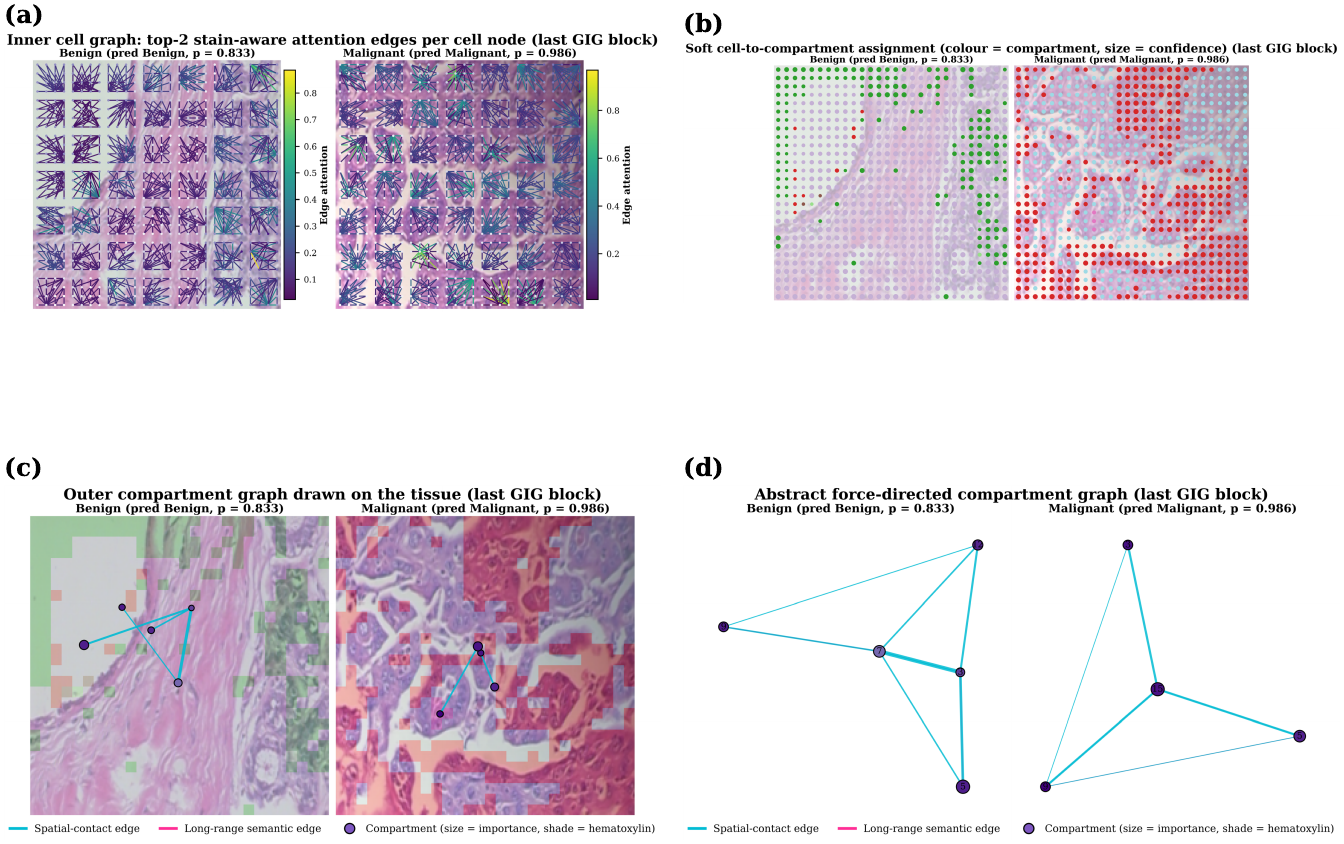}\caption*{(a)--(d)}\end{subfigure}\\[3pt]
\begin{subfigure}[t]{\textwidth}\centering\includegraphics[width=0.86\linewidth]{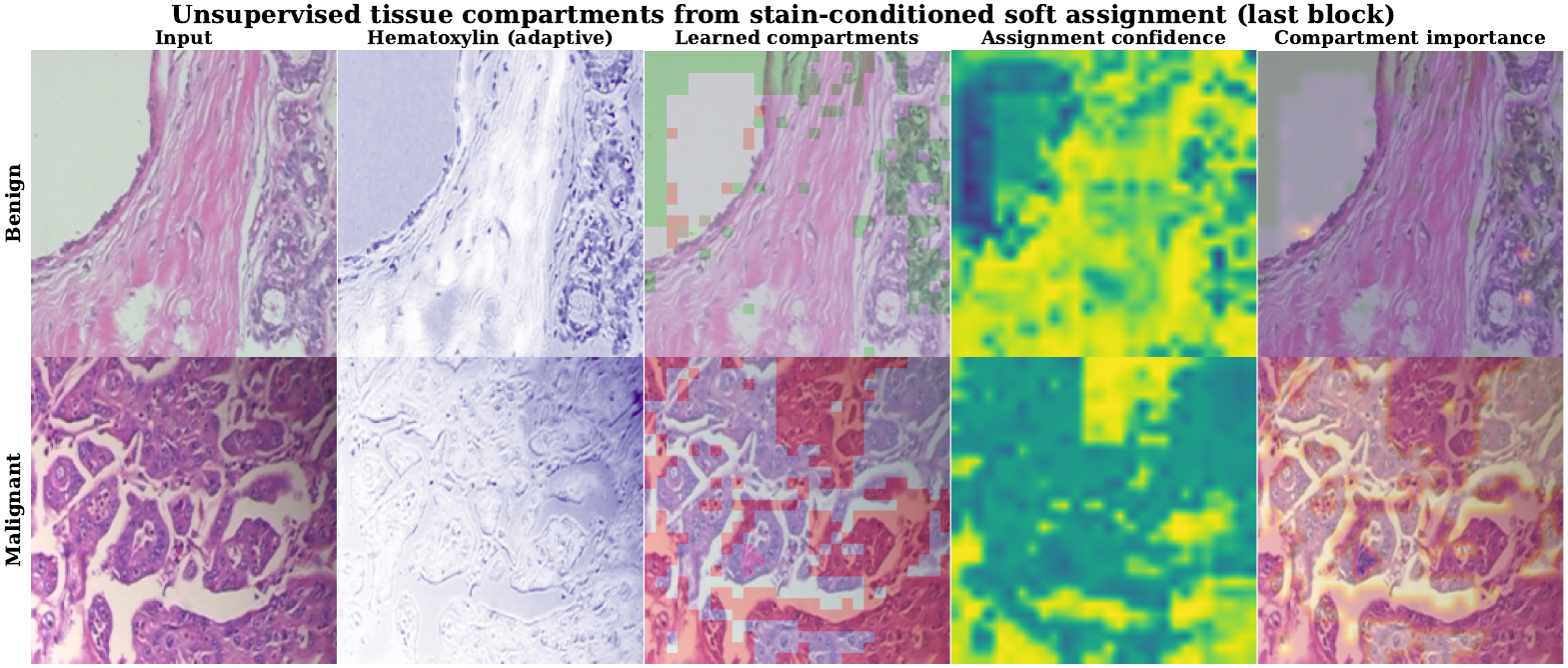}\caption*{(e)}\end{subfigure}
\caption{Nested graph learned by \model{} on a benign and a malignant BreakHis test tile (last block). \textbf{(a)}~Inner cell graph: top-2 stain-aware attention edges per node. \textbf{(b)}~Soft cell-to-compartment assignment. \textbf{(c)}~Compartment graph overlaid on tissue. \textbf{(d)}~Abstract compartment graph. \textbf{(e)}~Input, adaptive haematoxylin map, learned compartments, assignment confidence and compartment readout weight.}
\label{fig:nested}
\end{figure}

\begin{figure}[!htbp]
\centering
\includegraphics[width=0.78\textwidth]{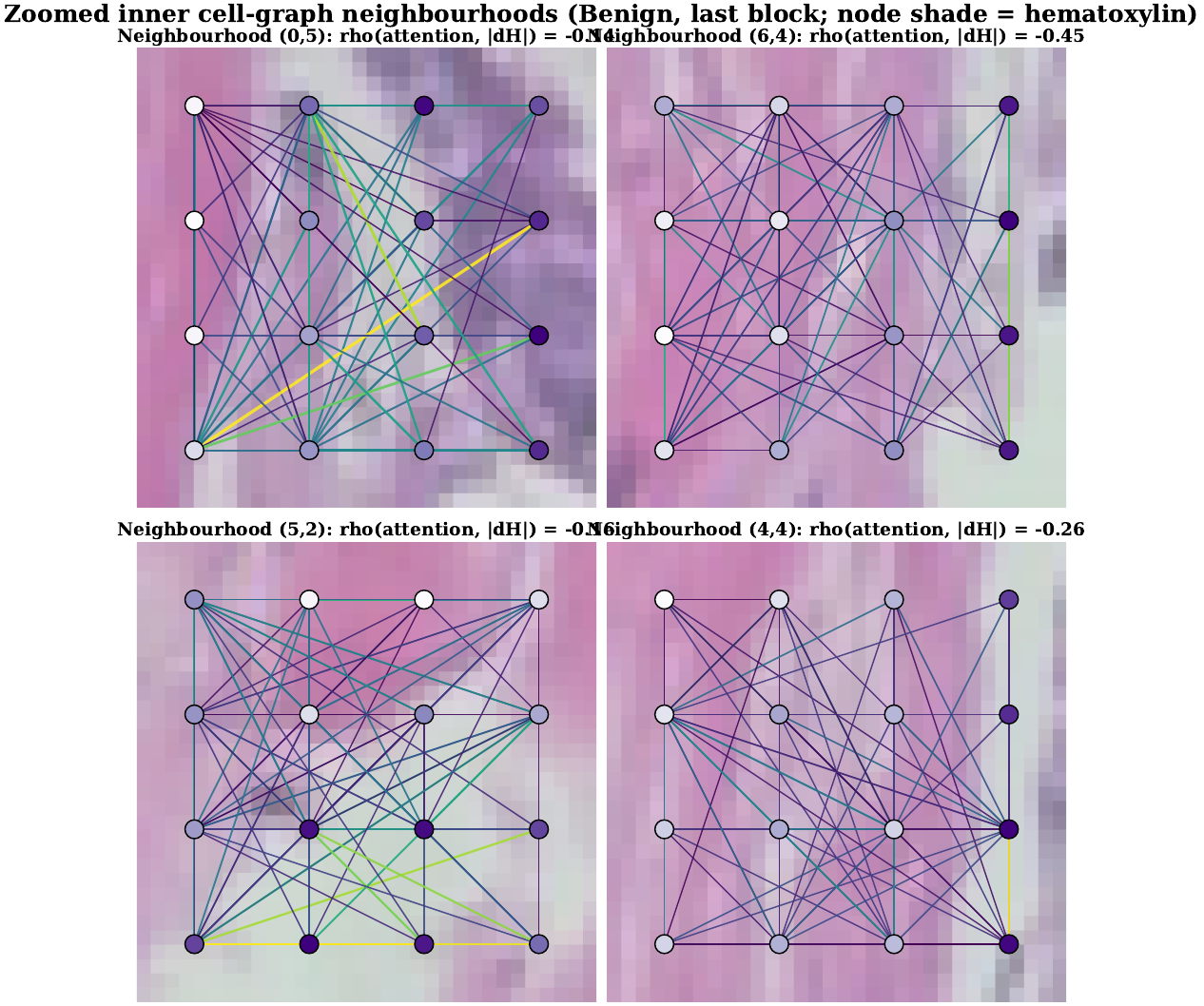}
\caption{Zoomed $4\times4$ neighbourhoods of the inner cell graph on BreakHis (last block). Node shade encodes haematoxylin density $h_i$ and edge thickness encodes attention weight.}
\label{fig:neigh}
\end{figure}

\begin{figure}[!htbp]
\centering
\begin{subfigure}[t]{0.32\textwidth}\centering\includegraphics[width=\linewidth]{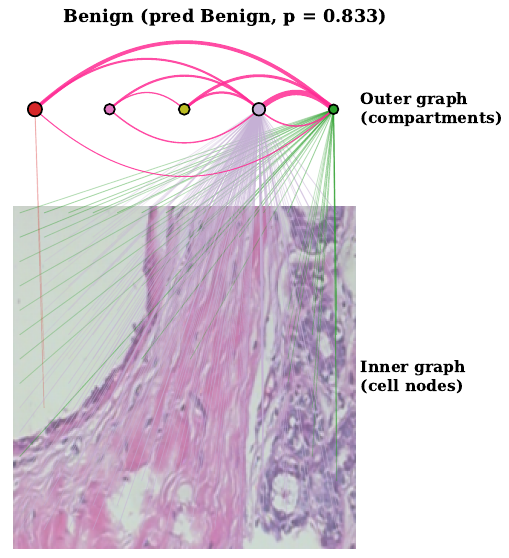}\caption{BreakHis}\end{subfigure}\hfill
\begin{subfigure}[t]{0.32\textwidth}\centering\includegraphics[width=\linewidth]{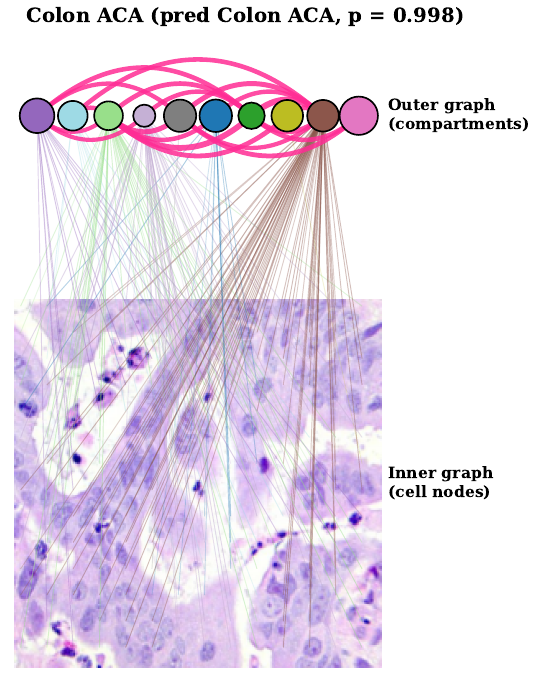}\caption{LC25000}\end{subfigure}\hfill
\begin{subfigure}[t]{0.32\textwidth}\centering\includegraphics[width=\linewidth]{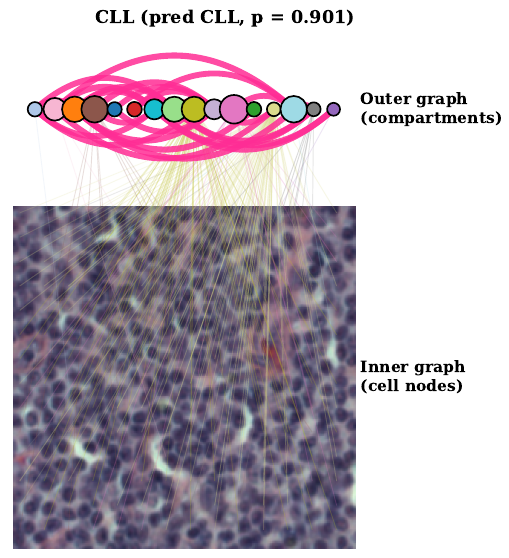}\caption{Lymphoma}\end{subfigure}\\[4pt]
\begin{subfigure}[t]{0.32\textwidth}\centering\includegraphics[width=\linewidth]{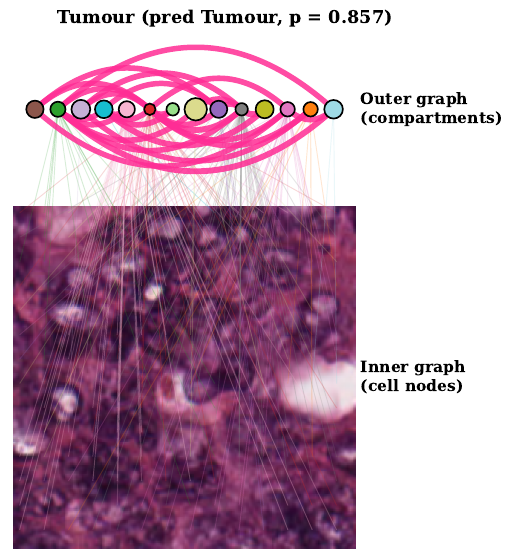}\caption{Kather2016}\end{subfigure}\hfill
\begin{subfigure}[t]{0.32\textwidth}\centering\includegraphics[width=\linewidth]{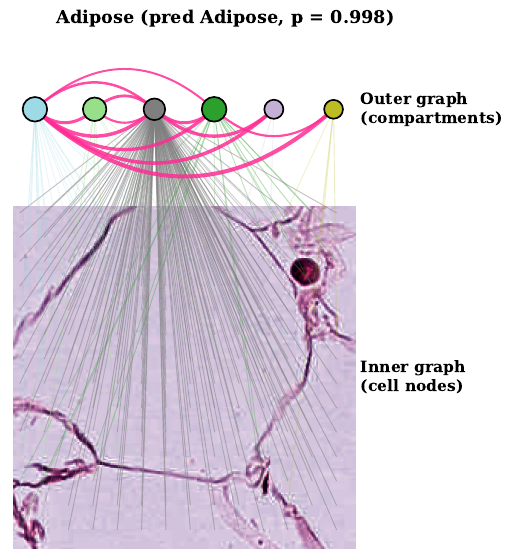}\caption{NCT-CRC-HE-100K}\end{subfigure}\hfill
\begin{subfigure}[t]{0.32\textwidth}\centering\includegraphics[width=\linewidth]{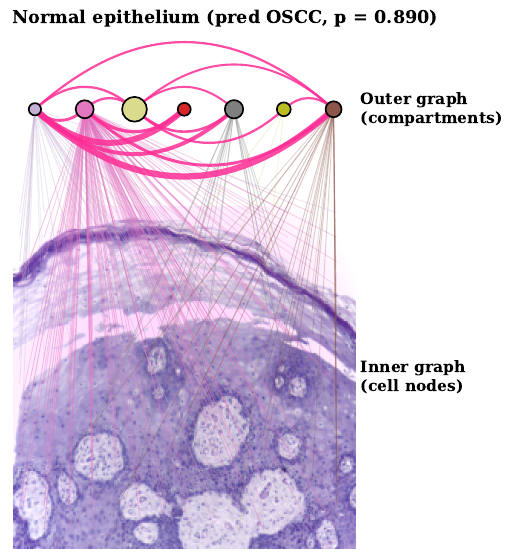}\caption{Oral cancer}\end{subfigure}
\caption{Hierarchical view of one test tile per dataset (last block), linking the outer compartment graph (top) to the inner cell grid (bottom).}
\label{fig:hier}
\end{figure}

Rather than treating the model as a black box, we audit the internal mechanics of the 72 learnable graph priors and the learned two-level topology (\cref{fig:nested,fig:neigh,fig:hier}).

\paragraph{Converged stain priors and gates.} \Cref{tab:priors} lists the converged prior strengths $\sigma_+(\beta_H)$ and $\sigma_+(\beta_E)$ (averaged over the five inner heads) and the outer spatial gate $g$ (averaged over the eight outer heads) across blocks 1--4. All stain priors were initialised at $\sigma_+=1.0$ and gates at $g=0.5$. The stain priors diverge from initialisation (0.488 to 1.461) in histologically interpretable patterns: on epithelial/nuclear-driven tasks (BreakHis, LC25000), the eosin penalty is suppressed in block 1 (0.488 and 0.524) while the haematoxylin penalty dominates across all four blocks. Conversely, on Kather2016 (which includes stroma, complex stroma, adipose and mucosa) and Oral Cancer (characterised by eosinophilic keratin pearls), the eosin prior $\sigma_+(\beta_E)$ overtakes $\sigma_+(\beta_H)$ in the deeper blocks (e.g., 1.145 vs.\ 0.941 in block 3 of Kather2016; 1.083 vs.\ 0.962 in block 4 of Oral Cancer). In contrast, the spatial--semantic gates $g$ remain close to 0.5 (0.446 to 0.594), indicating an approximately balanced mixture of spatial-contact and semantic compartment messaging.

\paragraph{Are the learned priors consistent with histology?} The ratio $\sigma_+(\beta_H)/\sigma_+(\beta_E)$ (last column of \cref{tab:priors}A) summarises which dye boundary the attention respects more. It was not supervised: both priors start at $1$, and the loss contains no stain term. The ratio nevertheless separates the datasets by the tissue feature on which their labels depend. On the two tasks defined by nuclear atypia of epithelium, BreakHis (benign vs.\ malignant breast tumour) and LC25000 (adenocarcinoma, squamous carcinoma and benign lung and colon), the haematoxylin prior exceeds the eosin prior in all four blocks (mean ratio $1.42$ and $1.62$). On the two tasks whose classes are distinguished largely by extracellular or cytoplasmic eosinophilic material, Kather2016 (stroma, complex stroma, mucosa, adipose, debris) and Oral Cancer (keratinising squamous epithelium with keratin pearls), the ratio falls below 1 in the deep blocks ($0.90$ and $0.91$), so deep-layer attention there respects eosin boundaries more than haematoxylin boundaries. Lymphoma, whose three subtypes differ in nuclear size, contour and chromatin within otherwise uniform lymphocyte sheets, has only a mild haematoxylin preference in the priors ($1.06$, haematoxylin-dominant in three of four blocks). However, its realised attention is the most haematoxylin-sensitive of all six datasets: in block 4 the correlation between attention and $|\Delta h|$ reaches $-0.354$, against $-0.292$ for $|\Delta e|$ (\cref{tab:edgerho}). The priors are only one factor in \cref{eq:attn}; the realised sensitivity also depends on how much haematoxylin contrast exists between neighbouring nodes, and that contrast is largest in densely nucleated lymphoma tiles. NCT-CRC-HE-100K, a nine-class mixture of epithelial, stromal and non-tissue classes, is eosin-leaning in blocks 1--2 and strongly haematoxylin-dominant in blocks 3--4 ($1.63$), consistent with coarse stromal partitioning early and nuclear detail late. Two caveats apply: the priors come from a single representative run (seed 42), and we did not test the correspondence against pathologist annotations. The qualitative pattern, haematoxylin-dominant priors where nuclei carry the diagnosis and eosin-leaning deep priors where extracellular matrix and keratin do, is nonetheless what Beer--Lambert stain chemistry predicts. We therefore treat it as a hypothesis-generating observation rather than as evidence of histological validity. Whether the priors carry information that pathologists would recognise requires multi-seed replication and a reader study with annotated tissue, which we leave to future work (\cref{sec:d_limitations}).

\begin{table}[!htbp]
\caption{Learned stain priors and realised attention. \textbf{(A)}~Converged graph priors per block (representative run, seed 42): $\sigma_+(\beta_H)$ / $\sigma_+(\beta_E)$ / $g$, averaged over heads. Initial values: $\sigma_+(\beta)=1$, $g=0.5$. The last column gives the haematoxylin-to-eosin prior ratio $\sigma_+(\beta_H)/\sigma_+(\beta_E)$ averaged over all four blocks and over the two deep blocks; values above 1 mean that the attention is penalised more for crossing haematoxylin boundaries than eosin boundaries. \textbf{(B)}~Spearman correlation between inner-graph attention and $|\Delta h|$ / $|\Delta e|$ / spatial distance per block (representative run, seed 42; 300{,}000 edges per block).}\label{tab:priors}\label{tab:edgerho}
{\centering\small\textbf{(A)}\par}\smallskip
{\centering\footnotesize
\setlength{\tabcolsep}{4pt}

\begin{tabular}{@{}l c c c c c@{}}
\toprule
\textbf{Dataset} & \textbf{Block 1} & \textbf{Block 2} & \textbf{Block 3} & \textbf{Block 4} & \textbf{H/E ratio (all / 3--4)}\\
\midrule
BreakHis & 0.868 / 0.488 / 0.543 & 0.970 / 0.693 / 0.524 & 1.044 / 0.834 / 0.512 & 1.081 / 0.878 / 0.528 & 1.42 / 1.24\\
LC25000 & 1.100 / 0.524 / 0.485 & 0.823 / 0.511 / 0.498 & 0.845 / 0.626 / 0.500 & 0.869 / 0.604 / 0.574 & 1.62 / 1.39\\
Lymphoma & 0.747 / 0.762 / 0.453 & 0.840 / 0.773 / 0.498 & 0.841 / 0.796 / 0.446 & 0.827 / 0.744 / 0.495 & 1.06 / 1.08\\
Kather2016 & 0.860 / 0.890 / 0.528 & 0.846 / 0.756 / 0.529 & 0.941 / 1.145 / 0.517 & 0.933 / 0.954 / 0.594 & 0.97 / 0.90\\
NCT-CRC-HE-100K & 1.046 / 1.089 / 0.455 & 1.090 / 1.461 / 0.580 & 1.446 / 0.867 / 0.578 & 1.173 / 0.737 / 0.448 & 1.24 / 1.63\\
Oral cancer & 0.932 / 0.752 / 0.469 & 0.811 / 0.862 / 0.470 & 0.959 / 1.034 / 0.469 & 0.962 / 1.083 / 0.452 & 1.00 / 0.91\\
\bottomrule
\end{tabular}\par}
\par\medskip{\centering\small\textbf{(B)}\par}\smallskip
{\centering\footnotesize
\setlength{\tabcolsep}{3.5pt}

\begin{tabular}{@{}l c c c c@{}}
\toprule
\textbf{Dataset} & \textbf{Block 1} & \textbf{Block 2} & \textbf{Block 3} & \textbf{Block 4}\\
\midrule
BreakHis & +0.011 / +0.037 / +0.116 & -0.070 / -0.056 / +0.030 & -0.090 / -0.078 / +0.013 & -0.111 / -0.121 / -0.010\\
LC25000 & +0.069 / +0.079 / +0.068 & +0.008 / +0.077 / +0.120 & -0.039 / +0.028 / +0.099 & -0.150 / -0.072 / +0.025\\
Lymphoma & +0.026 / +0.025 / +0.122 & -0.052 / -0.053 / +0.135 & -0.213 / -0.179 / +0.061 & -0.354 / -0.292 / -0.018\\
Kather2016 & -0.006 / -0.077 / +0.118 & -0.046 / -0.092 / +0.095 & -0.062 / -0.132 / +0.042 & -0.087 / -0.140 / +0.040\\
NCT-CRC-HE-100K & +0.010 / -0.039 / +0.121 & +0.009 / -0.008 / +0.105 & -0.073 / -0.072 / +0.090 & -0.110 / -0.042 / +0.057\\
Oral cancer & -0.034 / -0.030 / +0.098 & -0.016 / -0.029 / +0.139 & -0.186 / -0.189 / +0.012 & -0.321 / -0.314 / -0.107\\
\bottomrule
\end{tabular}\par}
\end{table}

\begin{figure}[!htbp]
\centering
\begin{subfigure}[t]{0.32\textwidth}\centering\includegraphics[width=\linewidth]{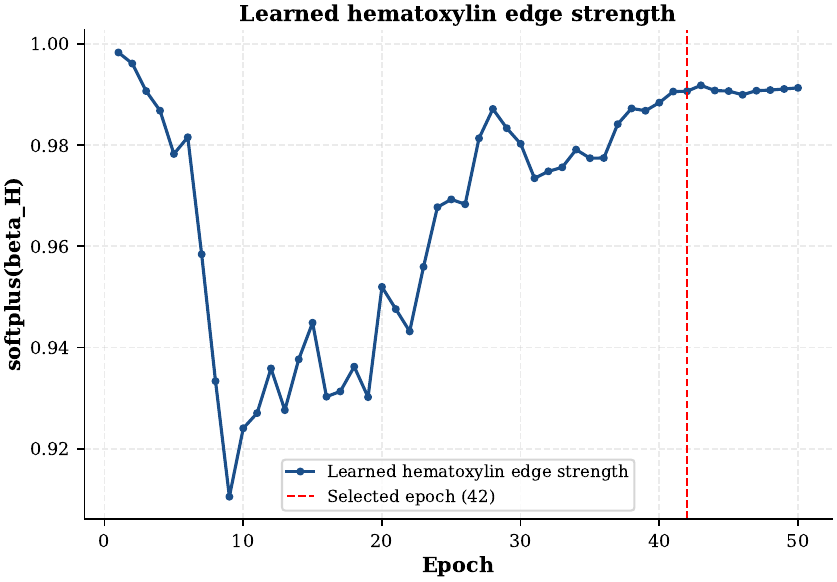}\caption{$\sigma_+(\beta_H)$ trajectory}\end{subfigure}\hfill
\begin{subfigure}[t]{0.32\textwidth}\centering\includegraphics[width=\linewidth]{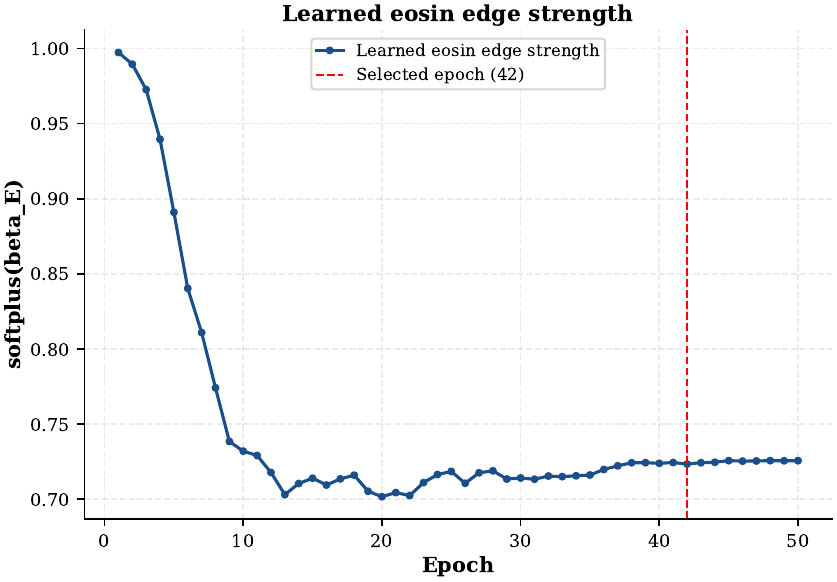}\caption{$\sigma_+(\beta_E)$ trajectory}\end{subfigure}\hfill
\begin{subfigure}[t]{0.32\textwidth}\centering\includegraphics[width=\linewidth]{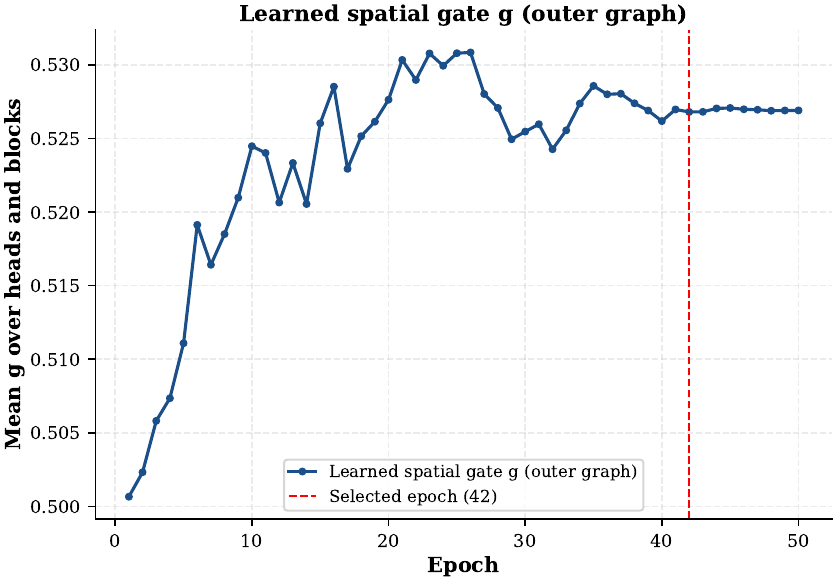}\caption{Spatial gate $g$ trajectory}\end{subfigure}\\[4pt]
\begin{subfigure}[t]{0.32\textwidth}\centering\includegraphics[width=\linewidth]{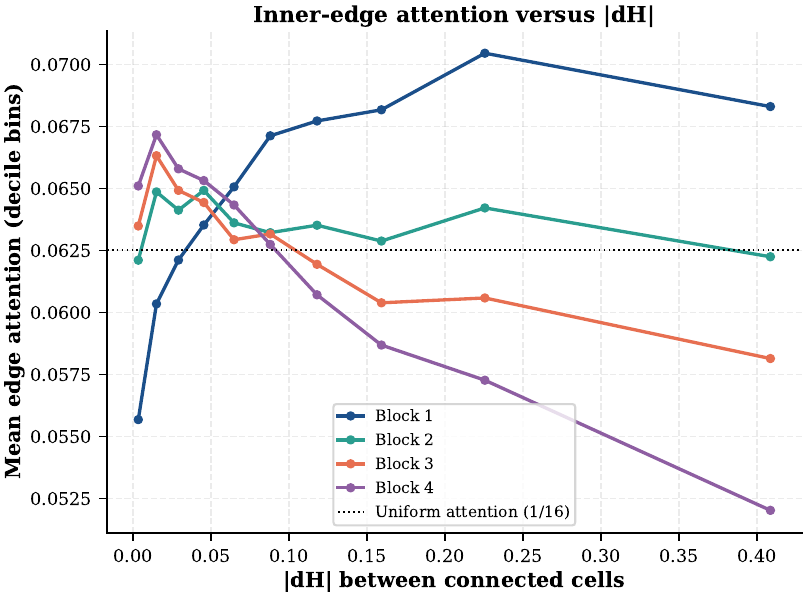}\caption{Attention vs $|\Delta h|$}\end{subfigure}\hfill
\begin{subfigure}[t]{0.32\textwidth}\centering\includegraphics[width=\linewidth]{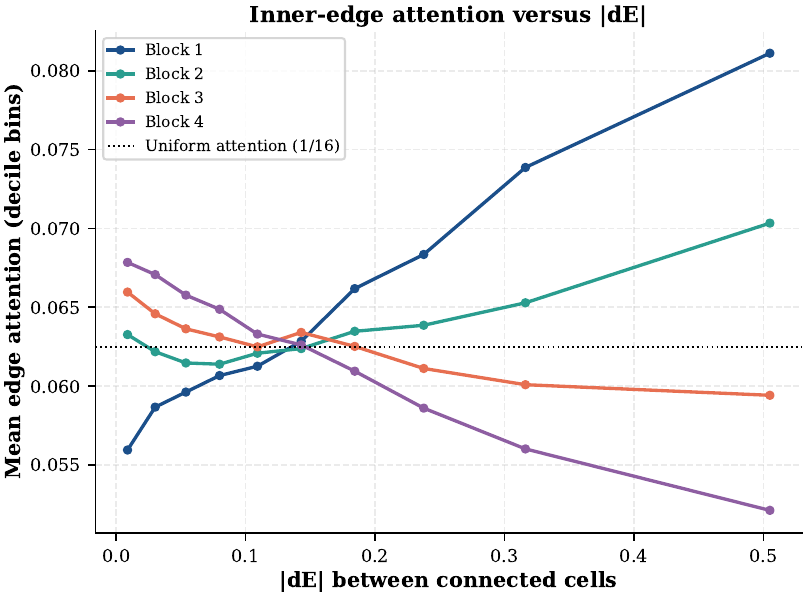}\caption{Attention vs $|\Delta e|$}\end{subfigure}\hfill
\begin{subfigure}[t]{0.32\textwidth}\centering\includegraphics[width=\linewidth]{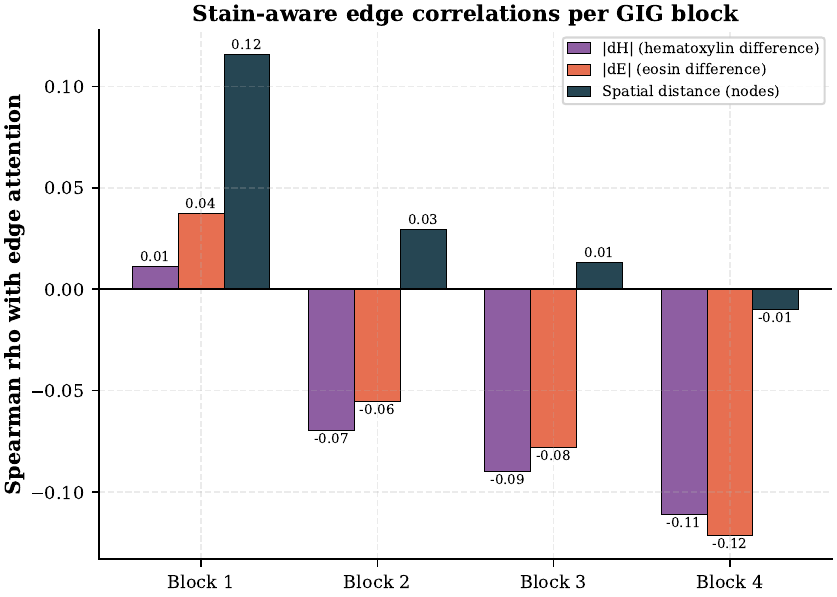}\caption{Correlations per block}\end{subfigure}
\caption{Stain-prior mechanics on BreakHis (representative run, seed 42).}
\label{fig:stainprior}
\end{figure}

\paragraph{Attention versus stain difference across depth.} \Cref{tab:edgerho} and \cref{fig:stainprior} report the Spearman rank correlation between realised inner-graph attention weights $\alpha_{ij}$ and stain differences $|\Delta h|,|\Delta e|$ across 300{,}000 sampled edges per block. While block 1 attention is governed primarily by content and relative position ($\rho\in[-0.08,+0.08]$), by block 4 attention correlates negatively with both haematoxylin difference ($\rho\in[-0.087,-0.354]$) and eosin difference ($\rho\in[-0.042,-0.314]$) on all six datasets. Because \cref{eq:attn} subtracts a non-negative stain term, a negative correlation in deep blocks is expected by construction and serves as a consistency check rather than independent evidence; the informative observation is that block-1 correlations stay near zero or positive despite the same penalty, so content and position dominate early attention.

\begin{figure}[!htbp]
\centering
\begin{subfigure}[t]{0.245\textwidth}\centering\includegraphics[width=\linewidth]{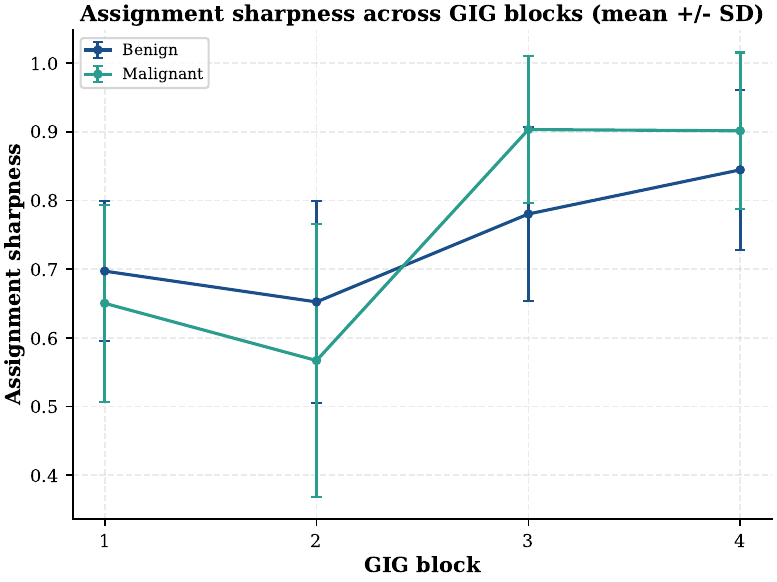}\caption{Assignment sharpness}\end{subfigure}\hfill
\begin{subfigure}[t]{0.245\textwidth}\centering\includegraphics[width=\linewidth]{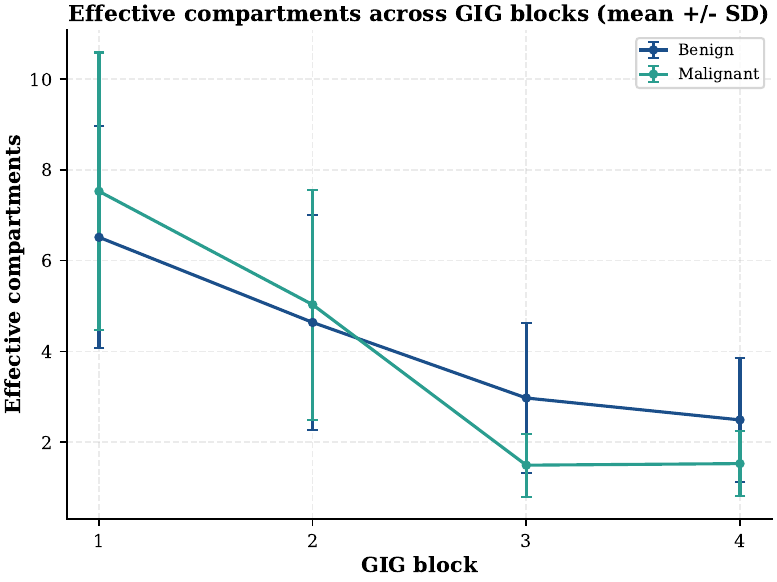}\caption{Effective compartments}\end{subfigure}\hfill
\begin{subfigure}[t]{0.245\textwidth}\centering\includegraphics[width=\linewidth]{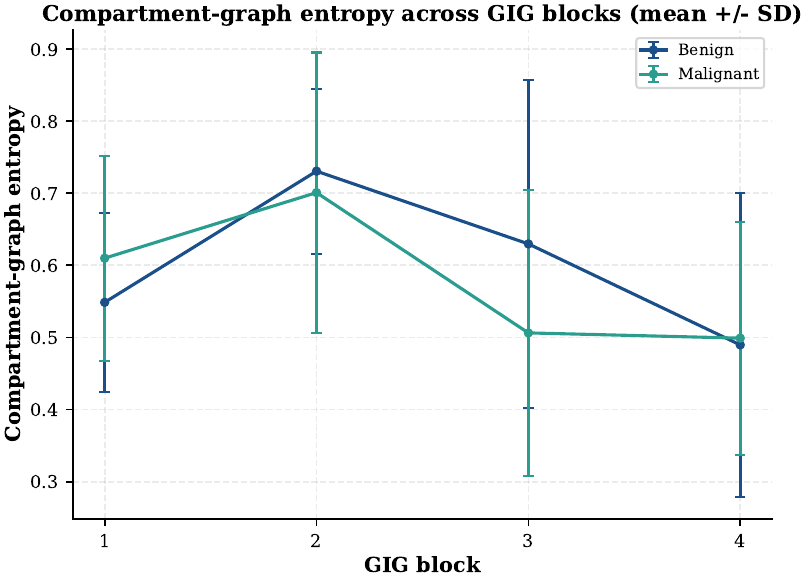}\caption{Graph entropy}\end{subfigure}\hfill
\begin{subfigure}[t]{0.245\textwidth}\centering\includegraphics[width=\linewidth]{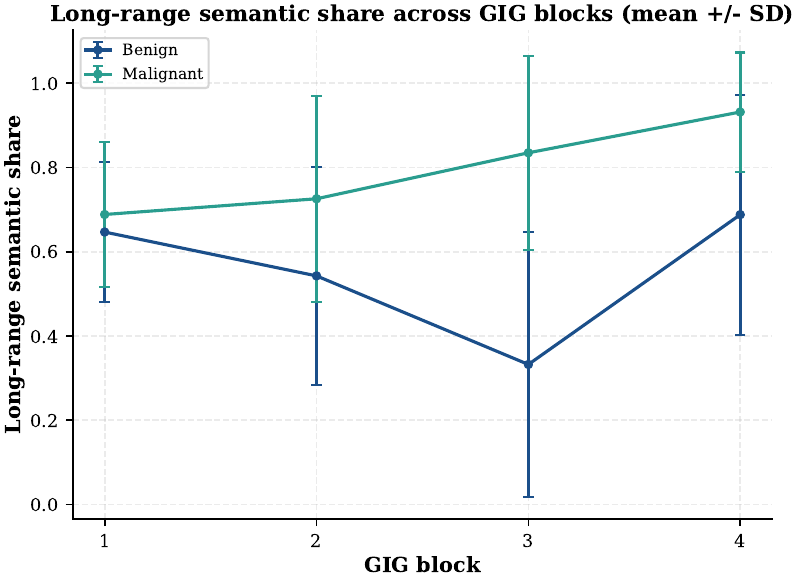}\caption{Long-range semantic share}\end{subfigure}
\caption{Layer-wise graph statistics on BreakHis by class (representative run, seed 42).}
\label{fig:layer}
\end{figure}

\begin{figure}[!htbp]
\centering
\includegraphics[width=0.92\textwidth]{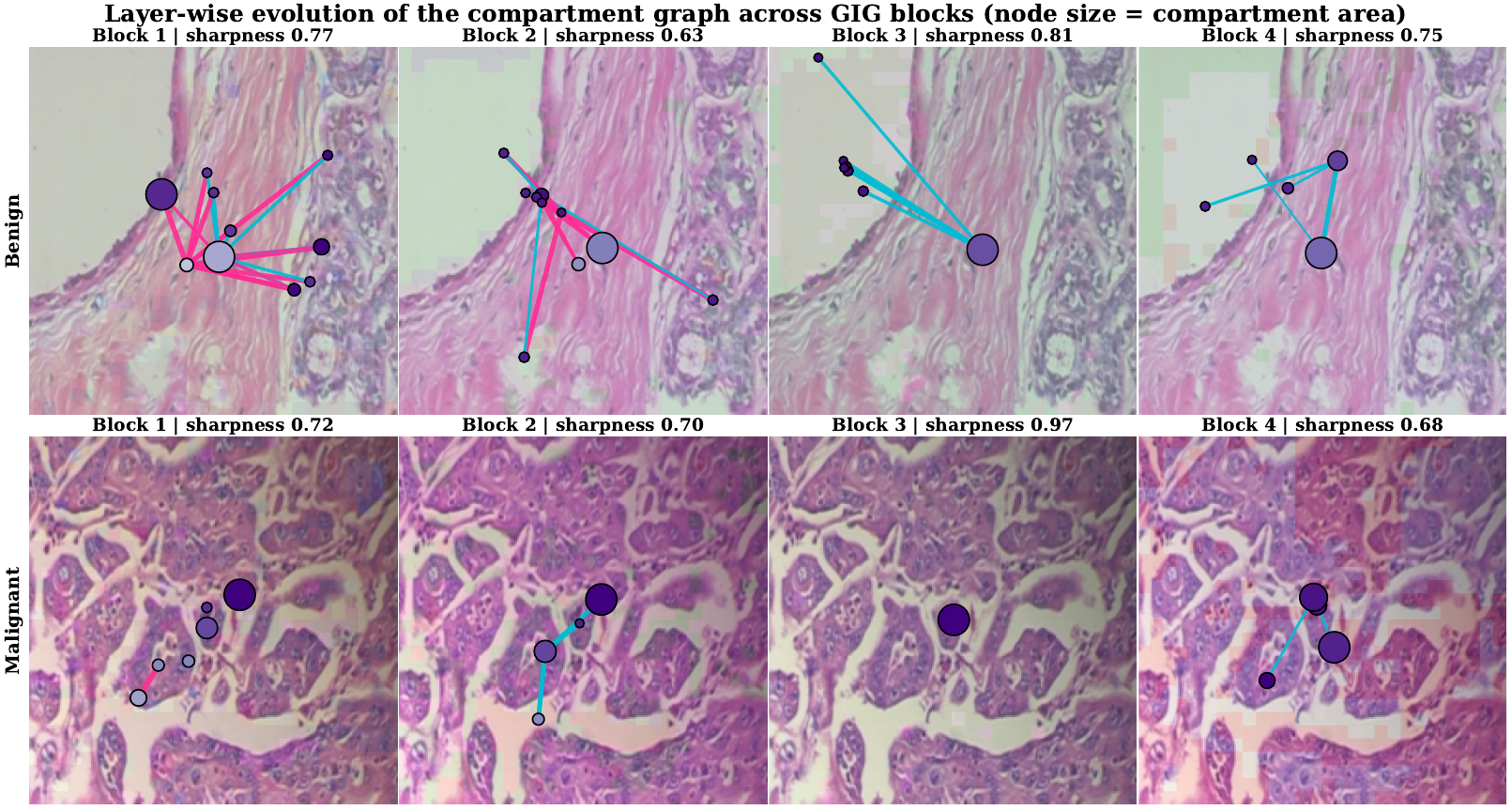}
\caption{Evolution of the compartment graph across the four nested blocks for a benign (top) and a malignant (bottom) BreakHis test tile.}
\label{fig:evolution}
\end{figure}

\begin{figure}[!htbp]
\centering
\begin{subfigure}[t]{0.245\textwidth}\centering\includegraphics[width=\linewidth]{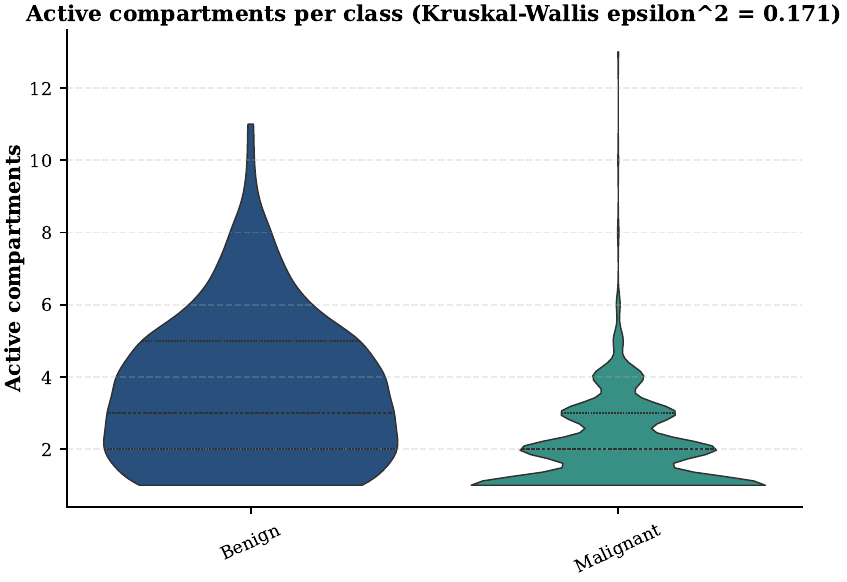}\caption{Active compartments}\end{subfigure}\hfill
\begin{subfigure}[t]{0.245\textwidth}\centering\includegraphics[width=\linewidth]{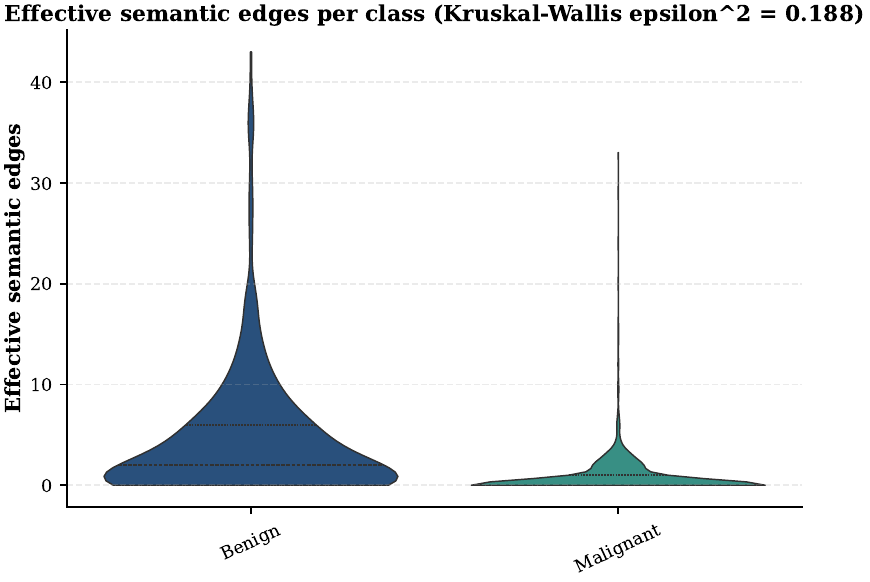}\caption{Effective semantic edges}\end{subfigure}\hfill
\begin{subfigure}[t]{0.245\textwidth}\centering\includegraphics[width=\linewidth]{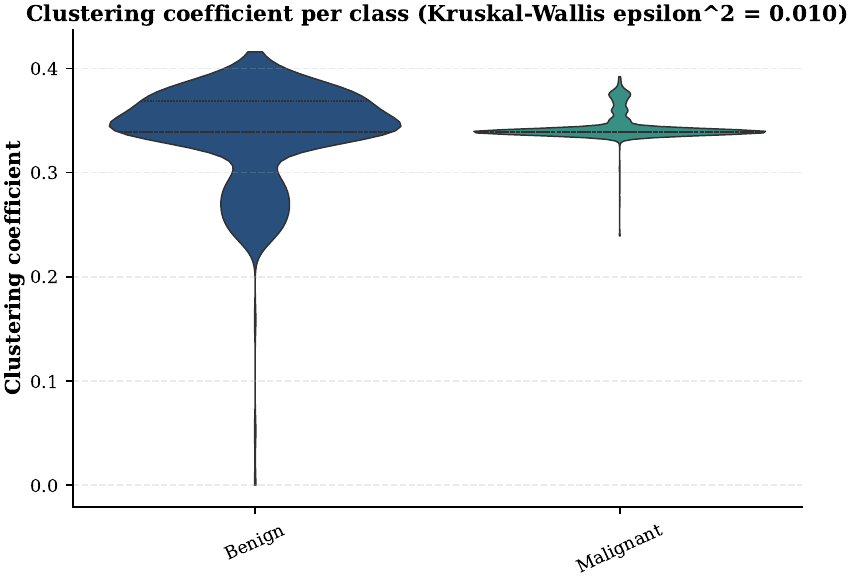}\caption{Clustering coefficient}\end{subfigure}\hfill
\begin{subfigure}[t]{0.245\textwidth}\centering\includegraphics[width=\linewidth]{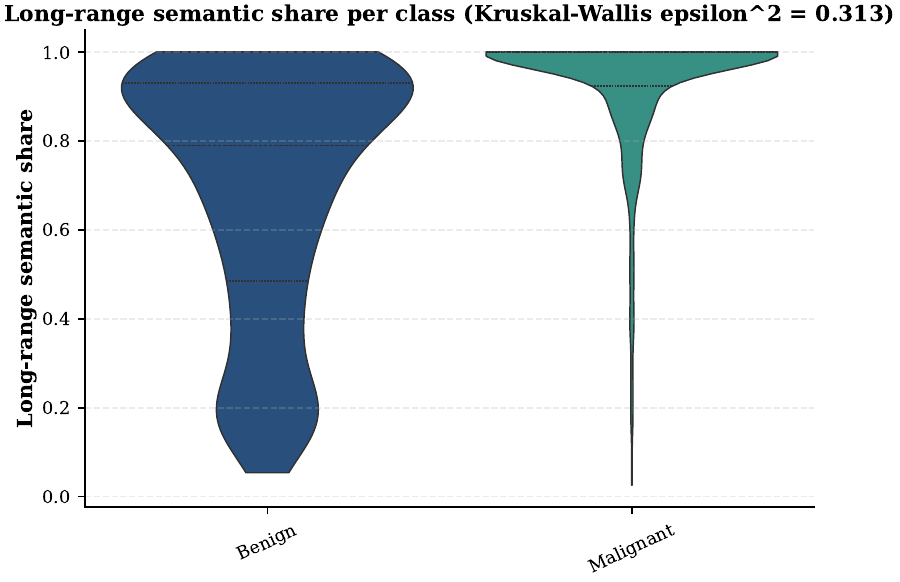}\caption{Long-range semantic share}\end{subfigure}\\[4pt]
\begin{subfigure}[t]{0.245\textwidth}\centering\includegraphics[width=\linewidth]{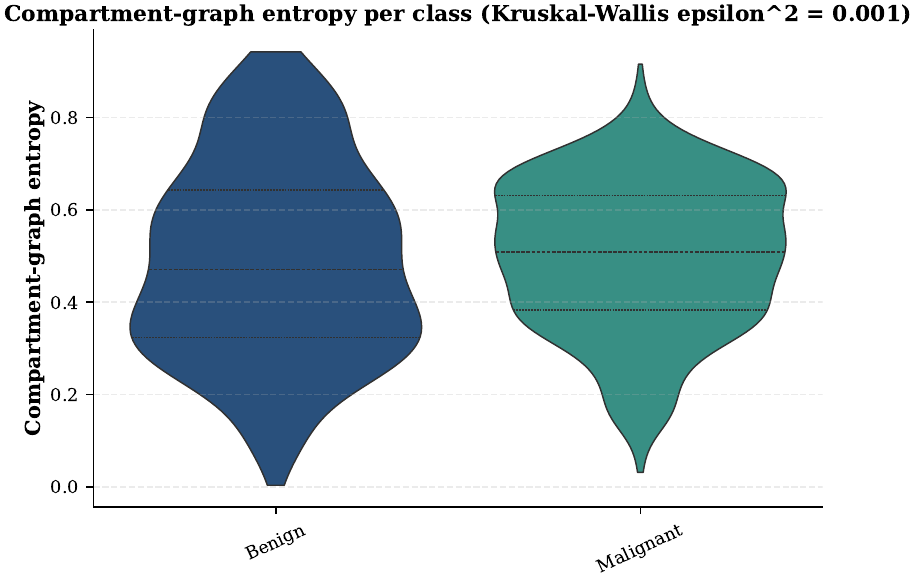}\caption{Compartment-graph entropy}\end{subfigure}\hfill
\begin{subfigure}[t]{0.245\textwidth}\centering\includegraphics[width=\linewidth]{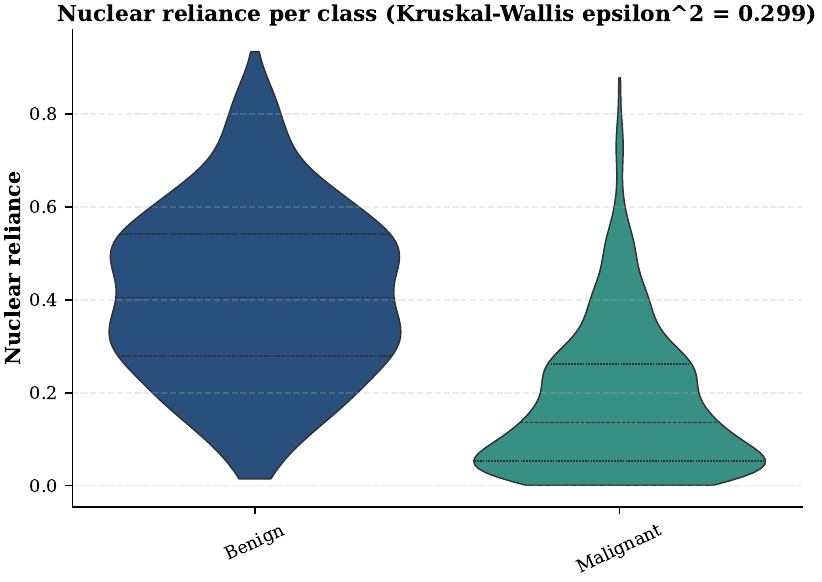}\caption{Nuclear reliance}\end{subfigure}\hfill
\begin{subfigure}[t]{0.245\textwidth}\centering\includegraphics[width=\linewidth]{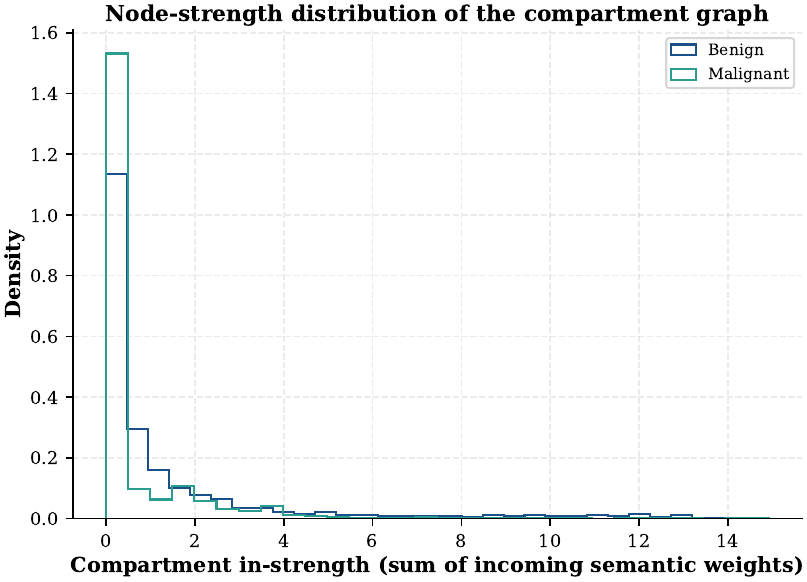}\caption{Node strength}\end{subfigure}\hfill
\begin{subfigure}[t]{0.245\textwidth}\centering\includegraphics[width=\linewidth]{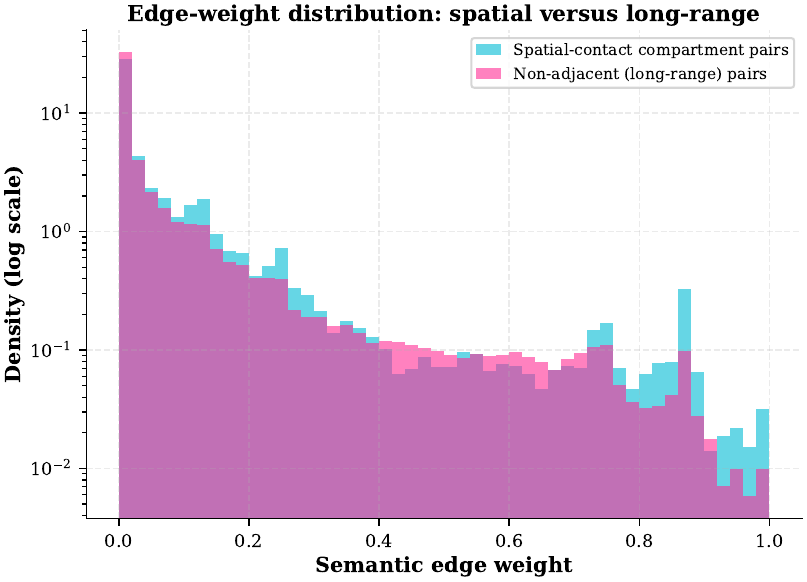}\caption{Edge weight}\end{subfigure}
\caption{Distributions of last-block graph statistics on the BreakHis test set by class.}
\label{fig:graphstats}
\end{figure}

\begin{figure}[!htbp]
\centering
\begin{subfigure}[t]{0.32\textwidth}\centering\includegraphics[width=\linewidth]{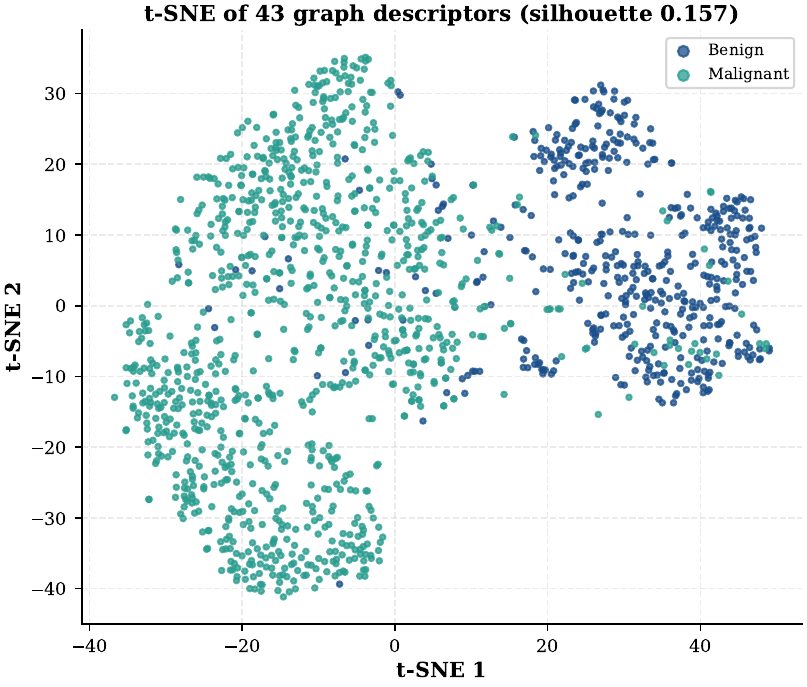}\caption{Descriptor t-SNE}\end{subfigure}\hfill
\begin{subfigure}[t]{0.32\textwidth}\centering\includegraphics[width=\linewidth]{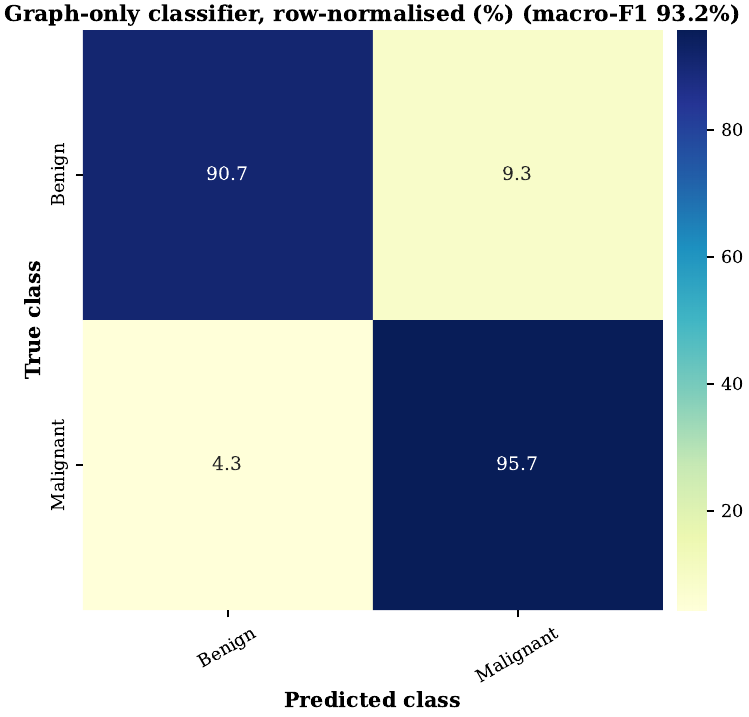}\caption{Graph-only confusion matrix}\end{subfigure}\hfill
\begin{subfigure}[t]{0.32\textwidth}\centering\includegraphics[width=\linewidth]{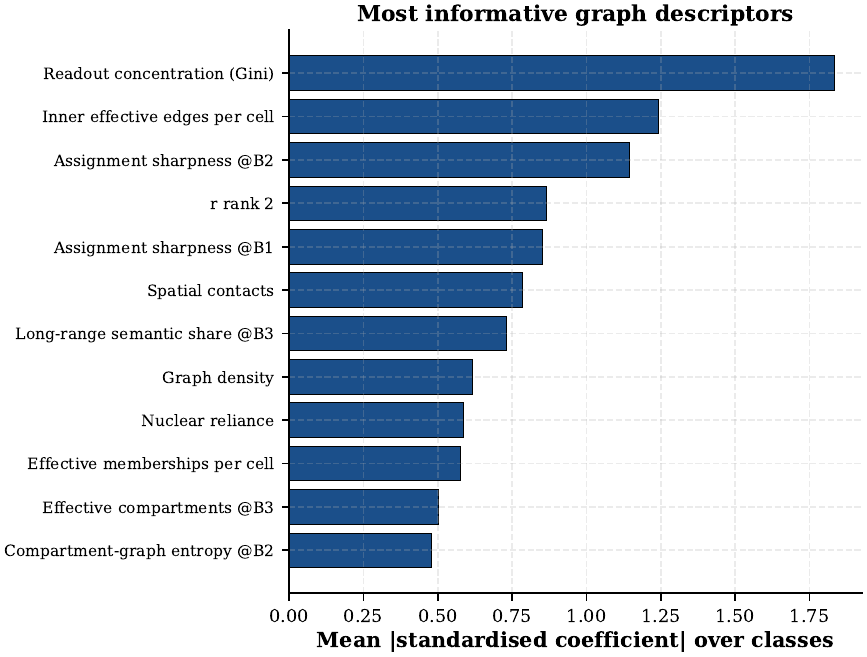}\caption{Descriptor importance}\end{subfigure}
\caption{Graph-descriptor linear probe on BreakHis (representative run, seed 42).}
\label{fig:probe}
\end{figure}

\paragraph{Layer-wise graph evolution and topological probe.} \Cref{tab:layer} and \cref{fig:layer,fig:evolution,fig:graphstats} track how compartment assignments evolve from block 1 to block 4. On architectural tasks with clear epithelial--stromal boundaries (BreakHis), assignment sharpness increases from 0.67 to 0.88 and effective compartments contract from 7.21 to 1.83, whereas on homogeneous sheets (Lymphoma, single-tissue Kather2016 tiles) assignments remain diffuse (8.63--12.98 effective compartments). Training a linear classifier solely on 43 scalar topological descriptors of the learned graphs without any neural feature vectors (\cref{tab:probe}, \cref{fig:probe}) recovers 89.73\% to 99.07\% accuracy across the six datasets (versus majority-class chance for a label-permuted control), confirming that diagnostic class information is encoded directly into the learned graph topology.

\begin{table}[!htbp]
\centering\footnotesize
\setlength{\tabcolsep}{4pt}
\caption{Layer-wise graph statistics, block~1 $\to$ block~4 (representative run, seed 42; mean over test tiles).}\label{tab:layer}
\begin{tabular}{@{}l c c c c c@{}}
\toprule
\textbf{Dataset} & \textbf{Assignment sharpness} & \textbf{Effective comp.} & \textbf{Graph entropy} & \textbf{Long-range sem.\ share} & \textbf{Active comp.}\\
\midrule
BreakHis & 0.67 $\to$ 0.88 & 7.21 $\to$ 1.83 & 0.59 $\to$ 0.50 & 0.68 $\to$ 0.86 & 9.90 $\to$ 2.48\\
LC25000 & 0.50 $\to$ 0.57 & 10.20 $\to$ 7.71 & 0.78 $\to$ 0.93 & 0.75 $\to$ 0.75 & 12.26 $\to$ 10.30\\
Lymphoma & 0.21 $\to$ 0.19 & 13.52 $\to$ 12.98 & 0.99 $\to$ 1.00 & 0.58 $\to$ 0.66 & 15.00 $\to$ 15.33\\
Kather2016 & 0.46 $\to$ 0.42 & 9.98 $\to$ 8.63 & 0.99 $\to$ 1.00 & 0.65 $\to$ 0.70 & 11.62 $\to$ 10.36\\
NCT-CRC-HE-100K & 0.73 $\to$ 0.72 & 8.15 $\to$ 6.38 & 0.63 $\to$ 0.73 & 0.69 $\to$ 0.77 & 10.12 $\to$ 8.47\\
Oral cancer & 0.58 $\to$ 0.59 & 9.39 $\to$ 6.35 & 0.80 $\to$ 0.96 & 0.69 $\to$ 0.74 & 12.50 $\to$ 8.29\\
\bottomrule
\end{tabular}
\end{table}

\begin{table}[!htbp]
\centering\footnotesize
\setlength{\tabcolsep}{4pt}
\caption{Graph-descriptor linear probe (representative run, seed 42): classification accuracy and macro-F1 (\%) using solely 43 scalar topological descriptors of the learned graphs.}\label{tab:probe}
\begin{tabular}{@{}l c c c c c c@{}}
\toprule
 & \textbf{Full model} & \multicolumn{2}{c}{\textbf{All 43 descriptors}} & \textbf{16 layer-wise} & \textbf{Permuted labels} & \textbf{Majority}\\
\cmidrule(lr){3-4}
\textbf{Dataset} & \textbf{Macro-F1} & \textbf{Acc.} & \textbf{Macro-F1} & \textbf{Macro-F1} & \textbf{Acc.} & \textbf{Acc.}\\
\midrule
BreakHis & 95.48 & 94.11 & 93.16 & 90.07 & 68.76 & 68.63\\
LC25000 & 99.92 & 99.07 & 99.08 & 97.36 & 19.80 & 20.62\\
Lymphoma & 95.87 & 89.73 & 89.83 & 83.51 & 35.73 & 37.33\\
Kather2016 & 94.92 & 90.57 & 90.65 & 91.06 & 12.07 & 12.78\\
NCT-CRC-HE-100K & 99.72 & 94.64 & 94.47 & 85.97 & 13.30 & 14.42\\
Oral cancer & 95.51 & 91.12 & 91.11 & 88.35 & 51.96 & 51.19\\
\bottomrule
\end{tabular}
\end{table}

%% ---------------------------------------------------------------------
\subsection{Audit of graph-native attributions versus post-hoc explanation faithfulness}\label{sec:r_xai}

\begin{figure}[!htbp]
\centering
\includegraphics[width=0.162\textwidth]{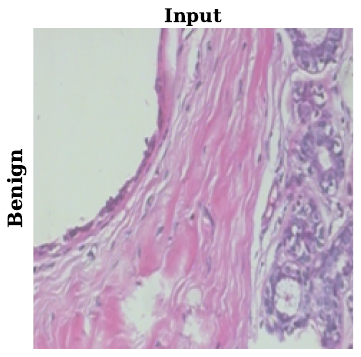}\hfill
\includegraphics[width=0.162\textwidth]{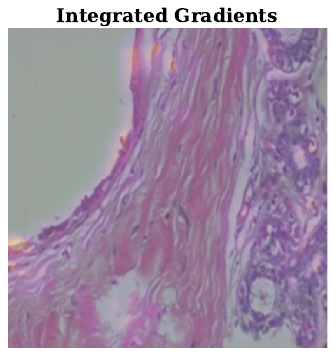}\hfill
\includegraphics[width=0.162\textwidth]{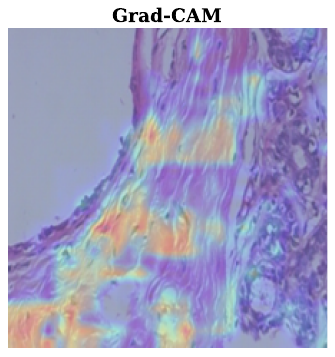}\hfill
\includegraphics[width=0.162\textwidth]{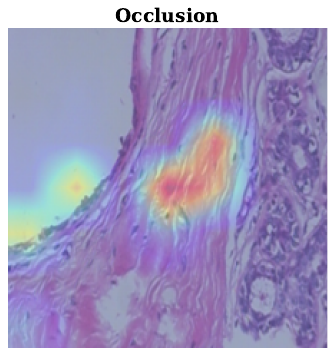}\hfill
\includegraphics[width=0.162\textwidth]{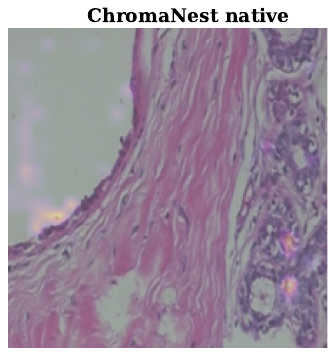}\hfill
\includegraphics[width=0.162\textwidth]{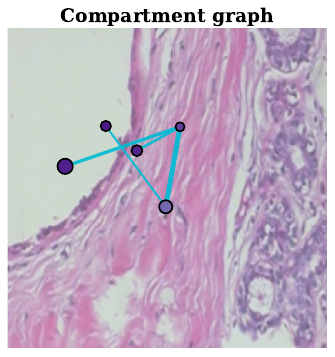}\\[1pt]
\includegraphics[width=0.162\textwidth]{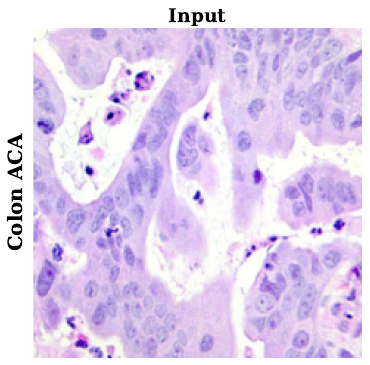}\hfill
\includegraphics[width=0.162\textwidth]{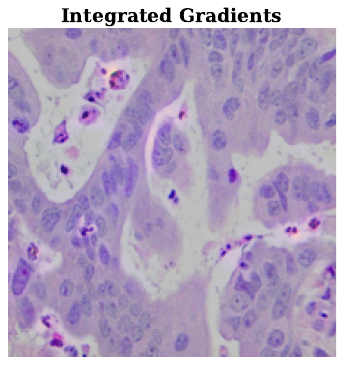}\hfill
\includegraphics[width=0.162\textwidth]{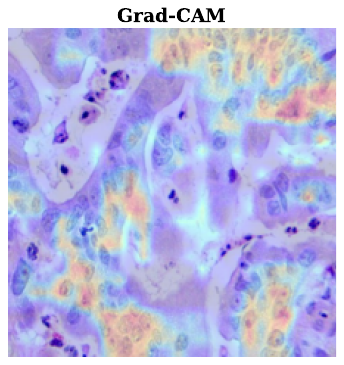}\hfill
\includegraphics[width=0.162\textwidth]{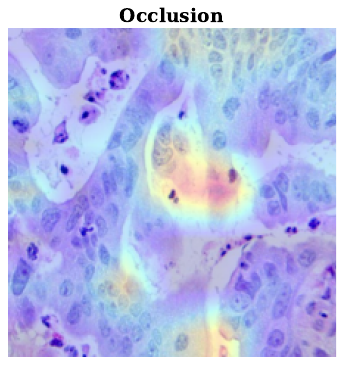}\hfill
\includegraphics[width=0.162\textwidth]{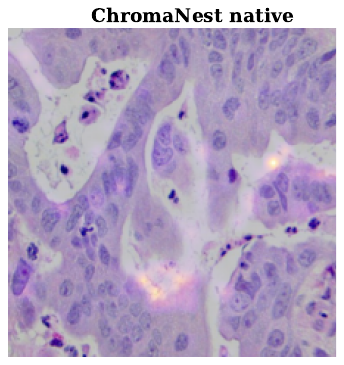}\hfill
\includegraphics[width=0.162\textwidth]{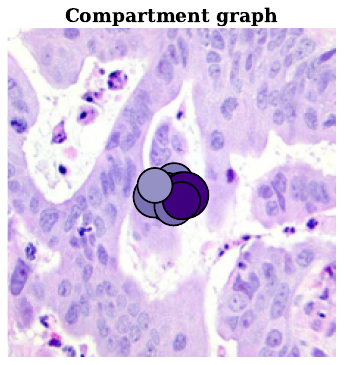}\\[1pt]
\includegraphics[width=0.162\textwidth]{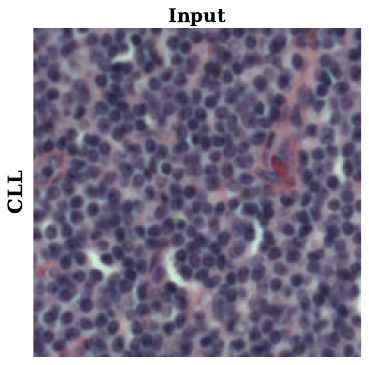}\hfill
\includegraphics[width=0.162\textwidth]{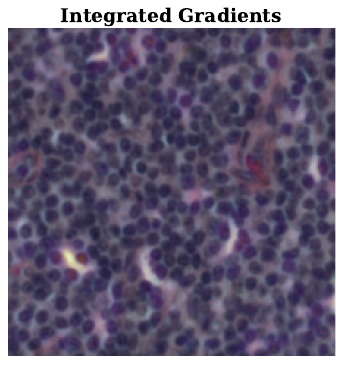}\hfill
\includegraphics[width=0.162\textwidth]{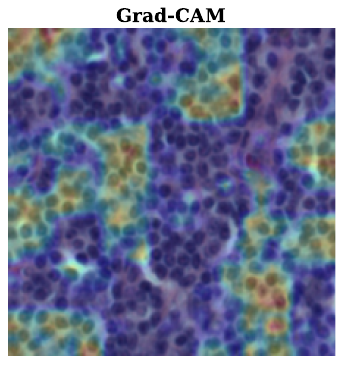}\hfill
\includegraphics[width=0.162\textwidth]{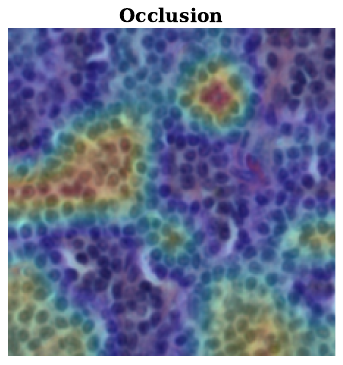}\hfill
\includegraphics[width=0.162\textwidth]{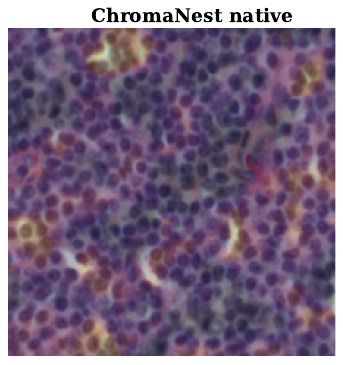}\hfill
\includegraphics[width=0.162\textwidth]{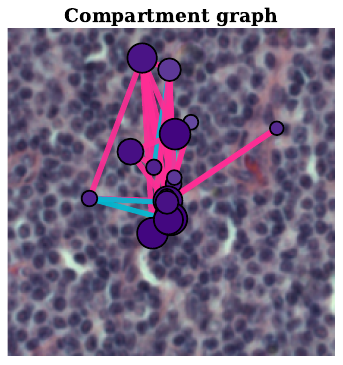}\\[1pt]
\includegraphics[width=0.162\textwidth]{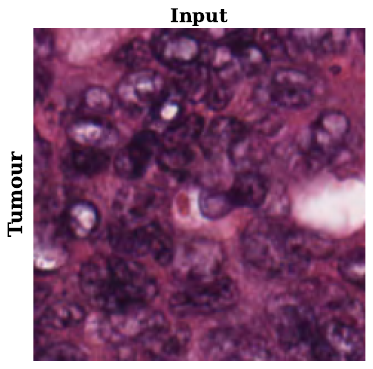}\hfill
\includegraphics[width=0.162\textwidth]{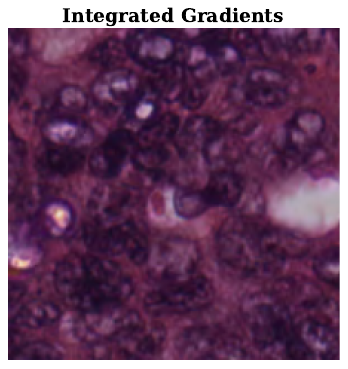}\hfill
\includegraphics[width=0.162\textwidth]{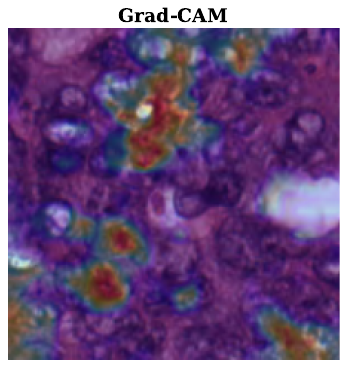}\hfill
\includegraphics[width=0.162\textwidth]{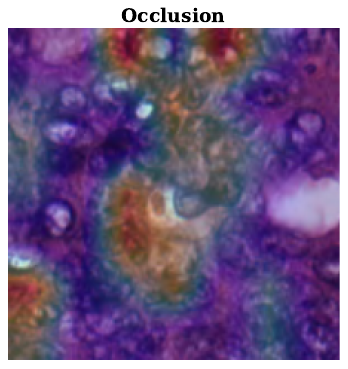}\hfill
\includegraphics[width=0.162\textwidth]{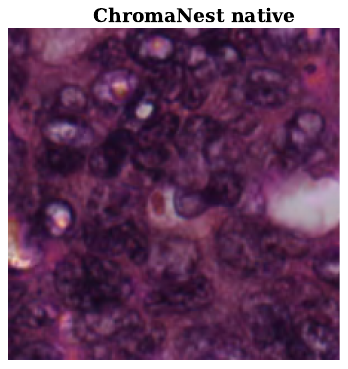}\hfill
\includegraphics[width=0.162\textwidth]{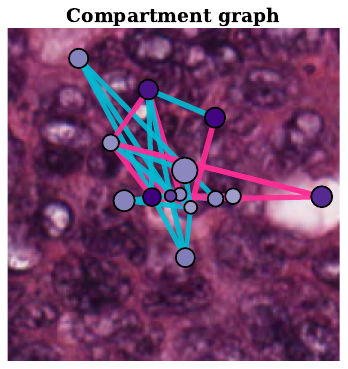}\\[1pt]
\includegraphics[width=0.162\textwidth]{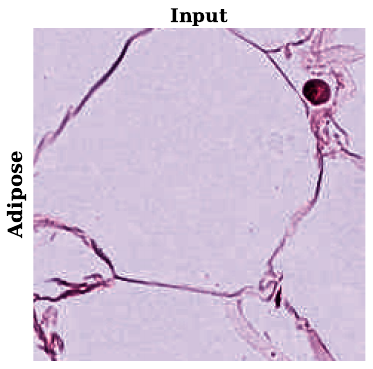}\hfill
\includegraphics[width=0.162\textwidth]{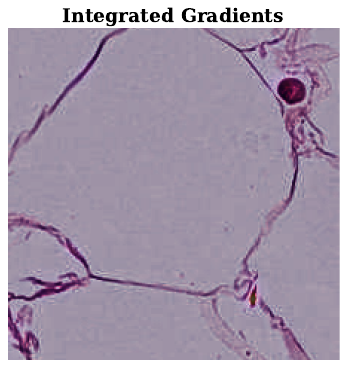}\hfill
\includegraphics[width=0.162\textwidth]{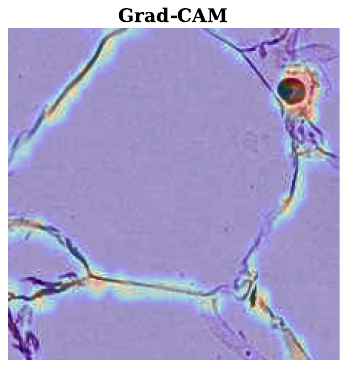}\hfill
\includegraphics[width=0.162\textwidth]{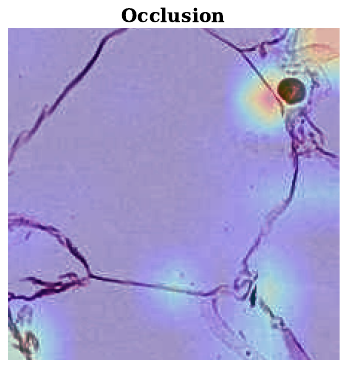}\hfill
\includegraphics[width=0.162\textwidth]{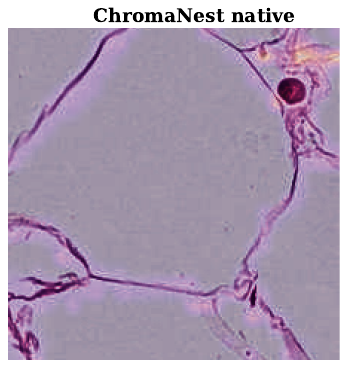}\hfill
\includegraphics[width=0.162\textwidth]{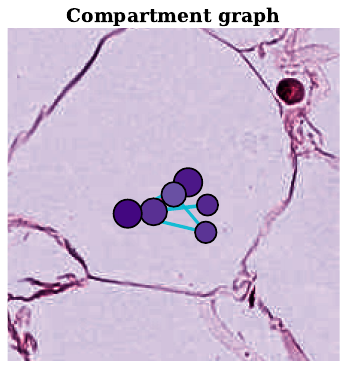}\\[1pt]
\includegraphics[width=0.162\textwidth]{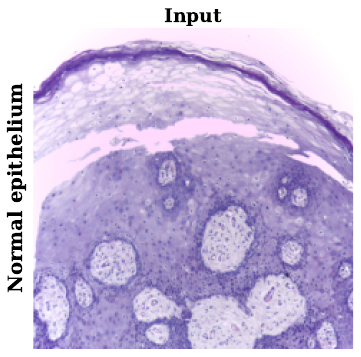}\hfill
\includegraphics[width=0.162\textwidth]{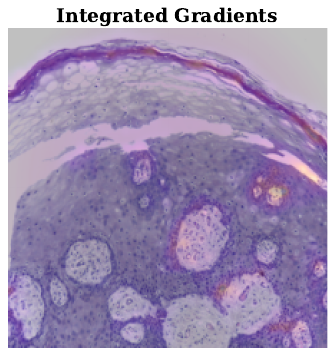}\hfill
\includegraphics[width=0.162\textwidth]{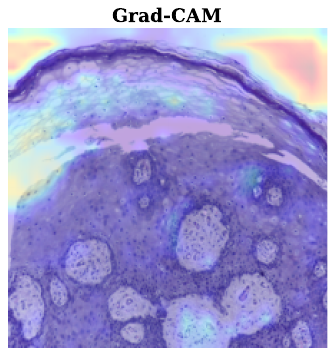}\hfill
\includegraphics[width=0.162\textwidth]{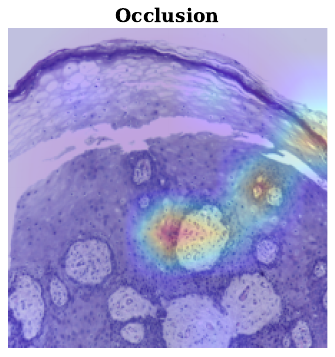}\hfill
\includegraphics[width=0.162\textwidth]{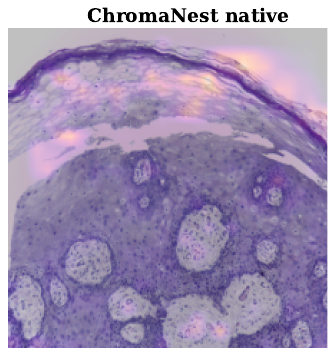}\hfill
\includegraphics[width=0.162\textwidth]{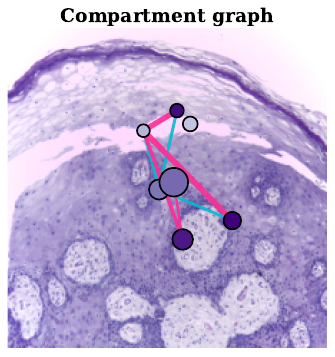}
\caption{Post-hoc attributions and learned compartment graphs on one test tile per dataset (top to bottom: BreakHis, LC25000, Lymphoma, Kather2016, NCT-CRC-HE-100K, Oral Cancer). Columns: input, Integrated Gradients, Grad-CAM, occlusion, \model{} native map, and the last-block compartment graph.}
\label{fig:xai_examples}
\end{figure}

\begin{figure}[!htbp]
{\centering
\begin{subfigure}[t]{0.32\textwidth}\centering\includegraphics[width=\linewidth]{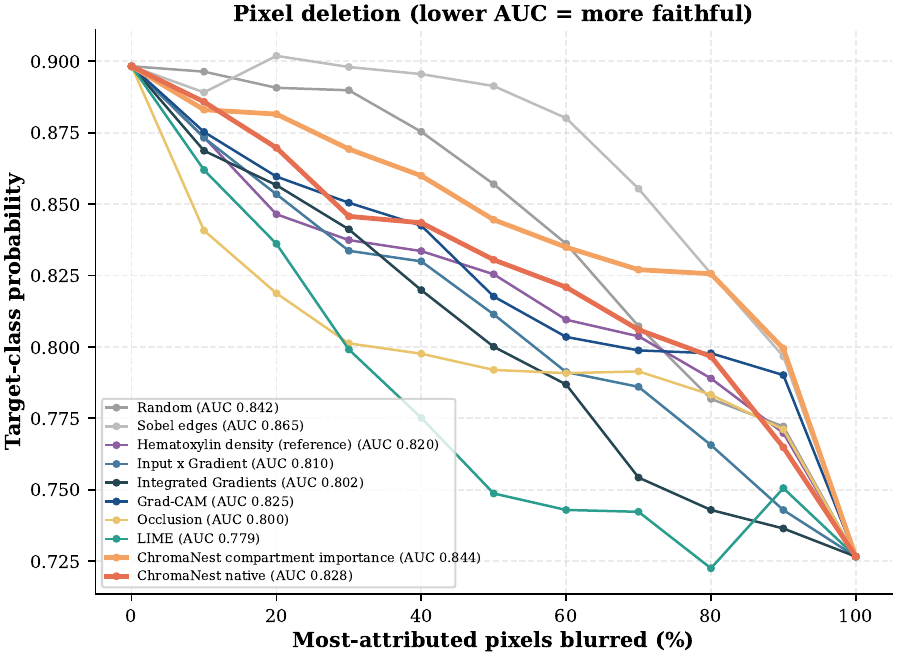}\caption{Pixel deletion}\end{subfigure}\hfill
\begin{subfigure}[t]{0.32\textwidth}\centering\includegraphics[width=\linewidth]{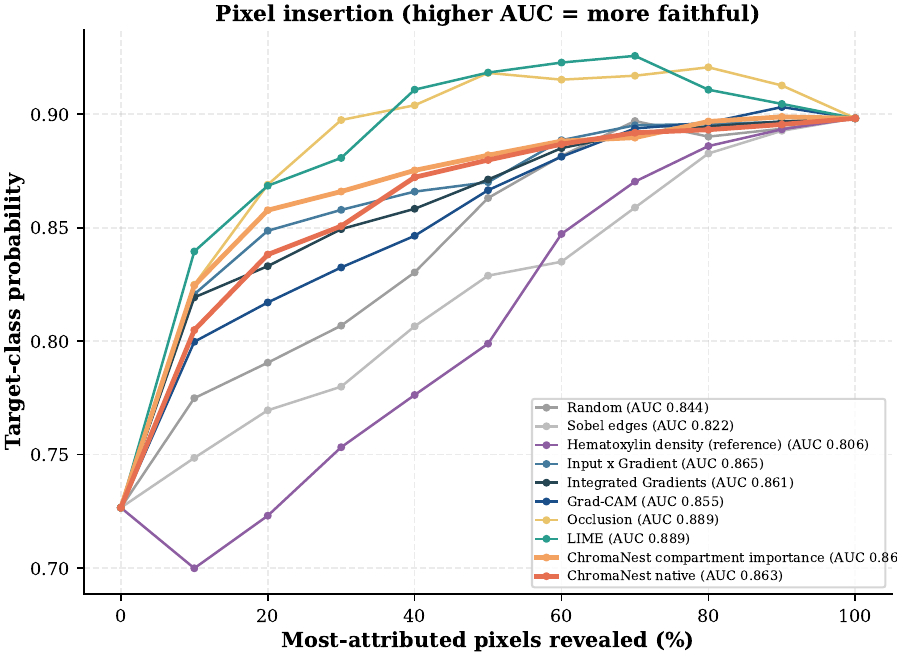}\caption{Pixel insertion}\end{subfigure}\hfill
\begin{subfigure}[t]{0.32\textwidth}\centering\includegraphics[width=\linewidth]{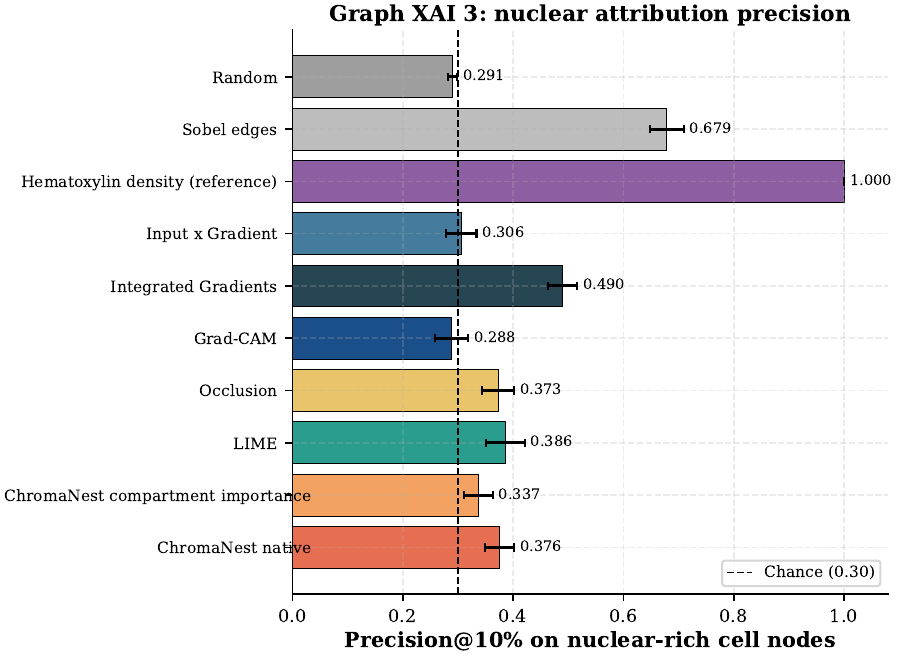}\caption{Nuclear precision@10\%}\end{subfigure}\\[4pt]
\begin{subfigure}[t]{0.32\textwidth}\centering\includegraphics[width=\linewidth]{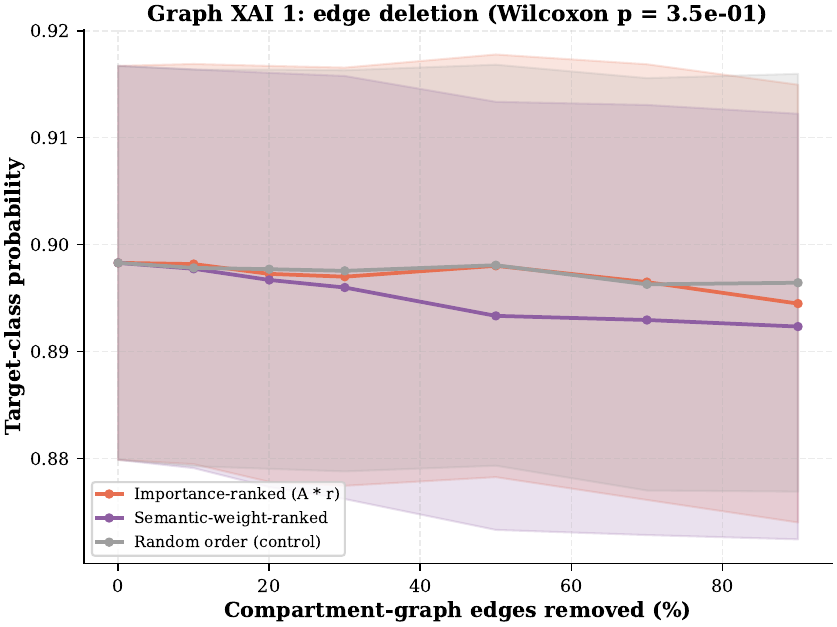}\caption{Edge deletion}\end{subfigure}\hfill
\begin{subfigure}[t]{0.32\textwidth}\centering\includegraphics[width=\linewidth]{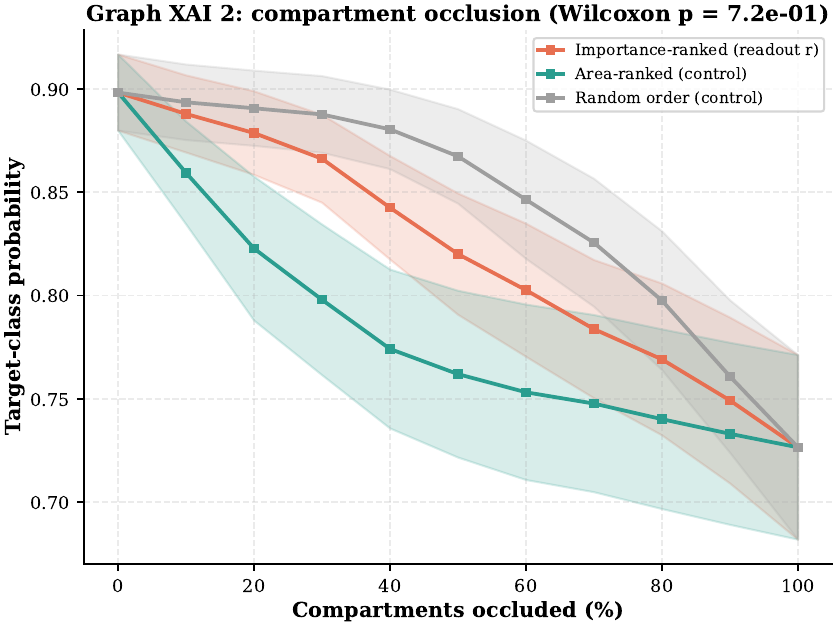}\caption{Compartment occlusion}\end{subfigure}\hfill
\begin{subfigure}[t]{0.32\textwidth}\centering\includegraphics[width=\linewidth]{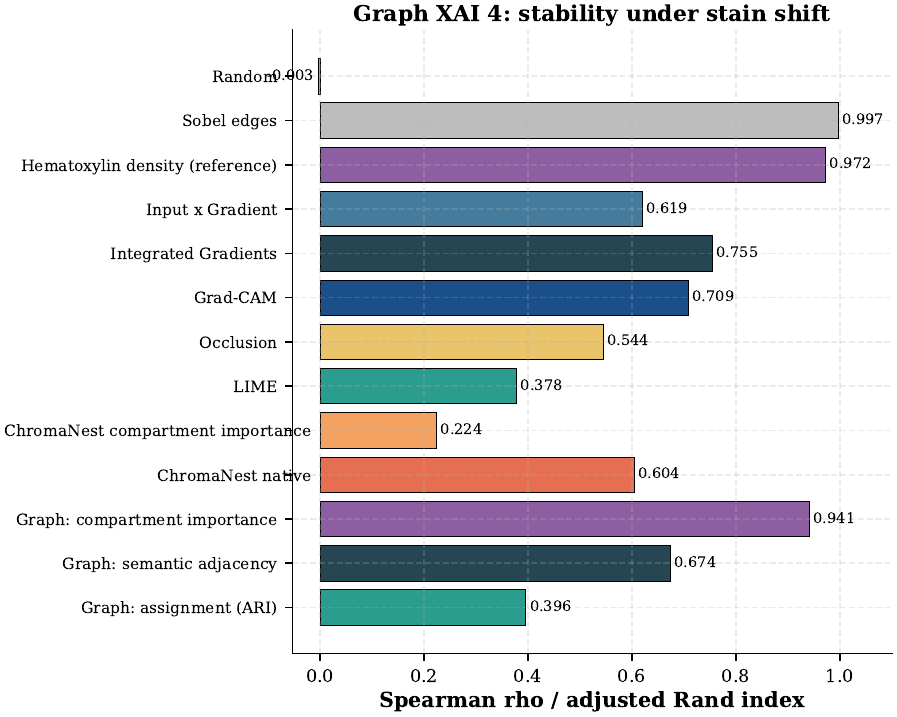}\caption{Stain-shift stability}\end{subfigure}\par}
\par\medskip
{\centering
\includegraphics[width=0.66\textwidth]{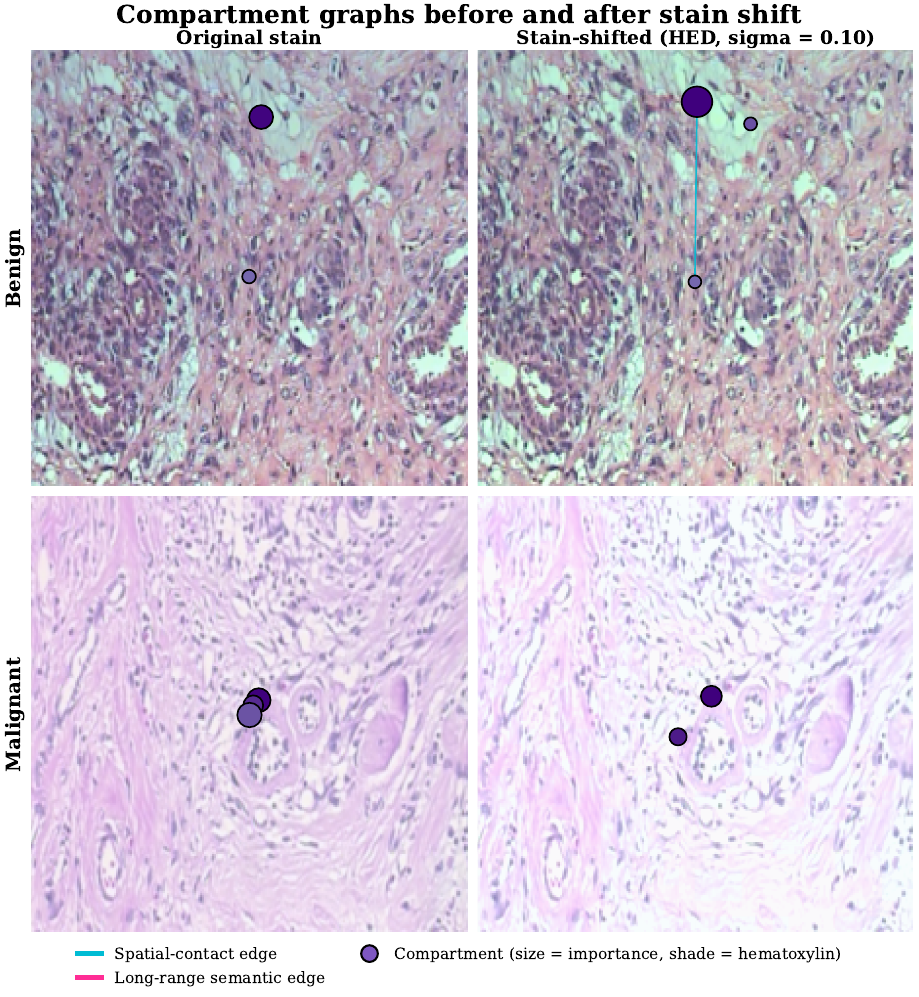}\par}
\caption{Faithfulness, stability and stain-shift audit on BreakHis. \textbf{(a)--(f)}~Faithfulness and stability audit on BreakHis (40 test images, representative run, seed 42). \textbf{Bottom:}~Learned compartment graphs on a benign (top) and a malignant (bottom) BreakHis test tile before (left) and after (right) a synthetic HED stain shift.}\label{fig:xai_curves}\label{fig:stainshift}
\end{figure}

A recurring pitfall in graph and attention models for pathology is assuming that built-in attention or pooling weights provide faithful post-hoc explanations~\cite{adebayo2018sanity,hense2024xmil}. We audit the pixel- and graph-level perturbation faithfulness of \model{}'s built-in maps (\cref{tab:xai_bh,tab:xai_cross,tab:xai_graph}; \cref{fig:xai_examples,fig:xai_curves,fig:stainshift}).

\paragraph{Stability and complementarity of built-in and post-hoc explanations.} At the pixel level (\cref{tab:xai_bh,tab:xai_cross}), \model{}'s native attribution map is highly concentrated (Gini 0.813), focuses on nuclei above chance (NP@10\% of 0.376 vs.\ 0.300 chance on BreakHis), and is more stable under synthetic HED stain shifts than LIME or occlusion across all six datasets. Under pixel deletion and insertion, the perturbation- and gradient-based methods LIME, Occlusion and Integrated Gradients achieve the strongest deletion and insertion scores. At the graph level (\cref{tab:xai_graph}), occluding compartments in order of their spatial area lowers prediction confidence substantially faster than random occlusion ($\Delta\mathrm{AUC}$ from $-0.066$ to $-0.308$), showing that the decision rests on spatially extensive tissue compartments; ordering by readout weight $r_m$ gives confidence curves close to random occlusion.

Under synthetic stain shifts (\cref{tab:xai_graph}, \cref{fig:stainshift}), the compartment weight ranking $r$ remains highly stable (Spearman $0.926$ to $0.993$) and predictions remain unchanged on 89.7\%--100\% of audited tiles. \model{}'s priors and compartment graphs therefore provide \emph{structural transparency} into how the network organises tissue, which complements pixel-level attribution from Integrated Gradients or LIME.

\begin{table}[!htbp]
\centering\footnotesize
\setlength{\tabcolsep}{3pt}
\caption{Pixel-level explanation audit on BreakHis (40 test images, representative run, seed 42). The haematoxylin-density row is a reference map that is nuclear by construction; its NP@10\% of 1.000 is not a property of \model{}, whose native map scores 0.376. Deletion/insertion: mean $\pm$ SD of the AUC of \cref{eq:delins}. NP@10\%: nuclear precision (chance 0.30). Stability: Spearman correlation before and after synthetic HED stain shift.}\label{tab:xai_bh}
\begin{tabular}{@{}l l c c c c c@{}}
\toprule
\textbf{Method} & \textbf{Type} & \textbf{Deletion AUC $\downarrow$} & \textbf{Insertion AUC $\uparrow$} & \textbf{NP@10\% $\uparrow$} & \textbf{Stability $\uparrow$} & \textbf{Gini}\\
\midrule
Random & Reference & 0.8419 $\pm$ 0.1507 & 0.8441 $\pm$ 0.1451 & 0.291 & -0.003 & 0.041\\
Hematoxylin density (reference) & Reference & 0.8201 $\pm$ 0.1682 & 0.8061 $\pm$ 0.1793 & 1.000 & 0.972 & 0.307\\
Input$\times$Gradient & Post-hoc & 0.8100 $\pm$ 0.1698 & 0.8652 $\pm$ 0.1378 & 0.306 & 0.619 & 0.419\\
Integrated Gradients & Post-hoc & 0.8019 $\pm$ 0.1891 & 0.8612 $\pm$ 0.1358 & 0.490 & 0.755 & 0.461\\
Grad-CAM & Post-hoc & 0.8248 $\pm$ 0.1848 & 0.8549 $\pm$ 0.1274 & 0.288 & 0.709 & 0.543\\
Occlusion & Post-hoc & 0.7999 $\pm$ 0.1736 & 0.8892 $\pm$ 0.1165 & 0.373 & 0.544 & 0.574\\
LIME & Post-hoc & \textbf{0.7792 $\pm$ 0.1806} & \textbf{0.8895 $\pm$ 0.1122} & 0.386 & 0.378 & 0.576\\
\model{} compartment importance & Graph-native & 0.8438 $\pm$ 0.1606 & 0.8692 $\pm$ 0.1291 & 0.337 & 0.224 & 0.120\\
\model{} native & Graph-native & 0.8276 $\pm$ 0.1624 & 0.8626 $\pm$ 0.1191 & 0.376 & 0.604 & 0.813\\
\bottomrule
\end{tabular}
\end{table}

\begin{table}[!htbp]
\centering\footnotesize
\setlength{\tabcolsep}{3.5pt}
\caption{Cross-dataset pixel-level faithfulness summary: \model{} native map (deletion / insertion AUC) versus random reference and the most faithful post-hoc method per dataset (representative run, seed 42).}\label{tab:xai_cross}
\begin{tabular}{@{}l c c c c c c@{}}
\toprule
 & \textbf{Native} & \textbf{Random} & \textbf{Best post-hoc} & \textbf{Best post-hoc} & \multicolumn{2}{c}{\textbf{Native map}}\\
\cmidrule(l){6-7}
\textbf{Dataset} & \textbf{del.\ / ins.} & \textbf{del.\ / ins.} & \textbf{deletion $\downarrow$} & \textbf{insertion $\uparrow$} & \textbf{NP@10\%} & \textbf{Stability}\\
\midrule
BreakHis & 0.828 / 0.863 & 0.842 / 0.844 & 0.779 (LIME) & 0.889 (LIME) & 0.376 & 0.604\\
LC25000 & 0.735 / 0.854 & 0.778 / 0.784 & 0.619 (IG) & 0.924 (LIME) & 0.305 & 0.897\\
Lymphoma & 0.610 / 0.656 & 0.528 / 0.531 & 0.443 (IG) & 0.767 (Occlusion) & 0.462 & 0.868\\
Kather2016 & 0.687 / 0.718 & 0.708 / 0.706 & 0.541 (LIME) & 0.794 (LIME) & 0.321 & 0.887\\
NCT-CRC-HE-100K & 0.581 / 0.707 & 0.703 / 0.701 & 0.467 (IG) & 0.808 (LIME) & 0.465 & 0.788\\
Oral cancer & 0.757 / 0.746 & 0.765 / 0.771 & 0.629 (LIME) & 0.830 (LIME) & 0.216 & 0.707\\
\bottomrule
\end{tabular}
\end{table}

\begin{table}[!htbp]
\centering\footnotesize
\setlength{\tabcolsep}{3pt}
\caption{Graph-level explanation tests and stain-shift stability (representative run, seed 42). $\Delta$AUC: normalised confidence-curve AUC of importance-ranked removal minus random removal (negative = faster confidence drop than random); $p$: paired Wilcoxon test.}\label{tab:xai_graph}
\begin{tabular}{@{}l r c c c c c c c@{}}
\toprule
 & & \multicolumn{2}{c}{\textbf{Edge deletion}} & \multicolumn{2}{c}{\textbf{Compartment occlusion ($r_m$)}} & \textbf{Area-ranked} & \textbf{Spearman of $r$} & \textbf{Prediction}\\
\cmidrule(lr){3-4}\cmidrule(lr){5-6}
\textbf{Dataset} & $n$ & $\Delta$\textbf{AUC} & $p$ & $\Delta$\textbf{AUC} & $p$ & $\Delta$\textbf{AUC} & \textbf{under shift} & \textbf{unchanged (\%)}\\
\midrule
BreakHis & 40 & -0.0003 & 0.354 & -0.0250 & 0.716 & -0.0660 & 0.941 $\pm$ 0.041 & 100.0\\
LC25000 & 40 & +0.0003 & 0.267 & -0.0091 & 0.656 & -0.2238 & 0.970 $\pm$ 0.029 & 97.5\\
Lymphoma & 39 & -0.0007 & 0.013 & -0.0358 & 0.135 & -0.2658 & 0.993 $\pm$ 0.005 & 89.7\\
Kather2016 & 40 & -0.0149 & $<0.001$ & +0.0205 & 0.491 & -0.1647 & 0.980 $\pm$ 0.033 & 100.0\\
NCT-CRC-HE-100K & 36 & -0.0001 & 0.810 & -0.0044 & 0.762 & -0.3084 & 0.926 $\pm$ 0.096 & 100.0\\
Oral cancer & 40 & -0.0031 & $<0.001$ & -0.0172 & 0.307 & -0.1019 & 0.942 $\pm$ 0.057 & 97.5\\
\bottomrule
\end{tabular}
\end{table}

%% ---------------------------------------------------------------------
\subsection{Calibration and selective prediction}\label{sec:r_calib}

\begin{figure}[!htbp]
\centering
\begin{subfigure}[t]{0.32\textwidth}\centering\includegraphics[width=\linewidth]{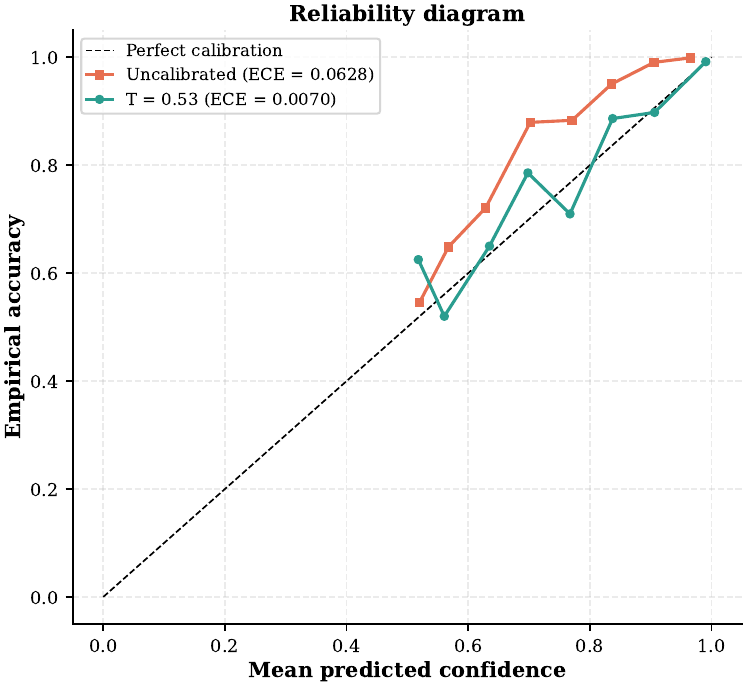}\caption{BreakHis}\end{subfigure}\hfill
\begin{subfigure}[t]{0.32\textwidth}\centering\includegraphics[width=\linewidth]{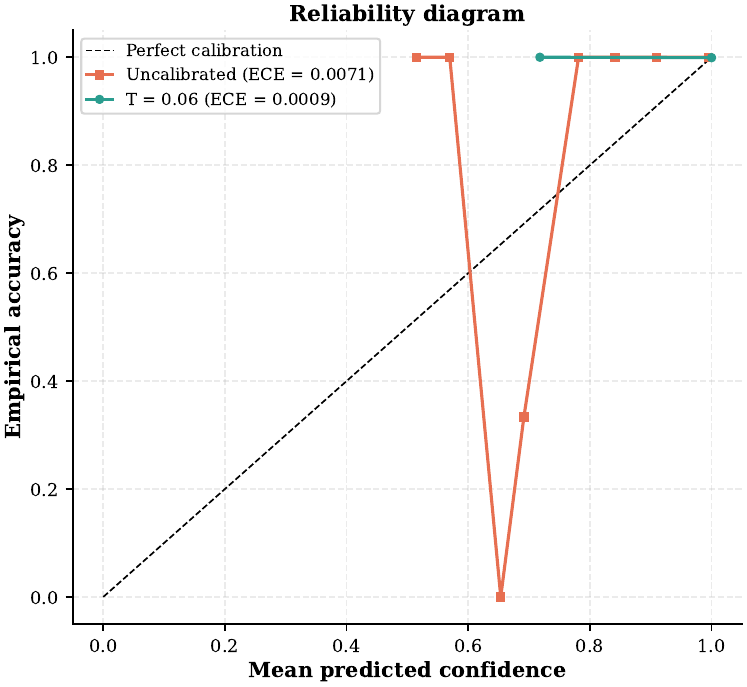}\caption{LC25000}\end{subfigure}\hfill
\begin{subfigure}[t]{0.32\textwidth}\centering\includegraphics[width=\linewidth]{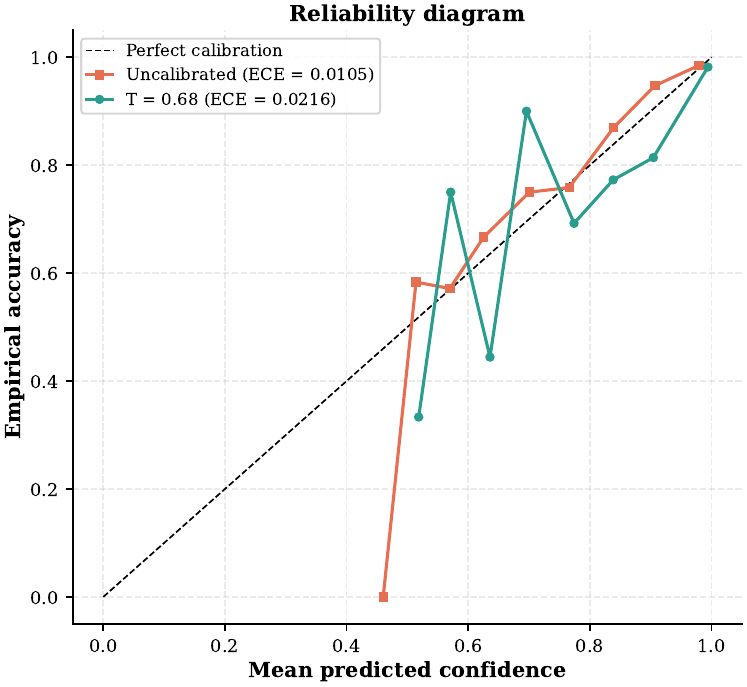}\caption{Lymphoma}\end{subfigure}\\[4pt]
\begin{subfigure}[t]{0.32\textwidth}\centering\includegraphics[width=\linewidth]{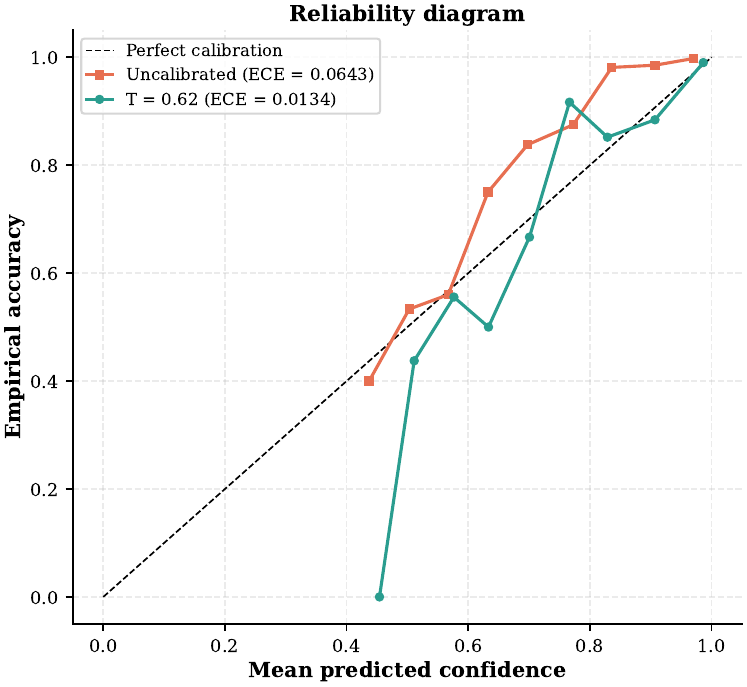}\caption{Kather2016}\end{subfigure}\hfill
\begin{subfigure}[t]{0.32\textwidth}\centering\includegraphics[width=\linewidth]{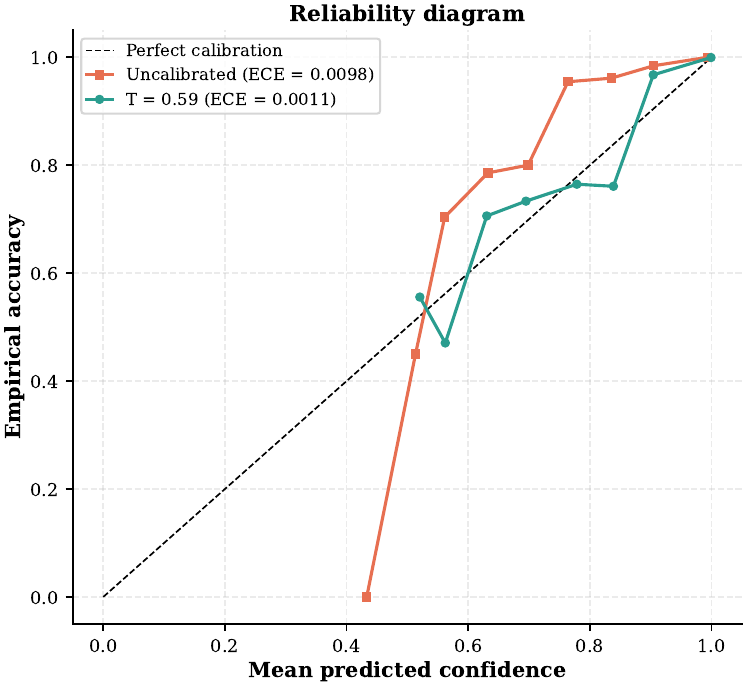}\caption{NCT-CRC-HE-100K}\end{subfigure}\hfill
\begin{subfigure}[t]{0.32\textwidth}\centering\includegraphics[width=\linewidth]{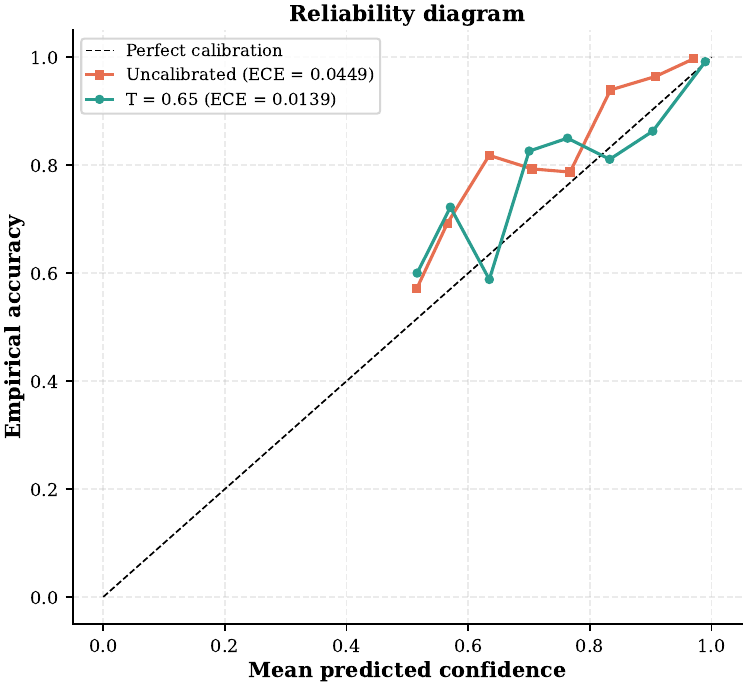}\caption{Oral cancer}\end{subfigure}
\caption{Reliability diagrams of \model{} on the six test sets post-temperature scaling.}
\label{fig:reliability}
\end{figure}

\Cref{tab:calib} and \cref{fig:reliability} report probability calibration before and after validation-fitted temperature scaling. Focal loss ($\gamma=2$) produces underconfident raw probabilities ($T^{\star}<1$ on all datasets). Post-scaling ECE falls to $\le 1.39\%$ on five of the six datasets, but temperature scaling made ECE worse on Lymphoma (from 1.05\% to 2.16\%) and also raised NLL on Lymphoma and LC25000 (\cref{tab:calib}). Notably, on Kather2016, while the global Expected Calibration Error (ECE) improves substantially from 6.43\% to 1.34\%, the Maximum Calibration Error (MCE) increases to 45.40\% post-scaling; this mathematical artifact occurs because global temperature scaling pulls the bulk distribution into alignment but pushes a very small number of overconfident, incorrect outlier predictions into a high-error bin at the extreme edge of the probability simplex. In selective prediction (\cref{tab:selective}), referring the 5\% most uncertain test tiles raises accuracy by 1.37--1.59\,pp on the four non-saturated benchmarks, and referring 20\% brings accuracy to $\ge 98.71\%$ across all six datasets.

\begin{table}[!htbp]
\caption{Calibration and selective prediction. \textbf{(A)}~Calibration before and after validation-fitted temperature scaling (representative run, seed 42; 15 equal-width bins; ECE and MCE in \%). \textbf{(B)}~Selective prediction: accuracy (\%) at decreasing coverage and area under the risk--coverage curve (AURC).}\label{tab:calib}\label{tab:selective}
{\centering\small\textbf{(A)}\par}\smallskip
{\centering\footnotesize
\setlength{\tabcolsep}{3.5pt}

\begin{tabular}{@{}l c cc cc cc cc@{}}
\toprule
 & & \multicolumn{2}{c}{\textbf{ECE}} & \multicolumn{2}{c}{\textbf{MCE}} & \multicolumn{2}{c}{\textbf{NLL}} & \multicolumn{2}{c}{\textbf{Brier}}\\
\cmidrule(lr){3-4}\cmidrule(lr){5-6}\cmidrule(lr){7-8}\cmidrule(l){9-10}
\textbf{Dataset} & $T^{\star}$ & pre & post & pre & post & pre & post & pre & post\\
\midrule
BreakHis & 0.532 & 6.28 & 0.70 & 17.74 & 10.71 & 0.1474 & 0.1062 & 0.0707 & 0.0598\\
LC25000 & 0.063 & 0.71 & 0.09 & 65.34 & 28.22 & 0.0079 & 0.0094 & 0.0017 & 0.0017\\
Lymphoma & 0.678 & 1.05 & 2.16 & 46.01 & 20.43 & 0.1671 & 0.1918 & 0.0679 & 0.0676\\
Kather2016 & 0.616 & 6.43 & 1.34 & 14.52 & 45.40 & 0.1913 & 0.1550 & 0.0869 & 0.0774\\
NCT-CRC-HE-100K & 0.588 & 0.98 & 0.11 & 43.31 & 9.16 & 0.0167 & 0.0094 & 0.0053 & 0.0044\\
Oral cancer & 0.651 & 4.49 & 1.39 & 18.38 & 15.17 & 0.1479 & 0.1268 & 0.0777 & 0.0730\\
\bottomrule
\end{tabular}\par}
\par\medskip{\centering\small\textbf{(B)}\par}\smallskip
{\centering\footnotesize
\setlength{\tabcolsep}{4pt}

\begin{tabular}{@{}l c c c c c@{}}
\toprule
\textbf{Dataset} & \textbf{100\%} & \textbf{95\%} & \textbf{90\%} & \textbf{80\%} & \textbf{AURC}\\
\midrule
BreakHis & 96.07 & 97.62 & 98.53 & 99.29 & 0.0105\\
LC25000 & 99.92 & 100.00 & 100.00 & 100.00 & 0.0001\\
Lymphoma & 95.87 & 97.24 & 98.13 & 99.02 & 0.0079\\
Kather2016 & 94.83 & 96.42 & 97.53 & 98.71 & 0.0098\\
NCT-CRC-HE-100K & 99.72 & 99.90 & 99.95 & 99.98 & 0.0004\\
Oral cancer & 95.51 & 96.94 & 97.88 & 98.92 & 0.0082\\
\bottomrule
\end{tabular}\par}
\end{table}

%% ---------------------------------------------------------------------
\subsection{Robustness to stain and acquisition drift}\label{sec:r_robust}

\begin{figure}[!htbp]
\centering
\begin{subfigure}[t]{0.32\textwidth}\centering\includegraphics[width=\linewidth]{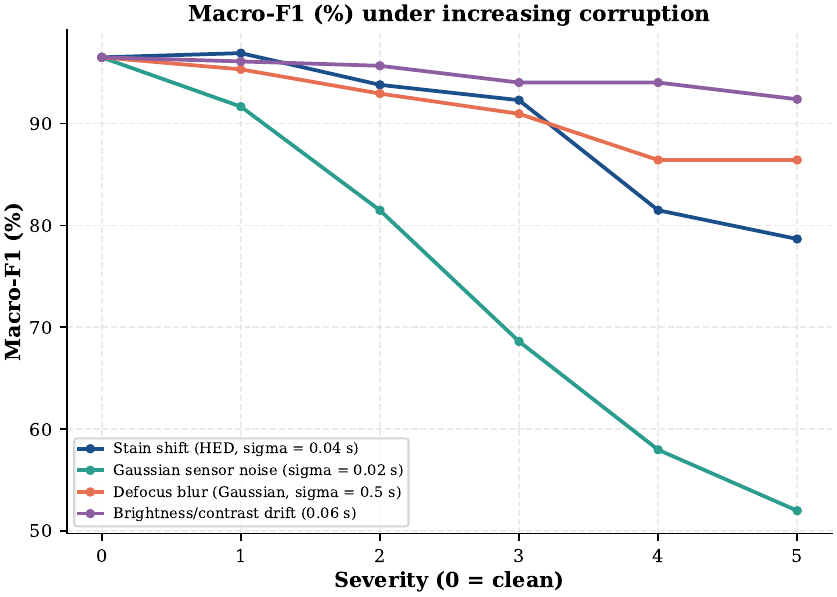}\caption{BreakHis}\end{subfigure}\hfill
\begin{subfigure}[t]{0.32\textwidth}\centering\includegraphics[width=\linewidth]{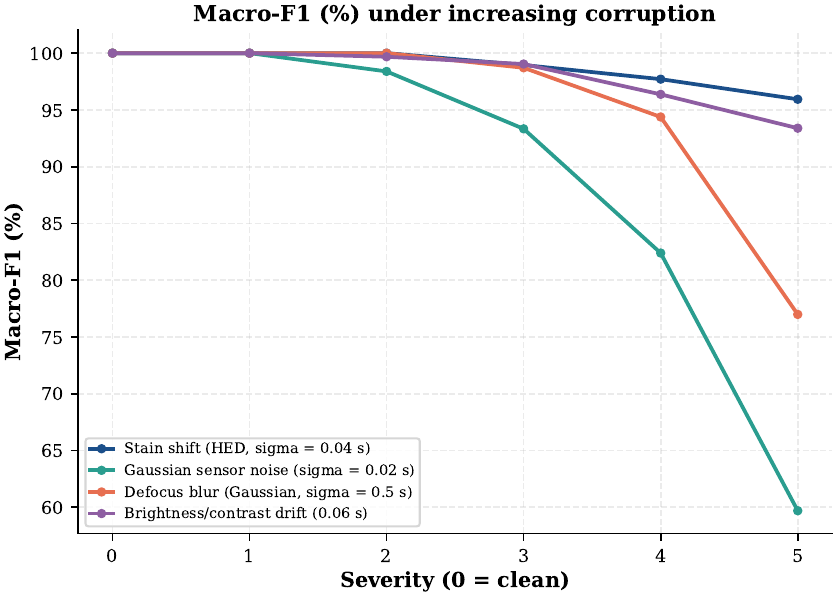}\caption{LC25000}\end{subfigure}\hfill
\begin{subfigure}[t]{0.32\textwidth}\centering\includegraphics[width=\linewidth]{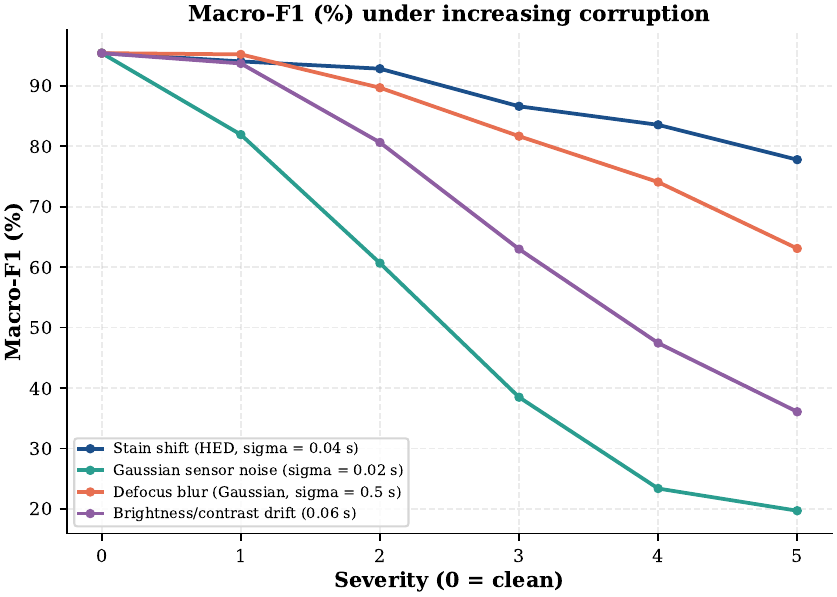}\caption{Lymphoma}\end{subfigure}\\[4pt]
\begin{subfigure}[t]{0.32\textwidth}\centering\includegraphics[width=\linewidth]{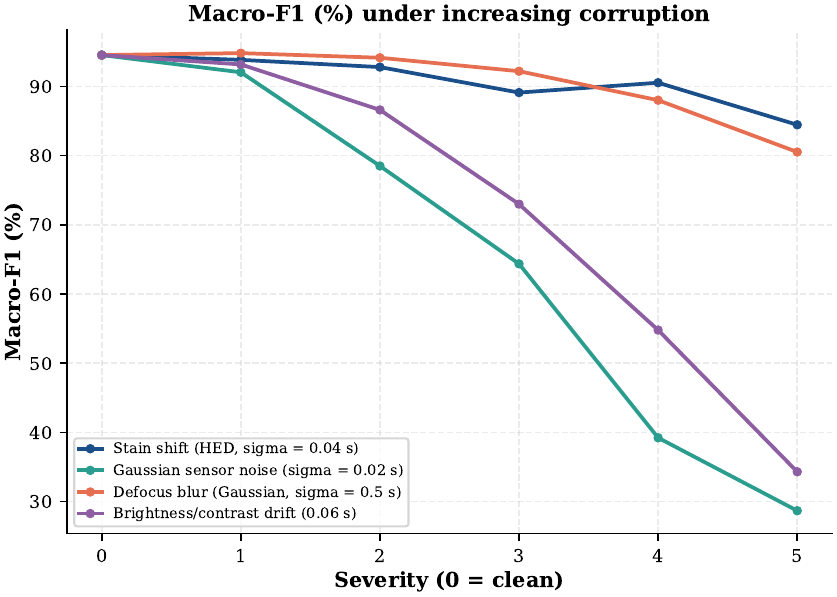}\caption{Kather2016}\end{subfigure}\hfill
\begin{subfigure}[t]{0.32\textwidth}\centering\includegraphics[width=\linewidth]{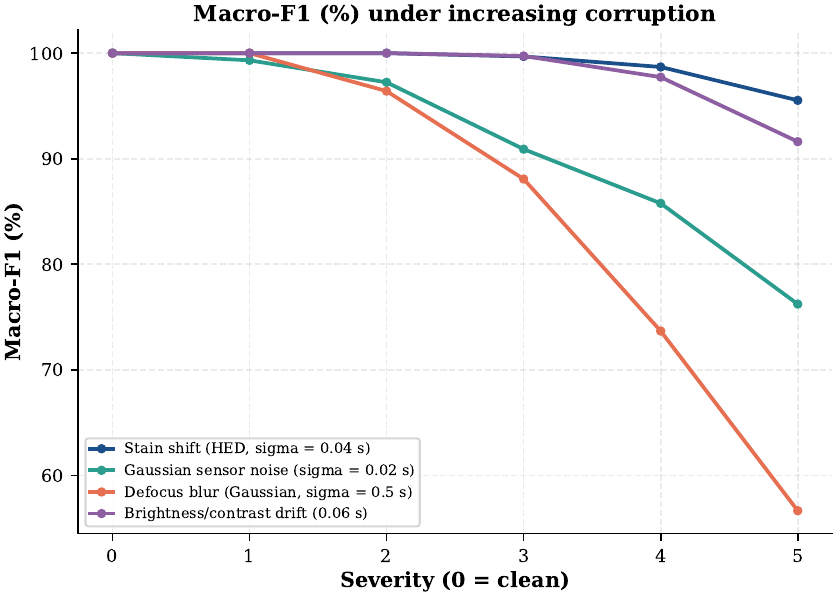}\caption{NCT-CRC-HE-100K}\end{subfigure}\hfill
\begin{subfigure}[t]{0.32\textwidth}\centering\includegraphics[width=\linewidth]{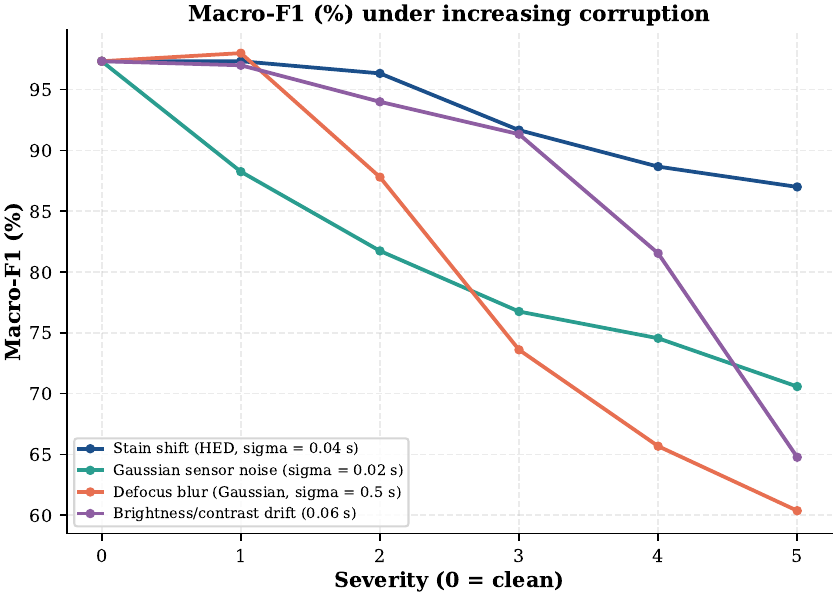}\caption{Oral cancer}\end{subfigure}
\caption{Macro-F1 of \model{} (trained without noise augmentation) against corruption severity across four corruption families.}
\label{fig:robust}
\end{figure}

\Cref{tab:robust} and \cref{fig:robust} evaluate macro-F1 across four corruption families (5 severities each) on 300 test tiles per dataset, and \cref{tab:robust_base} compares BreakHis robustness against Swin-Tiny, ConvNeXt-Tiny and CTransPath. Under standard training (without Gaussian noise augmentation), \model{} loses 1.2--1.5\,pp under synthetic HED stain shifts on LC25000 and NCT-CRC-HE-100K, 4.4--5.1\,pp on Kather2016 and Oral Cancer, and larger amounts on BreakHis (7.9\,pp) and Lymphoma (8.5\,pp), but degrades under high-frequency Gaussian sensor noise, which perturbs the $8\times8$ optical-density unmixing in Stage~0 before messages can be passed. Including additive Gaussian noise augmentation ($\sigma\sim\mathcal{U}(0,0.03)$, $p=0.5$) during training heavily mitigates this high-frequency sensor vulnerability (e.g., raising Lymphoma noise robustness from 44.83\% to 92.35\%) without substantially altering clean accuracy. In the BreakHis comparison (\cref{tab:robust_base}), \model{} degrades less than Swin-Tiny and ConvNeXt-Tiny under every stain condition, but the CTransPath architecture is more robust than \model{} (trained without noise augmentation) under every condition, both in absolute macro-F1 and in the drop from clean data (joint stain drift: $-2.9$ vs.\ $-4.2$\,pp).

\paragraph{Stain drift is where built-in stain physics pays off.} Two observations make robustness to stain variation the most defensible practical advantage of \model{}. First, within \cref{tab:robust}A, HED stain shift is the corruption family that \model{} tolerates best on five of the six datasets. The loss averaged over severities 1--5 is 1.22--1.48\,pp on NCT-CRC-HE-100K and LC25000 and 4.38--8.45\,pp on Kather2016, Oral Cancer and Lymphoma. These are smaller than the losses under sensor noise (10.11--50.58\,pp), defocus blur (4.59--20.24\,pp) and, where it is large, brightness/contrast drift. BreakHis is the exception, where brightness/contrast (2.06\,pp) and defocus (6.09\,pp) are tolerated better than stain shift (7.87\,pp). This pattern is expected: Stage~0 re-estimates the stain basis for every image, so a change in dye intensity or hue moves the optical-density maps but largely preserves their spatial structure, on which the edge penalties operate. Second, in the controlled BreakHis comparison (\cref{tab:robust_base}), averaged over the haematoxylin, eosin and joint $\pm30\%$ conditions, \model{} loses $2.88$\,pp from clean macro-F1, against $4.37$\,pp for Swin-Tiny (5.2 times more parameters) and $4.67$\,pp for ConvNeXt-Tiny (5.2 times more). This is a 34\% and 38\% smaller loss, respectively. Swin-Tiny is 0.50\,pp better than \model{} on clean BreakHis, yet \model{} is better under every stain condition (by 1.04--1.11\,pp), so stain drift reverses the ranking of the two models. The same property appears in the explanations: the compartment-importance rankings stay highly stable under synthetic stain shift (Spearman 0.926--0.993, \cref{sec:r_xai}). We do not claim universal stain robustness. The CTransPath architecture, which uses a convolutional stem in front of a Swin backbone and is also trained from scratch here, degrades less still ($2.23$\,pp), and we have not yet measured the robustness of the RGB-only and fixed-basis ablations, which would isolate how much of this advantage is due to Stage~0 itself (\cref{sec:d_limitations}).

\begin{table}[!htbp]
\caption{Robustness to stain and acquisition drift. \textbf{(A)}~Robustness of \model{}: macro-F1 (\%) on the clean 300-tile subset and averaged over severities 1--5 for each corruption family (representative run, seed 42). Columns evaluating Gaussian noise augmentation are shown distinctly. \textbf{(B)}~Robustness comparison on BreakHis (macro-F1, \%). The CTransPath architecture is more robust than \model{} (without noise augmentation) under every condition. The final column denotes \model{} trained with Gaussian sensor-noise augmentation.}\label{tab:robust}\label{tab:robust_base}
{\centering\small\textbf{(A)}\par}\smallskip
{\centering\footnotesize
\setlength{\tabcolsep}{3.5pt}

\begin{tabular}{@{}l c c c c c c@{}}
\toprule
\textbf{Dataset} & \textbf{Clean} & \textbf{HED stain shift} & \textbf{Gaussian noise (no/with aug.)} & \textbf{Defocus blur} & \textbf{Bright./contrast} & \textbf{Mean (no/with aug.)}\\
\midrule
BreakHis & 96.50 & 88.63 & 70.33 / 94.12 & 90.41 & 94.44 & 85.96 / 92.82\\
LC25000 & 100.00 & 98.52 & 86.76 / 99.55 & 94.01 & 97.70 & 94.25 / 97.96\\
Lymphoma & 95.41 & 86.96 & 44.83 / 92.35 & 80.74 & 64.16 & 69.17 / 83.92\\
Kather2016 & 94.52 & 90.14 & 60.55 / 93.10 & 89.93 & 68.36 & 77.24 / 87.21\\
NCT-CRC-HE-100K & 100.00 & 98.78 & 89.89 / 99.15 & 82.96 & 97.81 & 92.36 / 95.74\\
Oral cancer & 97.33 & 92.20 & 78.38 / 95.20 & 77.09 & 85.73 & 83.35 / 89.51\\
\bottomrule
\end{tabular}\par}
\par\medskip{\centering\small\textbf{(B)}\par}\smallskip
{\centering\footnotesize
\setlength{\tabcolsep}{4pt}

\begin{tabular}{@{}l c c c c c@{}}
\toprule
\textbf{Condition} & \textbf{Swin-Tiny} & \textbf{ConvNeXt-Tiny} & \textbf{CTransPath (arch.)} & \textbf{\model{}} & \textbf{\model{} (+noise aug.)}\\
\midrule
Clean & 95.98 & 93.64 & 97.11 & 95.48 & 95.25\\
Haematoxylin shift ($\pm30\%$) & 91.91 & 89.32 & 95.06 & 93.02 & 92.85\\
Eosin shift ($\pm30\%$) & 92.43 & 89.97 & 95.38 & 93.47 & 93.15\\
Joint stain drift ($\pm30\%$) & 90.49 & 87.62 & 94.21 & 91.30 & 91.10\\
Gaussian sensor noise & 91.36 & 88.77 & 94.76 & 90.66 & 94.55\\
Defocus blur ($3\times3$) & 92.06 & 89.47 & 95.01 & 93.08 & 92.80\\
\bottomrule
\end{tabular}\par}
\end{table}

%% =====================================================================
%% 5. DISCUSSION
%% =====================================================================
\section{Discussion}\label{sec:discussion}

\subsection{Principal findings}\label{sec:d_findings}

This study examined four questions: whether a lightweight tile classifier can learn a patch-to-compartment hierarchy end to end without offline nuclei segmentation; whether physical H\&E stain optical densities can be embedded directly into local message passing and differentiable node pooling; how strongly near-duplicate leakage inflates random-split histopathology benchmarks; and how the learned graph structure can be inspected quantitatively. Our results support three conclusions. First, conditioning windowed local attention~\cite{liu2021swin} and differentiable soft pooling~\cite{ying2018hierarchical,locatello2020object} on per-image Macenko optical densities yields competitive accuracy across six benchmarks within a 5.33\,M-parameter budget, achieving $91.84\pm2.37\%$ under strict patient-grouped cross-validation on BreakHis and $96.84\pm0.14\%$ external accuracy without adaptation on CRC-VAL-HE-7K. Second, random tile splits suffer from severe near-duplicate or same-source leakage on LC25000 (86.9\%), Lymphoma (100\%) and Oral Cancer (35.9\%). Third, the 72 learned stain priors and the compartment topologies provide \emph{structural transparency}, in the sense that the parameters and graphs can be read out and measured quantitatively: the graph topology alone encodes diagnostic class, and the compartment rankings are stable under stain shift. Across these findings, the evidence points to stain conditioning as the operative idea. It contributes at least as much accuracy as the whole compartment hierarchy, and this contribution comes almost entirely from the per-image H\&E optical-density input rather than from the stain penalty, whose effect on accuracy is within seed variation (\cref{sec:r_ablation}). Its learned priors differ between datasets in a way that is consistent with the tissue feature that defines each task, although this was observed in a single run and not validated by pathologists (\cref{sec:r_graphs}). Under stain drift, \model{} degrades 34--38\% less than Swin-Tiny and ConvNeXt-Tiny (\cref{sec:r_robust}). We regard the leakage audit and \ssip{} as a finding of equal weight: independent of any architecture, they change how the reported accuracies on three widely used benchmarks should be read.

\subsection{Why stain-conditioned nested graphs help, and where design intuition failed}\label{sec:d_why}

The trajectories of the learned stain penalties $\sigma_+(\beta_H)$ and $\sigma_+(\beta_E)$ (\cref{tab:priors}) suggest that the network adapts its relational inductive bias to tissue composition, although this rests on a single representative run and has not been checked against pathologist annotations. On epithelial tasks defined by nuclear pleomorphism (BreakHis, LC25000), the haematoxylin penalty dominates from block 1 while the eosin penalty is suppressed early. Conversely, on colorectal tissue classification (Kather2016) and oral squamous cell carcinoma (where stromal matrix, mucin and eosinophilic keratin pearls carry diagnostic signal), the eosin penalty overtakes the haematoxylin penalty in blocks 3 and 4. Spearman rank correlations across 300{,}000 edges per block (\cref{tab:edgerho}) are consistent with stain-coherent attention emerging progressively with depth, although a negative correlation is expected by construction of \cref{eq:attn}. Read together with the ablations, these patterns argue that the gains of \model{} come chiefly from where stain information is injected rather than from the depth of the hierarchy. The H\&E representation supplies most of the accuracy (\cref{sec:r_ablation}), and the per-dye, per-head penalties turn that representation into parameters whose values a pathologist can check against known stain chemistry.

At the same time, two architectural behaviours departed from our initial design expectations. First, the spatial--semantic outer gate $g_h$ remained near 0.50 across all datasets (\cref{tab:priors}), indicating that the compartment graph relies on an approximately equal blend of grid-contact spatial adjacency and semantic attention rather than specialising strictly per dataset. Second, while compartments sharpen into discrete epithelial and stromal regions on BreakHis (\cref{tab:layer}), they remain diffuse on Lymphoma and Kather2016, where individual tiles consist of uniform lymphocyte sheets or single tissue types lacking internal architectural boundaries. This indicates the risk of forcing rigid compartment assumptions on homogeneous tiles, a dynamic confirmed by the non-monotonic ablation drops when pooling is applied without robust adjacency interactions (\cref{tab:ablation_step}).

\subsection{Benchmark validity, near-duplicate isolation (\ssip{}) and patient-level splitting}\label{sec:d_leakage}

Our leakage audit highlights a critical methodological issue in public tile benchmarks: datasets constructed by offline augmentation or tiling of a small image pool (LC25000 from 1{,}250 originals, Lymphoma from 374 source images, Oral Cancer from 978 originals~\cite{diniz2025oscc}) exhibit 35.9\%--100\% near-duplicate or same-source overlap under standard random splitting.

At the same time, we emphasise the exact boundary of what perceptual-hash grouping (\ssip{}) can and cannot guarantee:
\begin{enumerate}
  \item \emph{Internal consistency vs.\ completeness:} Because the audit metric $\lambda$ (\cref{eq:leakage}) uses the same $D_4$-pHash threshold as \cref{alg:ssip}, $\lambda=0\%$ under \ssip{} holds by construction. On LC25000, \ssip{} collapses 25{,}000 tiles into 7{,}680 connected components—where 86.9\% is the random-split leakage rate, i.e.\ the share of test tiles with a detected near-duplicate in training under a random split—but cannot merge arbitrary-angle continuous rotations back to the 1{,}250 original images. Hence, the 99.88\% accuracy on LC25000 still reflects residual same-source similarity.
  \item \emph{Near-duplicates vs.\ patient isolation:} Perceptual hashing groups visually overlapping images, not distinct fields of view from the same patient. On BreakHis, comparing image-level \ssip{} ($95.71\pm0.62\%$) against strict 81-patient cross-validation ($91.84\pm2.37\%$, \cref{tab:cv}) directly quantifies the $\approx$3.9\,pp patient-level correlation effect. Similarly, on colorectal tissue, the external evaluation without adaptation on CRC-VAL-HE-7K ($96.84\pm0.14\%$, \cref{tab:external}) provides the true patient- and laboratory-independent generalisation estimate.
\end{enumerate}

\begingroup\revon
\paragraph{Why the split unit differs between datasets.} The choice of split unit follows from the metadata, not from the results. Where a public release records which patient or source image a tile came from (BreakHis, Lymphoma, and the patient-disjoint CRC-VAL-HE-7K cohort), we split on that identifier, because no pixel-based method can recover patient identity as reliably. Where the release records no identifier (LC25000, Kather2016, NCT-CRC-HE-100K, Oral Cancer), a patient split is impossible, and the alternative used in most prior work, a random tile split, leaves 35.9--86.9\% of test tiles with a near-duplicate in training on the augmented datasets. \ssip{} is the strongest separation that the pixels alone permit: it groups every detectable rotated, flipped or re-cropped copy, transitively, before any partition is drawn. All later steps that could reintroduce test information, namely preprocessing, augmentation, hyperparameter and checkpoint selection, temperature fitting and decision thresholds, use training or validation data only (\cref{tab:leakctrl}). The remaining gap between \ssip{} and patient isolation is therefore one of grouping granularity, which we measure directly on BreakHis ($\approx$3.9\,pp) and report as a limitation, rather than a pathway through which test labels enter model development.\par
\endgroup

\paragraph{Implications for published benchmark results.} The audit is not a footnote to the architecture but a result in its own right, and it does not depend on \model{}. Under a random tile split, 86.9\% of LC25000 test tiles, every Lymphoma test tile and 35.9\% of Oral Cancer test tiles have a near-duplicate or same-source partner in the training set (\cref{tab:ssip}). As far as their protocols are reported, most of the near-ceiling accuracies on these benchmarks were obtained under such random splits (\cref{sec:rw_tile}). Where that is the case, for example for the 99.20--99.62\% reported on LC25000~\cite{hasan2024lwms,jim2025xllcnet}, the 99--99.87\% on lymphoma~\cite{rajadurai2024precisionlymphonet,sikkandar2025hctnlc} and the 98.33\% on oral squamous cell carcinoma~\cite{islam2024oscclime}, the reported accuracy cannot be separated from recognition of near-identical training images. Such numbers should not be compared with results obtained under source-separated splits, and differences of a few tenths of a percentage point between them carry little information. Where leakage is low (BreakHis 2.65\%, Kather2016 1.70\%, NCT-CRC-HE-100K 0.51\%), random and \ssip{} splits are nearly equivalent, which makes the audit informative in both directions: it identifies which benchmarks need source-separated splits and which do not. Because \ssip{} needs only the images, it can be applied retroactively to any tile archive without patient metadata. The notebooks that regenerate the group assignments and split indices for all six datasets are available on Zenodo (Data and code availability), so that future studies can report results on identical leakage-free partitions.

\subsection{Structural transparency versus post-hoc explanation faithfulness}\label{sec:d_xai}

Our explanation audit (\cref{sec:r_xai}) shows that structural transparency and post-hoc attribution play complementary roles. \model{} is structurally transparent: its 72 scalar priors have a direct physical interpretation in Beer--Lambert optical-density space, its learned graph topology alone encodes 89.7\%--99.1\% diagnostic accuracy (\cref{tab:probe}), and its compartment weight rankings are highly invariant to stain shifts (Spearman 0.93--0.99). The native map places more of its top-10\% attribution on nuclear pixels than chance (NP@10\% = 0.376 vs.\ 0.300 on BreakHis); the NP@10\% of 1.000 in \cref{tab:xai_bh} belongs to the haematoxylin-density reference row, which is nuclear by construction, not to \model{}.

For practical deployment, this implies a clear functional division: pathologists and developers can use \model{}'s native compartment maps and edge priors for structural inspection of how tissue is organised and for monitoring stability under stain shift, and pixel-level attribution methods such as LIME or Integrated Gradients when identifying the exact pixels that drive an individual prediction is required.

\subsection{Computational complexity and clinical deployment considerations}\label{sec:d_complexity}\label{sec:d_positioning}

Every operator in \model{} scales linearly with the number of cell nodes $N$: windowed cell attention costs $\mathcal{O}(Nw^2d)$, the convolutional FFN costs $\mathcal{O}(Nd^2)$, soft assignment and FiLM broadcast cost $\mathcal{O}(NMd)$, and the coarse compartment graph costs $\mathcal{O}(M^2d+Md^2)$ independently of $N$. Compared with global ViTs ($\mathcal{O}(N^2d)$) and 300\,M--632\,M parameter pathology foundation models (\cref{tab:positioning}), \model{} trains in 3.6\,GB of VRAM on a single entry-level Tesla T4 GPU, achieves calibrated probabilities (post-scaling ECE $\le 2.16\%$) suitable for confidence-based clinical referral (\cref{tab:selective}), and requires only 5.33\,M parameters.

\begin{table}[!htbp]
\centering\footnotesize
\setlength{\tabcolsep}{3.5pt}
\caption{Positioning against published results on the same datasets. Published values are as reported under the original authors' splitting protocols; \model{} is reported across leakage-audited image-level (\ssip{}), source-image, patient-grouped and external-cohort protocols.}\label{tab:positioning}
\begin{tabularx}{\textwidth}{@{}l l X r l r@{}}
\toprule
\textbf{Dataset} & \textbf{Study} & \textbf{Model} & \textbf{Params} & \textbf{Split protocol} & \textbf{Accuracy (\%)}\\
\midrule
LC25000 (5-class) & \citet{hasan2024lwms} & LW-MS CNN & 1.1\,M & random, tile level & 99.20\\
LC25000 (lung, 3-class) & \citet{jim2025xllcnet} & XLLC-Net & 3\,M & random, tile level & $99.62 \pm 0.16$\\
BreakHis (binary) & \citet{ranjbar2024dct} & DCT-HistoTransformer & 7.97\,M & not reported & 96.00\\
NCT-CRC-HE-100K & \citet{abdelghany2024efficientnet} & EfficientNet-B0 & $\approx$5.3\,M & internal / CRC-VAL-7K & 99.87 / 99.00\\
Lymphoma & \citet{rajadurai2024precisionlymphonet} & PrecisionLymphoNet & $\approx$47\,M & random, tile level & 99.00\\
Lymphoma & \citet{sikkandar2025hctnlc} & HCTN-LC & Not Reported & no source-level split & 99.87\\
Oral cancer & \citet{islam2024oscclime} & EfficientNetB3 + LIME & $\approx$12\,M & provided split (126 test) & 98.33\\
\midrule
CRC-100K (raw) & \citet{chen2024uni} & UNI (ViT-L, foundation) & $\approx$300\,M & linear probe, released split & 92.5$^{\ddagger}$\\
CRC-100K (raw) & \citet{lu2024conch} & CONCH (foundation) & 90\,M + 110\,M & linear probe, released split & 94.1$^{\ddagger}$\\
Pan-cancer & \citet{vorontsov2024virchow} & Virchow (foundation) & 632\,M & specimen level & AUC 0.95\\
\midrule
BreakHis (patient-CV) & \multicolumn{2}{l}{\model{} (this work)} & 5.33\,M & 5-fold patient-grouped (81 pat.) & $91.84 \pm 2.37$\\
BreakHis (\ssip{}) & \multicolumn{2}{l}{\model{} (this work)} & 5.33\,M & \ssip{} pHash groups (5 seeds) & $95.92 \pm 0.38$\\
CRC-VAL-HE-7K (ext.) & \multicolumn{2}{l}{\model{} (this work)} & 5.33\,M & external, no adaptation (50 pat.) & $96.84 \pm 0.14$\\
NCT-CRC-HE-100K & \multicolumn{2}{l}{\model{} (this work)} & 5.33\,M & \ssip{} pHash groups (5 seeds) & $99.69 \pm 0.03$\\
LC25000 (5-class) & \multicolumn{2}{l}{\model{} (this work)} & 5.33\,M & \ssip{} pHash groups (5 seeds) & $99.88 \pm 0.04$\\
Lymphoma & \multicolumn{2}{l}{\model{} (this work)} & 5.33\,M & source-image split (tile / image) & $95.41 \pm 0.72$ / 98.67$^{\dagger}$\\
Kather2016 & \multicolumn{2}{l}{\model{} (this work)} & 5.33\,M & \ssip{} pHash groups (5 seeds) & $94.66 \pm 0.51$\\
Oral cancer & \multicolumn{2}{l}{\model{} (this work)} & 5.33\,M & \ssip{} pHash groups (5 seeds) & $95.28 \pm 0.61$\\
\bottomrule
\multicolumn{6}{@{}l}{\scriptsize $^{\dagger}$5-seed tile-level accuracy / seed-42 source-image-aggregated accuracy (75 test source images). $^{\ddagger}$Balanced accuracy on unnormalised CRC-100K.}\\
\end{tabularx}
\end{table}

\subsection{Limitations and future work}\label{sec:d_limitations}\label{sec:d_future}

This study has several limitations. The stain penalty's contribution to accuracy is within seed noise (V2$\to$V3 in \cref{tab:ablation_step}: $+0.05$ to $+0.25$\,pp), so its value is mainly interpretive. On BreakHis, Kather2016 and NCT-CRC-HE-100K, \ssip{} is close to an image-level split, and $D_4$-pHash cannot recover arbitrary-angle rotations or non-overlapping tiles of one slide; only the BreakHis patient-grouped (81 identifiers against 82 reported patients) and CRC-VAL-HE-7K results isolate patients. The foundation models were only linear-probed, no baselines were run under patient-grouped cross-validation, and only ResNet-50 and ConvNeXt-Tiny were evaluated on the external cohort. We used the Macenko-normalised NCT-CRC-HE-100K release, which reduces stain variation and carries known colour and compression shortcuts~\cite{ignatov2024nctcrche}. Finally, \model{} operates on $224\times224$ tiles, several analyses come from a single representative run, temperature scaling worsened calibration on Lymphoma, the CTransPath architecture is more robust to stain shift on BreakHis, and Stage~0 unmixing is sensitive to Gaussian noise without noise augmentation. The robustness comparison does not include the Stage-0 ablations (RGB only, fixed basis), so the share of the stain-drift advantage attributable to per-image unmixing has not been isolated. The histological interpretation of the learned priors rests on a single seed and on dataset-level class definitions rather than on pathologist annotation.

Future work will run all baselines under patient-grouped cross-validation and on CRC-VAL-HE-7K, fine-tune the foundation encoders, retrain on the NCT-CRC-HE-100K-NONORM release and repeat every single-run analysis over five seeds, including the converged stain priors and the stain-drift evaluation of the Stage-0 ablations. Because block complexity is $\mathcal{O}(N)$, linking compartment nodes across adjacent tiles into a slide-level graph or multiple-instance aggregator on multi-centre cohorts with patient and site metadata (such as TCGA or CAMELYON) is a natural next step, together with reader studies with board-certified pathologists to validate the histological correspondence of the learned compartments.

%% =====================================================================
%% 6. CONCLUSION
%% =====================================================================
\section{Conclusion}\label{sec:conclusion}

We introduced \model{}, a parameter-efficient (5.33\,M parameters, 3.04\,GFLOPs, linear in node count) stain-aware nested graph transformer for H\&E histopathology tile classification. Its novelty lies in treating stain chemistry as a signal rather than a nuisance: per-image Macenko optical densities enter local windowed attention as learnable, head-specific edge penalties, condition the soft assignment of patch nodes to tissue compartments, and return to every node through zero-initialised FiLM feedback, so that physical stain information shapes both levels of a learned patch-to-compartment hierarchy without nuclei segmentation or pretrained encoders. Evaluated across six public datasets (150{,}581 internal images), \model{} reaches $91.84\pm2.37\%$ accuracy under strict patient-grouped cross-validation on BreakHis and $96.84\pm0.14\%$ on the independent 50-patient CRC-VAL-HE-7K cohort without adaptation, is best on lymphoma and within 0.03--1.81\,pp of considerably larger from-scratch, ImageNet-fine-tuned and foundation-model baselines elsewhere, and exposes 72 inspectable priors whose learned values differ between datasets in a way consistent with stain chemistry. The ablations identify stain conditioning, rather than the hierarchy as such, as the main source of these gains, and within stain conditioning the per-image H\&E optical-density input supplies almost all of the accuracy ($+0.60$ to $+2.15$\,pp), whereas the stain edge penalty adds only $+0.05$ to $+0.25$\,pp, within seed variation, and is valuable mainly because it makes the use of each dye inspectable. In a single representative run, the priors became haematoxylin-dominant on nuclear-atypia tasks and eosin-leaning in deep blocks on stroma- and keratin-rich tasks, a pattern consistent with stain chemistry but not yet validated against pathologist annotations, and under $\pm30\%$ stain drift the model loses 34--38\% less macro-F1 than Swin-Tiny and ConvNeXt-Tiny. As a contribution of equal weight, the accompanying Source-Separated Image Partitioning protocol shows that random tile splits leak near-duplicates for up to 100\% of test images on widely used benchmarks (86.9\% on LC25000). This means that many published near-ceiling accuracies cannot be read as generalisation, and \ssip{}, with notebooks that regenerate its split files, provides a practical, reproducible alternative when patient metadata are unavailable. Together with the openly available notebooks and a training footprint of 3.6\,GB on a single entry-level GPU, these contributions make \model{} a transparent, reproducible and resource-efficient foundation for slide-level, multi-centre and resource-constrained computational pathology. The Kaggle notebooks that generate all results, tables and figures are available on Zenodo at \url{https://doi.org/\zenodoDOI}; they were run in the Kaggle environment, which is required to run them.

%% =====================================================================
%% DATA AND CODE AVAILABILITY
%% =====================================================================
\section*{Data and code availability}\label{sec:availability}
All six internal datasets and the external CRC-VAL-HE-7K validation cohort are publicly available from their original repositories. BreakHis~\cite{spanhol2016dataset} is distributed by the VRI Laboratory, Federal University of Paran\'a (\url{https://web.inf.ufpr.br/vri/databases/breast-cancer-histopathological-database-breakhis/}) for non-commercial research with acknowledgement of the source and citation of~\cite{spanhol2016dataset}; the distribution page also states a CC BY 4.0 licence. LC25000~\cite{borkowski2019lc25000} is distributed by its authors (\url{https://github.com/tampapath/lung_colon_image_set}) as de-identified images freely available to researchers, without a stated licence. The lymphoma images belong to the IICBU-2008 benchmark of the U.S. National Institute on Aging~\cite{shamir2008iicbu,orlov2010lymphoma}, distributed for free download without a stated licence. Kather2016~\cite{kather2016multiclass} is available from Zenodo (\url{https://doi.org/10.5281/zenodo.53169}) under CC BY 4.0. NCT-CRC-HE-100K (Macenko-normalised release) and CRC-VAL-HE-7K~\cite{kather2019predicting,kather2018nctzenodo} are available from Zenodo (\url{https://doi.org/10.5281/zenodo.1214456}) under CC BY 4.0. The oral cancer images derive from the histopathological imaging database for oral cancer analysis~\cite{rahman2020oral} (Mendeley Data, \url{https://doi.org/10.17632/ftmp4cvtmb.2}), released under CC BY 4.0. Only the notebooks are uploaded: the per-dataset Kaggle notebooks, which contain the \model{} source code, the \ssip{} partitioning code and the training and evaluation code, and which generate all tables and figures, are available on Zenodo at \url{https://doi.org/\zenodoDOI}. Trained weights, split files and other outputs are not archived. The notebooks were run in the Kaggle environment (Kaggle notebooks with NVIDIA Tesla T4 GPUs) and require it to be executed.

\section*{CRediT authorship contribution statement}
Touhid Alam and Sumaiya Sarkar Shimla contributed equally to this work and share first authorship.

\noindent\textbf{Touhid Alam:} Conceptualisation, Methodology, Software, Validation, Formal analysis, Investigation, Data curation, Visualisation, Writing -- original draft, Writing -- review \& editing.

\noindent\textbf{Sumaiya Sarkar Shimla:} Conceptualisation, Methodology, Software, Validation, Formal analysis, Investigation, Data curation, Visualisation, Writing -- original draft, Writing -- review \& editing.

\noindent\textbf{Md. Manzurul Hasan:} Conceptualisation, Methodology, Validation, Formal analysis, Resources, Supervision, Project administration, Writing -- review \& editing.

\noindent\textbf{Md. Rafiqul Islam:} Conceptualisation, Methodology, Validation, Formal analysis, Resources, Supervision, Project administration, Writing -- review \& editing.

\section*{Funding}
This research received no specific grant from any funding agency in the public, commercial or not-for-profit sectors.

\section*{Declaration of competing interest}
The authors declare that they have no known competing financial interests or personal relationships that could have appeared to influence the work reported in this paper.

\section*{Declaration of generative AI and AI-assisted technologies in the writing process}
During the preparation of this work the authors used Claude (Anthropic) in order to improve the readability and language of the manuscript. After using this tool, the authors reviewed, verified and edited all content as needed and take full responsibility for the content of the published article.

\section*{Ethics statement}
This study used only publicly available, de-identified histopathology datasets released by their original authors for research use. No new human data were collected, and no ethics approval was therefore required.

\bibliographystyle{elsarticle-num-names}
\bibliography{references}

\end{document}